\documentclass{article}

\usepackage{PRIMEarxiv}

\usepackage[utf8]{inputenc}
\usepackage[T1]{fontenc}
\usepackage{hyperref}
\usepackage{url}
\usepackage{booktabs}
\usepackage{amsfonts}
\usepackage{nicefrac}
\usepackage{microtype}
\usepackage{graphicx}
\usepackage{multirow}
\usepackage{array}
\usepackage{siunitx}
\usepackage{xcolor}
\usepackage{listings}
\usepackage{longtable}

\usepackage[noabbrev,nameinlink]{cleveref}

\definecolor{codegreen}{rgb}{0,0.6,0}
\definecolor{codegray}{rgb}{0.5,0.5,0.5}
\definecolor{codepurple}{rgb}{0.58,0,0.82}
\definecolor{backcolour}{rgb}{0.97,0.97,0.97}

\title{\includegraphics[width=0.99\textwidth]{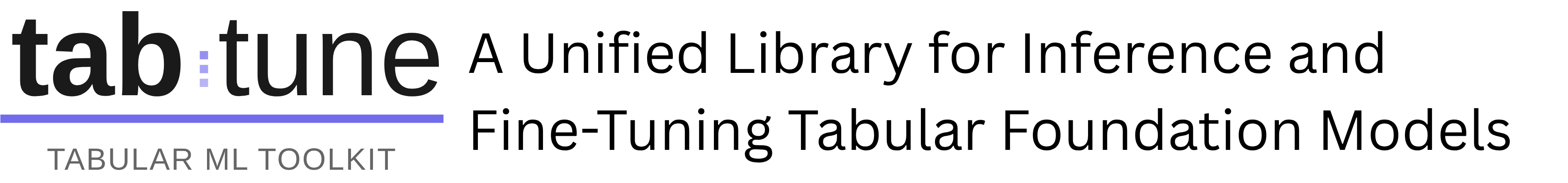}}

\author{
  Aditya Tanna, Pratinav Seth, Mohamed Bouadi \\
  Utsav Avaiya, Vinay Kumar Sankarapu \\
  \affiliation{Lexsi Labs, India \& France}
}
\setabstract{%
Tabular foundation models represent a growing paradigm in structured data learning, extending the benefits of large-scale pretraining to tabular domains. However, their adoption remains limited due to heterogeneous preprocessing pipelines, fragmented APIs, inconsistent fine-tuning procedures, and the absence of standardized evaluation for deployment-oriented metrics such as calibration and fairness.
We present \textsc{TabTune}, a unified library that standardizes the complete workflow for tabular foundation models through a single interface. TabTune provides consistent access to multiple state-of-the-art models supporting multiple adaptation strategies, including zero-shot inference, meta-learning, supervised fine-tuning (SFT), and parameter-efficient fine-tuning (PEFT).
The framework automates model-aware preprocessing, manages architectural heterogeneity internally, and integrates evaluation modules for performance, calibration, and fairness. Designed for extensibility and reproducibility, TabTune enables consistent benchmarking of adaptation strategies of tabular foundation models. The library is open source and available at \url{https://github.com/Lexsi-Labs/TabTune} .
}
\setkeywords{Tabular Foundation Models, Meta-Learning, PEFT, Calibration, Fairness, Benchmarking.}
\runningtitle{TabTune: A Unified Library for Inference and Fine-Tuning Tabular Foundation Models}

\begin{document}
\maketitle

\section{Introduction}

Tabular data i.e. structured collections of records (rows) and attributes (columns)—underpins a vast array of real-world applications, from healthcare and finance to industrial analytics and scientific research. These datasets form the backbone of modern decision-making systems: hospitals rely on patient records for diagnostic risk prediction, financial institutions assess creditworthiness and fraud using structured features, and scientific studies aggregate multi-sensor measurements to discover new insights. Yet despite this ubiquity, deep learning on tabular data has long lagged behind the breakthroughs achieved in domains such as natural language processing and computer vision. 

Traditional approaches like gradient-boosted decision trees (e.g., XGBoost~\cite{chen2016xgboost}, LightGBM~\cite{ke2017lightgbm}, CatBoost~\cite{prokhorenkova2018catboost}) remain dominant because they reliably handle heterogeneous features, variable sample sizes, and small datasets—challenges that standard neural networks often struggle with. Deep models frequently underperform on tabular benchmarks due to the lack of spatial or temporal structure, making feature extraction less amenable to convolutional or sequential architectures. Moreover, tabular datasets often exhibit high heterogeneity—mixing numerical, categorical, and ordinal variables—and suffer from missing values and skewed distributions, which violate many of the inductive biases that make deep learning successful in other modalities. As a result, practitioners continue to favor tree ensembles and gradient boosting for their robustness, interpretability, and low data requirements.

Recently, the advent of \emph{tabular foundation models (TFMs)} has begun to reshape this landscape. These models extend the ``foundation model'' paradigm—pretraining on large, diverse corpora followed by downstream adaptation—to structured data domains. Models such as \textsc{TabPFN}~\cite{tabpfn,TabPFN2,tabpfn25}, \textsc{TabICL}~\cite{tabicl}, \textsc{ContextTAB}~\cite{contexttab}, \textsc{OrionMSP}~\cite{orionmsp2025}, and \textsc{OrionBiX}~\cite{orionbix2025} leverage large-scale pre-training on synthetic or massive tabular datasets to provide generic, out-of-the-box learning capabilities. These developments mark a significant step toward ``general-purpose'' tabular learners that can adapt rapidly to new datasets with minimal supervision. 

However, despite their promise, adopting these models in real-world practice remains hampered by several frictions:

\begin{itemize}
  \item \textsc{Diverse preprocessing requirements}: Each model demands its own data encoding, feature normalization, and missing-value strategy—requiring practitioners to build bespoke pipelines for each model family. For example, TabPFN expects numerically encoded categorical features consistent with its synthetic priors, whereas TabICL and Orion architectures rely on learned embeddings or set-transformer encoders for categorical attributes.
  \item \textsc{Fragmented APIs and training protocols}: Some models operate purely in zero-shot inference mode (zero fine-tuning), others support supervised fine-tuning (SFT) or parameter-efficient fine-tuning (PEFT), specifically Low-Rank Adaptation (LoRA; see Section~\ref{sec:related_work} for details). Managing these disparate workflows is laborious and error-prone, particularly when comparing multiple model families under consistent experimental settings.
  \item \textsc{Evaluation gaps in deployment-relevant metrics}: While accuracy improvements are well documented, aspects such as calibration of probability estimates, fairness across subgroups, and resource-efficiency trade-offs are under-explored in a unified framework. This absence of standardization makes it difficult to assess whether performance gains translate to trustworthy or deployable behavior.
  \item \textsc{Model selection complexity}: With multiple models and tuning strategies available, users face non-trivial questions: Which model best suits a small vs. large dataset? What is the memory/latency trade-off? How do calibration and fairness behave under different tuning regimes? In practice, this uncertainty often discourages the adoption of TFMs despite their potential benefits.
\end{itemize}

To bridge these gaps, we introduce \textsc{TabTune}, a unified, scikit-learn-compatible toolkit that standardizes the full modelling workflow for tabular foundation models. TabTune offers:

\begin{itemize}
  \item A \textsc{single interface} for multiple TFM families, handling model-specific preprocessing internally and exposing consistent fit–predict semantics.
  \item Support for a \textsc{spectrum of adaptation strategies}: \texttt{zero-shot inference}, \texttt{meta-learning}, \texttt{full supervised fine-tuning (SFT)}, and \texttt{parameter-efficient fine-tuning (PEFT)}. This design allows practitioners to switch between inference paradigms seamlessly using a unified API.
  \item Built-in diagnostics for \textsc{calibration} (Expected Calibration Error, Maximum Calibration Error, Brier score) and \textsc{fairness} (Statistical Parity Difference, Equalised Odds Difference, Equalised Opportunity Difference), enabling a holistic assessment of model trustworthiness and risk.
  \item A systematic benchmarking module across standard tabular suites (e.g., TALENT~\cite{talent}, OpenML-CC18~\cite{openmlcc18}) that evaluates accuracy, calibration, fairness, and resource efficiency in a unified ranking framework, facilitating reproducible and comparative studies of TFM performance.
\end{itemize}

With TabTune, practitioners can quickly experiment with multiple model/tuning configurations using consistent functions (e.g., \texttt{.fit()}, \texttt{.predict()}, \texttt{.evaluate()}), without rewriting preprocessing logic or training loops. Moreover, TabTune serves as an experimental bed for studying the interplay between learning paradigms—such as zero-shot inference versus parameter-efficient tuning—and their downstream effects on calibration, fairness, and computational efficiency. For detailed system architecture, see Section~\ref{sec:system_design}. Through comprehensive experiments, we demonstrate that TabTune not only simplifies workflows, but also provides new insights into the trade-off space of accuracy, calibration, fairness and efficiency across tabular foundation models.

In the remainder of the paper, Section \ref{sec:related_work} reviews prior work on tabular foundation models and adaptation techniques. Section \ref{sec:system_design} details the system design of TabTune, Section \ref{sec:experiments} describes our experimental setup and benchmark results, and Section \ref{sec:discussion} offers practical guidance on model selection, discusses limitations, and outlines future directions.

\section{Related Work}
\label{sec:related_work}

\subsection{Traditional Machine Learning and Deep Learning for Tabular Data}

Tabular data—structured as records (rows) and features (columns)—is pervasive across applications such as healthcare, finance, industrial analytics, and scientific research. For decades, models based on gradient-boosted decision trees (GBDTs) such as XGBoost, LightGBM, and CatBoost have dominated supervised learning on tabular tasks. These methods reliably handle heterogeneous feature types, missing values, and small to medium sample sizes, while requiring only modest hyper-parameter tuning. Their robustness and simplicity have made them the de facto standard for tabular prediction tasks in both research and production settings.

In recent years, deep neural networks—such as feed-forward multilayer perceptrons, attention-based architectures, and hybrid tree–neural models—have been applied to tabular data~\cite{borisov2022tabular,somvanshi2024tabular}. Despite significant progress, these models have often struggled to consistently outperform GBDTs in real-world benchmarks. This gap arises from several challenges: the heterogeneity of tabular data, limited dataset sizes, complex feature–target relationships, and the absence of large-scale pre-training strategies analogous to those that revolutionized vision and language modeling.

\subsection{Tabular Foundation Models}

A major paradigm shift has recently emerged with the advent of \emph{tabular foundation models (TFMs)}—large pretrained architectures that aim to generalize across diverse tasks, often through in-context learning (ICL) or single forward-pass inference without task-specific gradient updates. 
In the context of TFMs, \emph{zero-shot inference} refers to the "fit then predict" paradigm: during the fit phase, the model receives training samples as context (without gradient-based parameter updates), and during the predict phase, it performs in-context learning through forward passes alone to make predictions on test samples. Unlike traditional machine learning where "fit" implies gradient updates and training, in TFMs the fit step simply provides context samples to the pretrained model, enabling rapid adaptation without requiring training time. 

\textsc{TabPFN} (Tabular Prior-data Fitted Network) represents one of the earliest and most influential TFMs. It employs a transformer architecture trained on vast collections of synthetic tabular datasets, effectively learning to approximate Bayesian inference across varying generative priors. This pre-training strategy enables TabPFN to achieve strong few-shot and zero-shot performance on small to medium datasets (typically up to $\sim$10,000 samples), offering competitive accuracy without explicit fine-tuning~\cite{tabpfn,TabPFN2,tabpfn25}. 
Building on this foundation, \textsc{TabICL} extends the in-context learning paradigm to larger-scale tabular datasets by introducing a novel two-stage \emph{column-then-row} attention mechanism. Through large-scale synthetic pre-training (up to $\sim$60K samples) and robust scaling to half a million samples at inference time, TabICL demonstrates that transformer-based ICL can challenge traditional tree ensembles even on large tabular domains~\cite{tabicl}. 
Another line of work, \textsc{TabDPT}, introduces an in-context learning (ICL)-based tabular foundation model trained with denoising-style, self-supervised objectives on real-world tables~\cite{tabdpt}. In parallel, other approaches explore the role of \emph{mixed synthetic priors}, showing that combining causal, statistical, and randomized priors during pre-training can substantially improve downstream generalization and calibration performance across classification and regression tasks.

Collectively, these developments illustrate that, with sufficient architectural bias and large-scale pre-training, transformer-based TFMs can achieve broad generalization on structured data—challenging long-held assumptions about the superiority of tree-based ensembles in tabular learning. Recent comprehensive surveys have catalogued the landscape of tabular foundation models~\cite{jiang2025representation,badaro2023transformers}, highlighting their potential to transform tabular learning. Recent work has demonstrated that careful fine-tuning strategies can yield significant improvements for TFMs~\cite{rubachev2025finetuning}. Meanwhile, researchers have argued that tabular foundation models should be a research priority~\cite{breugel2024position}, given their potential impact across domains.

\textsc{OrionMSP}~\cite{orionmsp2025} introduces a transformer architecture for \emph{tabular in-context learning} that combines multi-scale sparse attention—capturing both local and global feature dependencies efficiently—with a Perceiver-style latent memory that enables safe bidirectional communication between training and test samples. It uses column-wise embeddings via Set Transformers to model feature distributions and a split-masked causal attention mechanism for proper ICL reasoning. The design achieves near-linear attention scaling while maintaining high accuracy across diverse datasets, rivaling or outperforming TabICL and TabPFN—especially on heterogeneous, high-dimensional, and imbalanced datasets—thus demonstrating robust, efficient, and generalizable tabular learning without gradient updates.
Similarly, \textsc{OrionBiX}~\cite{orionbix2025} employs a biaxial attention mechanism for tabular in-context learning, enabling bidirectional context modeling across both rows and columns of tabular data. This dual-axis attention pattern allows the model to capture complex feature interactions and sample relationships simultaneously, providing a more comprehensive representation of tabular structure. The biaxial design facilitates effective information flow between training and test samples, enabling strong in-context learning performance across diverse tabular datasets while maintaining computational efficiency.

\subsection{Representation Learning and Alternative Approaches}

While transformer-based TFMs dominate recent work, alternative representation learning paradigms have been explored. Graph neural networks have been adapted for tabular data, modeling feature relationships through graph structures~\cite{li2024gnn}. Extensions of foundation models to time series forecasting demonstrate their versatility; for instance, TabPFN-v2 has shown competitive performance on temporal tasks~\cite{hoo2025tables}, suggesting that tabular foundation models may generalize beyond traditional classification settings.

\subsection{Tabular ML Toolkits and Frameworks}

Several frameworks have emerged to facilitate deep learning with tabular data, each addressing different aspects of the modeling pipeline. \textsc{AutoGluon}~\cite{erickson2020autogluon} is an AutoML framework that automates model selection and hyperparameter tuning for tabular data, employing neural architecture search and ensemble methods to achieve strong predictive performance with minimal manual configuration. \textsc{PyTorch Tabular}~\cite{joseph2021pytorch} provides a modular framework for deep learning with tabular data, offering reusable components for neural network architectures tailored to structured data. However, these frameworks primarily focus on training models from scratch or performing architecture search, rather than working with pretrained tabular foundation models.

In contrast, \textsc{TabTune} is the first unified framework specifically designed for tabular foundation models, bridging the gap between pretrained TFM architectures and practical deployment workflows. While existing toolkits excel at discovering or building models from scratch, TabTune addresses the unique challenges of working with pretrained TFMs: standardizing diverse preprocessing requirements, unifying adaptation strategies (zero-shot inference, meta-learning, supervised fine-tuning, and parameter-efficient fine-tuning), and providing built-in evaluation diagnostics for calibration and fairness. This specialization enables practitioners to leverage the powerful generalization capabilities of pretrained TFMs without the complexity of managing model-specific interfaces and workflows.

\subsection{Adaptation Techniques: Meta-Learning, Fine-Tuning, and PEFT}

Although inference-only TFMs exhibit strong generalization, many real-world applications benefit from adapting to target distributions. Three key adaptation paradigms have emerged.

\textbf{Meta-learning} (or episodic fine-tuning) uses support–query splits from training data to retain in-context generalization while improving task-specific performance~\cite{finn2017model}. This approach has proven effective for tabular foundation models such as TabICL, where episodic training enhances accuracy and supports few-shot adaptation to new distributions~\cite{tabicl,tabdpt}.

\textbf{Supervised fine-tuning (SFT)} updates all model parameters on labeled data when sufficient resources are available~\cite{howard2018universal}. In tabular domains, full fine-tuning often yields notable gains over inference-only use, especially on large datasets where models learn domain-specific patterns~\cite{tabicl,tabdpt}. However, it remains computationally and storage intensive, as every parameter must be updated per task.

\textbf{Parameter-efficient fine-tuning (PEFT)} methods, such as Low-Rank Adaptation (LoRA)~\cite{hu2021lora}, limit updates to low-dimensional subspaces, greatly reducing compute and memory costs while preserving performance. Surveys categorize PEFT into additive (adapters), selective (partial updates), and reparameterization-based (low-rank) approaches~\cite{liu2024peft}. Empirical studies show that PEFT matches or surpasses full fine-tuning with far lower resource demands~\cite{razuvayevskaya2024peft}. For tabular foundation models, PEFT enables efficient domain adaptation under tight resource budgets, supporting rapid experimentation and scalable deployment. Together, these paradigms bridge the gap between zero-shot inference and fully supervised modeling, promoting adaptable TFMs for diverse applications.

\subsection{Calibration, Fairness, and Responsible Deployment}

As tabular foundation models progress from research to deployment, challenges of trustworthiness, calibration, fairness, and robustness have become increasingly central. 

\textbf{Calibration reliability}—the alignment between predicted probabilities and actual outcomes—is vital for high-stakes decision-making. Guo et al.~\cite{guo2017calibration} showed that modern neural networks often exhibit systematic miscalibration, motivating metrics such as Expected Calibration Error (ECE) and Maximum Calibration Error (MCE) to measure these gaps. For tabular foundation models, early results suggest that pretrained transformers yield better-calibrated confidence estimates than classical ensembles, especially in low-data settings where in-context learning supports more nuanced uncertainty estimation.

\textbf{Fairness} requires that models avoid systematic bias across demographic groups~\cite{barocas2019fairness}. Fairness frameworks propose criteria such as \textit{equalized odds}~\cite{hardt2016equality}—requiring equal true and false positive rates—and \textit{equal opportunity}, focusing on true positive rate parity. \cite{pleiss2017fairness} revealed a fundamental tension between perfect calibration and strict fairness, showing that these objectives can conflict. In-context learning setups may further amplify biases present in demonstration data. Mitigation strategies include group-balanced sampling, fair demonstration selection, and uncertainty-based data curation.

\textbf{Robustness}—the stability of model behavior under distributional shift, perturbations, and adversarial attacks—remains an open challenge. While tabular foundation models benefit from large-scale, heterogeneous pretraining, their behavior under shift and noise is not yet fully characterized. These considerations highlight the need for unified evaluation frameworks that jointly assess predictive performance, calibration, fairness, and robustness to ensure the responsible deployment of tabular foundation models.

\section{System Design}
\begin{figure}
    \centering
    \includegraphics[width=0.54\linewidth]{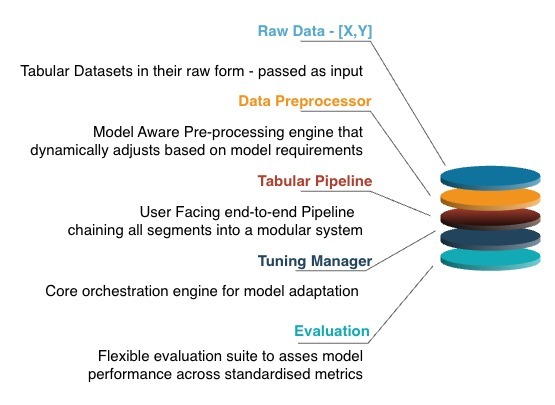}
    \caption{Overview of the TabTune architecture. Raw tabular data are processed through a model-aware pre-processing engine and modular tuning pipeline, coordinated by a central orchestration engine with integrated evaluation assessment modules.}
    \label{fig:placeholder}
\end{figure}

\begin{figure}[pt]
    \centering
    \includegraphics[width=0.93\textwidth]{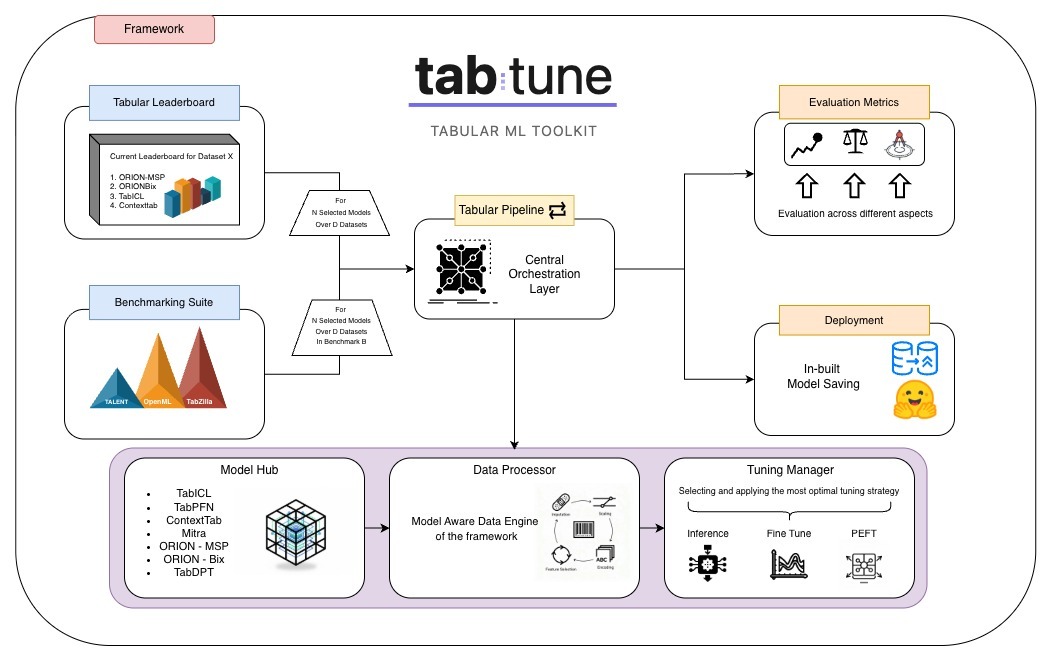}
    \caption{Modular architecture of TabTune. The \texttt{TabularPipeline} orchestrates the workflow by chaining the model-aware \texttt{DataProcessor}, the encapsulated TFM, and the adaptive \texttt{TuningManager}, forming a cohesive and control framework.}
    \label{fig:architecture}
\end{figure}

The design of \textbf{TabTune} emphasizes abstraction of the operational complexity associated with modern tabular foundation models (TFMs), while preserving flexibility for advanced adaptation, benchmarking, and experimentation. The framework adopts a strictly modular architecture in which preprocessing, tuning, and orchestration layers are decoupled yet interoperable, ensuring extensibility and transparency across research and production workflows (Figure~\ref{fig:architecture}).

\subsection{Design Principles}

\textbf{Unified Interface.}  
TabTune exposes a consistent, high-level API that provides unified access to all supported TFMs, including TabPFN, TabDPT, TabICL, Mitra, ContextTab, OrionMSP and OrionBiX . Through a shared syntax for model initialization, training, and evaluation, it eliminates the need for users to adapt to model-specific conventions or data representations, enabling seamless experimentation across architectures.

\textbf{Model-Aware Automation.}  
The framework automatically identifies the chosen model and configures preprocessing, hyperparameters, and adaptation procedures accordingly. Whether deploying a zero-shot TabPFN, performing episodic meta-learning with TabICL, or applying parameter-efficient LoRA fine-tuning, TabTune dynamically selects optimized defaults aligned with the model’s internal design and capacity. This automation reduces manual configuration overhead and ensures consistency across runs.

\textbf{Unified Workflow Compatibility.}  
All models in TabTune follow a consistent workflow based on standardized method calls such as \texttt{fit()}, \texttt{predict()}, and \texttt{evaluate()}. This design allows easy integration into broader machine learning pipelines and benchmarking frameworks, enabling reproducible and interpretable experimentation without requiring specialized wrappers or retraining scripts.

\subsection{Modular Architecture}
\label{sec:system_design}

TabTune’s architecture is composed of four interdependent modules that collectively handle data processing, model adaptation, inference, and evaluation. These components are designed to remain logically independent yet harmonized through a unified orchestration layer.

\begin{itemize}
    \item \textbf{TabularPipeline} module, which serves as the main user interface and orchestrator of the modeling workflow. It encapsulates the selected TFM, manages persistence through \texttt{save()} and \texttt{load()} utilities, and coordinates the complete pipeline—from preprocessing and fine-tuning to evaluation and model comparison.
    \item The \textbf{DataProcessor} component is responsible for handling model-specific preprocessing logic. Upon initialization, it dynamically loads specialized processors and performs imputation, normalization, encoding, and embedding generation as required.
    \item The \textbf{TuningManager} module constitutes the computational backbone of TabTune. It executes the selected adaptation strategy—ranging from zero-shot inference to supervised fine-tuning, meta-learning, and parameter-efficient fine-tuning (PEFT). The manager implements training routines in a unified format, abstracts gradient updates, and integrates lightweight adaptation methods such as LoRA to reduce computational and memory costs while maintaining model generalization. For meta-learning configurations, episodic training is supported through dynamic support–query sampling, preserving the in-context adaptability of transformer-based models.
    \item Complementing these components, the \textbf{TabularLeaderboard} module enables systematic evaluation and large-scale comparison of model variants. It automates training and evaluation over consistent data partitions and produces ranked summaries across metrics such as AUC, F1-score, and inference efficiency. This functionality transforms comparative model assessment into a reproducible and configurable experiment suite.
\end{itemize}

\subsection{Supported Models}

TabTune includes built-in support for seven prominent tabular foundation models, default tuning configurations, and aligned preprocessing pipelines. This design ensures that every model can be instantiated, fine-tuned, and evaluated with minimal configuration effort, while preserving model-specific fidelity.

\begin{table}[pt]
\centering
\caption{Tabular foundation models supported in TabTune. Paradigm indicates the learning approach—ICL, Prior-Fitted Network, Denoising Transformer, or Scalable ICL. Key Feature summarizes each model’s core architectural element (e.g., attention).}
\label{tab:models}
\footnotesize
\begin{tabular}{lll}
\toprule
Model & Paradigm & Key Feature \\
\midrule
OrionMSP & In-Context Learning & Multi-scale sparse attention with Perceiver-style latent memory \\
OrionBiX & In-Context Learning & Biaxial attention with bidirectional context modeling \\
TabICL & In-Context Learning & Two-stage column-then-row attention for large-scale tabular ICL \\
Mitra & Scalable ICL & 2D attention (row and column) with mixed synthetic priors \\
ContextTab & Semantics-Aware ICL & Utilizes modality-specific semantic embeddings \\
TabDPT & In-Context Learning & Employs diffusion-based self-supervised pre-training \\
TabPFN & Prior-Fitted Network & Approximates Bayesian inference on synthetic priors \\
\bottomrule
\end{tabular}
\end{table}
%Together, these models span the principal paradigms of modern tabular foundation modeling, from Bayesian-inspired inference to semantic and denoising-based representation learning.

\section{Implementation}
\label{sec:implementation}

This section describes how \textbf{TabTune}'s implementation is organized around three core components—the \texttt{DataProcessor}, \texttt{TuningManager}, and \texttt{TabularPipeline}—which together enable seamless preprocessing, adaptation, and inference through a standardized interface.

\subsection{Model-Aware Preprocessing}
The \texttt{DataProcessor} dynamically loads model-specific routines to format data correctly with minimal user setup.
For \textsc{TabPFN}, features are mapped to numerical or integer-encoded tensors to align with its synthetic priors. \textsc{ContextTab}~\cite{contexttab} adds semantic vectors for column names and categorical values. \textsc{TabICL} and \textsc{Mitra} apply normalization and model-tailored transformations.
For \textsc{OrionMSP}~\cite{orionmsp2025} and \textsc{OrionBiX}~\cite{orionbix2025}, preprocessing is distribution- and hierarchy-aware: columns are embedded with Set Transformer-based induced set attention on train rows only, numerical features are standardized and categorical values embedded, multi-scale row inputs and block-sparse attention masks are constructed, Perceiver-style memory is written by train rows and read by all rows, and label injection uses split-masked attention to prevent test-to-train leakage. Compared to \textsc{TabICL}~\cite{tabicl}, ORION adds distribution-aware column embeddings and multi-scale sparsity for scalable, ICL-safe processing.
Together, these routines ensure each TFM receives inputs in its expected schema.

\subsection{Flexible Tuning Strategies}

The \texttt{TuningManager} module governs model adaptation and optimization, executing the selected strategy specified during pipeline initialization. It provides a unified training controller that supports four complementary regimes—inference, supervised fine-tuning, meta-learning, and parameter-efficient fine-tuning (PEFT)—under a common abstraction.
\begin{itemize}
    \item \textbf{Zero-shot inference}, TabTune performs zero-shot prediction using pretrained weights, exploiting the generalization capacity of models such as TabPFN for small datasets without parameter updates.
    \item \textbf{Supervised fine-tuning} extends this by optimizing all model parameters through gradient-based learning on labeled data, allowing users to specify training details (e.g., number of epochs or learning rate) via the configuration dictionary. For in-context learning models such as TabICL and Mitra, batches are split into pseudo-episodes where the first half serves as support context and the second half as query targets, maintaining the episodic structure required by the model architecture.
    \item \textbf{Meta-learning fine-tuning} introduces episodic adaptation, where models are trained on dynamically sampled support–query pairs to maintain in-context generalization while improving task-specific accuracy. Each episode randomly samples disjoint support and query sets from the training data, with class labels normalized to contiguous indices within each episode to ensure compatibility with fixed-size classification heads. This approach is particularly effective for transformer-based architectures like TabICL and TabDPT that rely on in-context reasoning.
    \item \textbf{Parameter-Efficient fine-tuning (PEFT)}—implemented via Low-Rank Adaptation (LoRA)—reduces computational cost by updating only small low-rank matrices within attention layers. LoRA is configured with rank $r=8$ and scaling factor $\alpha=16$, injecting trainable adapters into attention projection layers. When LoRA application fails due to architectural constraints, the framework automatically falls back to standard fine-tuning. This technique significantly lowers memory requirements while preserving performance and is supported for both supervised and meta-learning modes.
\end{itemize}

\subsubsection{PEFT Implementation Details:} 
The fine-tuning framework determines target layers for parameter-efficient adaptation automatically based on model architecture. Updates are typically restricted to attention projections and core transformer components, while dynamic or task-specific heads are excluded. If no predefined targets are detected, adapters are applied to all eligible linear layers. When adapter injection is incompatible with the underlying architecture, the system reverts to standard full-parameter fine-tuning.

\subsubsection{Episodic Training Mechanics:} 
For meta-learning adaptation, episodes are formed by sampling disjoint support and query sets from the training data. Each episode contains $S$ labeled support examples and $Q$ query examples used for evaluation. The model is conditioned on the support labels and optimized to predict the query labels within each episode. To ensure compatibility with fixed-size classification heads, class labels are remapped to contiguous indices $\{0, 1, \ldots, K-1\}$ according to the support set. Episodes in which query samples include unseen classes are excluded to preserve consistency.

\begin{table}[pt]
\centering
\caption{Fine-tuning strategy support matrix across tabular foundation models. $\checkmark$ = full support; $\ast$ = experimental (may revert to base fine-tuning); -- = not supported. SFT = Supervised Fine-Tuning; Meta = Meta-Learning; PEFT SFT = Parameter-Efficient Supervised Fine-Tuning; PEFT Meta = Parameter-Efficient Meta-Learning.}
\label{tab:finetune_support}
\footnotesize
\begin{tabular}{lcccc}
\toprule
Model & SFT & Meta-Learning & PEFT SFT & PEFT Meta-Learning \\
\midrule
OrionMSP & $\checkmark$ & $\checkmark$ & $\checkmark$ & $\checkmark$ \\
OrionBiX & $\checkmark$ & $\checkmark$ & $\checkmark$ & $\checkmark$ \\
TabPFN & $\checkmark$ & $\checkmark$ & $\ast$ & $\ast$ \\
TabICL & $\checkmark$ & $\checkmark$ & $\checkmark$ & $\checkmark$ \\
TabDPT & $\checkmark$ & $\checkmark$ & $\checkmark$ & $\checkmark$ \\
Mitra & $\checkmark$ & $\checkmark$ & $\checkmark$ & $\checkmark$ \\
ContextTab & $\checkmark$ & -- & $\ast$ & -- \\
\bottomrule
\end{tabular}
\end{table}

\subsubsection{Data Sampling Strategies}
\label{sec:data_sampling}
The framework incorporates optional resampling techniques to address label imbalance and promote a well-diversified dataset for model context, while remaining independent of the core modeling workflow. Resampling is applied exclusively to the training split to avoid any data leakage into the validation or test sets.

\textbf{Supported methods :} The available options include: \texttt{smote} (Synthetic Minority Over-sampling Technique), \texttt{random\_over} (random oversampling), \texttt{random\_under} (random undersampling), \texttt{tomek} (Tomek links removal), \texttt{kmeans} (cluster centroids undersampling), and \texttt{knn} (neighborhood cleaning rule). These methods enable flexible control over class balance and support reproducible experimentation across diverse tabular datasets.

\subsection{Evaluation Utilities}

TabTune provides comprehensive evaluation capabilities across performance, calibration, and fairness. 
Beyond these standard metrics, TabTune also includes built-in utilities for assessing model calibration and fairness—critical dimensions for responsible deployment in high-stakes applications.

\begin{itemize}
    \item The \texttt{evaluate()} method computes standard classification metrics including accuracy, precision, recall, F1-score, and AUC-ROC, enabling rapid assessment of predictive performance across models and tuning strategies (Listing~\ref{lst:performance_eval}).
    \item The \texttt{evaluate\_calibration()} method computes Expected Calibration Error (ECE), Maximum Calibration Error (MCE), and Brier score to quantify the reliability of predicted probabilities. These metrics reveal whether a model's confidence estimates align with actual prediction accuracy, essential for decision-making systems in healthcare, finance, and autonomous applications.
    \item The \texttt{evaluate\_fairness()} method assesses demographic parity and equalized odds across sensitive attributes, measuring whether predictions exhibit systematic bias across subgroups. Users specify sensitive features (e.g., gender, race, age) and receive Statistical Parity Difference (SPD), Equalised Odds Difference (EOD), and more, enabling practitioners to identify and mitigate algorithmic bias before deployment.
\end{itemize}
\subsection{Installation}
\label{sec:installation}

TabTune is designed to be easily integrated into existing Python environments. The library is open-source and available on the Python Package Index (PyPI), ensuring straightforward installation for practitioners.

To install the latest stable release, users can execute the following command:

\begin{lstlisting}[language=bash]
pip install tabtune
\end{lstlisting}

Alternatively, for the latest development version, the package can be installed directly from the source repository:

\begin{lstlisting}[language=bash]
git clone https://github.com/Lexsi-Labs/TabTune.git
cd TabTune
pip install -r requirements.txt
pip install -e .
\end{lstlisting}

\subsection{Pipeline Abstraction and Workflow}

The \texttt{TabularPipeline} serves as the primary orchestration layer for integrating the preprocessing and tuning components. It manages the end-to-end flow—data ingestion, preprocessing, adaptation, and prediction—through a unified API that abstracts low-level operations.

The following demonstrates the core workflow for model initialization, training, and inference. The pipeline detects the model type, loads the corresponding preprocessing backend, and delegates adaptation to the \texttt{TuningManager}. In this example, the user specifies only the model and adaptation mode. TabTune internally triggers three automated steps: (1) the \texttt{DataProcessor} formats the dataset to match the model's expected schema, (2) the \texttt{TuningManager} selects and executes the correct adaptation strategy, and (3) the \texttt{TabularPipeline} aggregates predictions and handles evaluation. This design illustrates how the framework abstracts operational complexity behind a minimal interface, enabling quick prototyping and reproducible experimentation.

\begin{lstlisting}[language=Python, label=lst:performance_eval]
from tabtune import TabularPipeline
# Initialize the pipeline with model configuration
pipeline = TabularPipeline(model_name="OrionMSP",tuning_strategy="inference")
# Train and predict
pipeline.fit(X_train, y_train)
predictions = pipeline.predict(X_test)
\end{lstlisting}

\subsection{Benchmarking with TabularLeaderboard}

To facilitate large-scale model comparison and reproducibility, TabTune includes the \texttt{TabularLeaderboard} module. This component automates benchmarking across multiple TFMs and adaptation modes under consistent data splits, producing ranked summaries based on selected metrics such as AUC, F1-score, and training efficiency.  
Listing~\ref{lst:leaderboard} shows a representative use case where several models and tuning regimes are benchmarked on the same dataset. Each configuration defines its own adaptation strategy and hyperparameters, while TabTune ensures standardized preprocessing and evaluation across all runs.The leaderboard system illustrates the high-level orchestration capabilities of TabTune's architecture. Internally, it constructs multiple \texttt{TabularPipeline} instances, each managing its respective preprocessing and tuning configuration, and executes them in parallel under uniform conditions. The results are aggregated and ranked automatically, providing a reproducible, interpretable, and scalable benchmarking framework for evaluating tabular foundation models.
Following demonstrates both evaluation modes, which can be applied to any TabTune model after training.
\newpage
\begin{lstlisting}[language=Python, label=lst:leaderboard]
from tabtune import TabularLeaderboard
# Initialize leaderboard
leaderboard = TabularLeaderboard(X_train, X_test, y_train, y_test)
# Add model configurations
leaderboard.add_model(model_name='TabPFN',tuning_strategy='inference')
leaderboard.add_model(model_name='OrionBix',tuning_strategy='finetune',
                                        tuning_params={'epochs': 5})
leaderboard.add_model(model_name='OrionMSP',tuning_strategy='inference')
# Execute and rank models
results = leaderboard.run(rank_by='roc_auc_score')
\end{lstlisting}

\section{Usage Scenarios}
\label{sec:usage_scenarios}

This section demonstrates TabTune's capabilities through several scenarios, showcasing the framework's functionality. These examples provide hands-on illustrations of the system components described in Section~\ref{sec:implementation}.
\
\subsection{Basic Usage and Multi-Model Switching}
TabTune's unified API allows seamless experimentation across foundation models with identical workflows. Switching between the seven supported models requires only a parameter change—no need to learn new APIs or preprocessing steps. This consistency enables rapid prototyping, performance comparison, and deployment without rewriting data or training code. Model persistence ensures reproducibility and version control.

\begin{lstlisting}[language=Python]
from tabtune import TabularPipeline
# Example 1: TabPFN Model
pipeline_pfn = TabularPipeline(model_name="TabPFN",tuning_strategy="inference")
pipeline_pfn.fit(X_train, y_train)
predictions_pfn = pipeline_pfn.predict(X_test)
# Example 2: OrionMSP Model with semantic features
pipeline_OrionMSP = TabularPipeline(model_name="OrionMSP",tuning_strategy="inference")
pipeline_OrionMSP.fit(X_train, y_train)
predictions_OrionMSP = pipeline_OrionMSP.predict(X_test)
\end{lstlisting}

\subsection{Model Comparison and Results Analysis}
This scenario showcases systematic benchmarking of multiple models and tuning strategies with detailed performance insights. The leaderboard automates cross-validation, metric computation, and ranking across all configurations. Results can be exported as structured DataFrames for analysis, visualization, or automated model selection—enabling data-driven decisions on the best model–tuning combination for a given dataset and deployment setting.

\begin{lstlisting}[language=Python]
from tabtune import TabularLeaderboard
leaderboard = TabularLeaderboard(X_train, X_test,y_train, y_test)
# Add multiple model configurations
leaderboard.add_model(model_name='TabPFN',tuning_strategy='inference')
leaderboard.add_model(model_name='TabICL',tuning_strategy='finetune',tuning_params={'epochs':5})
leaderboard.add_model(model_name='OrionMSP',tuning_strategy='inference')
# Run comparison and get detailed results
results = leaderboard.run(rank_by='roc_auc_score')
\end{lstlisting}

\subsection{Fine-Tuning Demonstration}

This scenario demonstrates the different fine-tuning strategies available in TabTune. Each approach offers distinct trade-offs between performance, and generalization capability.

\subsubsection{Supervised Fine-Tuning (SFT)}

Supervised fine-tuning optimizes all model parameters using standard mini-batch gradient descent. This approach typically achieves the highest accuracy when sufficient training data is available.

\begin{lstlisting}[language=Python]
from tabtune import TabularPipeline
# Full supervised fine-tuning
sft_pipeline = TabularPipeline(
    model_name='OrionMSP',
    tuning_strategy='finetune',
    tuning_params={'finetune_mode': 'sft','epochs': 5,'learning_rate': 1e-4}
)
sft_pipeline.fit(X_train, y_train)
sft_metrics = sft_pipeline.evaluate(X_test, y_test)
\end{lstlisting}

\subsubsection{Meta-Learning Fine-Tuning}

Meta-learning fine-tuning uses episodic training, where the model learns to adapt from support sets to query sets within each episode. This approach preserves in-context learning capabilities while improving task-specific performance.

\begin{lstlisting}[language=Python]
from tabtune import TabularPipeline
# Meta-learning fine-tuning
meta_pipeline = TabularPipeline(
    model_name='OrionMSP',
    tuning_strategy='finetune',
    tuning_params={
        'finetune_mode': 'meta-learning', 
        'epochs': 3, 'learning_rate': 5e-5, 'support_size': 48, 'query_size': 32
        }
)
meta_pipeline.fit(X_train, y_train)
meta_metrics = meta_pipeline.evaluate(X_test, y_test)
\end{lstlisting}

\subsubsection{Parameter-Efficient Fine-Tuning with SFT (PEFT SFT)}

PEFT SFT combines LoRA adapters with supervised fine-tuning, achieving comparable performance to full fine-tuning while reducing memory usage by 60--80\%.

\begin{lstlisting}[language=Python]
from tabtune import TabularPipeline
# PEFT with supervised fine-tuning
peft_sft_pipeline = TabularPipeline(
    model_name='TabICL',
    tuning_strategy='peft',
    tuning_params={
        'finetune_mode': 'sft', 'epochs': 5,
        'peft_config': {'r': 8,'lora_alpha': 16,'lora_dropout': 0.05}
    }
)
peft_sft_pipeline.fit(X_train, y_train)
peft_sft_metrics = peft_sft_pipeline.evaluate(X_test, y_test)
\end{lstlisting}

\subsubsection{Parameter-Efficient Fine-Tuning with Meta-Learning (PEFT Meta-Learning)}

PEFT meta-learning combines LoRA adapters with episodic training, offering the benefits of both parameter efficiency and in-context generalization.
\newpage
\begin{lstlisting}[language=Python]
from tabtune import TabularPipeline
# PEFT with meta-learning
peft_meta_pipeline = TabularPipeline(
    model_name='TabICL',
    tuning_strategy='peft',
    tuning_params={'finetune_mode': 'meta-learning', 'epochs': 3, 
    'support_size': 48, 'query_size': 32,
    'peft_config': {'r': 8, 'lora_alpha': 16, 'lora_dropout': 0.05}}
)
peft_meta_pipeline.fit(X_train, y_train)
peft_meta_metrics = peft_meta_pipeline.evaluate(X_test, y_test)
\end{lstlisting}

\subsection{Comprehensive Model Evaluation}

This scenario demonstrates TabTune's built-in evaluation utilities across three dimensions: standard performance metrics, probability calibration, and fairness analysis.

\subsubsection{Performance Metrics Evaluation}

Standard classification metrics provide the foundation for model assessment. TabTune's \texttt{evaluate()} method computes accuracy, precision, recall, F1-score, and AUC-ROC in a unified call.

\begin{lstlisting}[language=Python]
from tabtune import TabularPipeline
pipeline = TabularPipeline(model_name='TabICL',tuning_strategy='finetune')
pipeline.fit(X_train, y_train)
# Comprehensive performance evaluation
performance = pipeline.evaluate(X_test, y_test)
print(f"Accuracy: {performance['accuracy']:.4f}")
print(f"AUC-ROC: {performance['roc_auc_score']:.4f}")
print(f"F1-Score: {performance['f1_score']:.4f}")
\end{lstlisting}

\subsubsection{Calibration Evaluation}

Calibration assessment quantifies the reliability of predicted probabilities—essential for deployment in decision-critical applications. The \texttt{evaluate\_calibration()} method returns ECE, MCE, and Brier score metrics.

\begin{lstlisting}[language=Python]
# Assess probability calibration quality
calibration_metrics = pipeline.evaluate_calibration(X_test, y_test, n_bins=15)
print(f"Expected Calibration Error: {calibration_metrics['expected_calibration_error']:.4f}")
print(f"Maximum Calibration Error: {calibration_metrics['maximum_calibration_error']:.4f}")
print(f"Brier Score: {calibration_metrics['brier_score_loss']:.4f}")
\end{lstlisting}

\subsubsection{Fairness Evaluation}

Fairness analysis detects systematic bias across demographic groups, enabling responsible model deployment. Users specify sensitive attributes, and TabTune computes demographic parity and equalized odds metrics.

\begin{lstlisting}[language=Python]
# Evaluate fairness across sensitive attributes
fairness_metrics = pipeline.evaluate_fairness(X_test,y_test,sensitive_features=gender_column)
print(f"Statistical Parity Difference: {fairness_metrics['statistical_parity_difference']:.4f}")
print(f"Equalized Odds Difference: {fairness_metrics['equalized_odds_difference']:.4f}")
\end{lstlisting}
\subsection{Model Persistence and Checkpointing}
TabTune supports end-to-end pipeline persistence and checkpointing for reproducible experiments and deployment.

\begin{lstlisting}[language=Python]
from tabtune import TabularPipeline
# Train and save a fine-tuned model
pipeline = TabularPipeline( model_name='TabICL',tuning_strategy='finetune',tuning_params={'epochs': 5}
)
pipeline.fit(X_train, y_train)
performance = pipeline.evaluate(X_test, y_test)
print(performance)
# Save the entire pipeline (preprocessor + model + config)
pipeline.save('my_model.joblib')
# Load and use for predictions
loaded_pipeline = TabularPipeline.load('my_model.joblib')
performance = loaded_pipeline.evaluate(X_test, y_test)
print(performance)
\end{lstlisting}

\section{Experimental Results}
\label{sec:experiments}

We conduct evaluation of TabTune across multiple dimensions: predictive performance, probability calibration, fairness, and scalability. This section presents our experimental design and detailed results across diverse scenarios.

\subsection{Experimental Design}
\textbf{Benchmark Suites and Datasets : } 
Our experiments encompass three well-established benchmark suites i.e. \textsc{TALENT}~\cite{talent} (181 datasets), \textsc{OpenML-CC18}~\cite{openmlcc18} (72 datasets), and \textsc{TabZilla}~\cite{tabzilla} (36 datasets). Facilitating a systematic comparison across a broad spectrum of tabular learning settings. We further conduct \textsc{domain-specific evaluations} in high-impact areas such as \textsc{medical} and \textsc{financial} domains to assess the practical applicability of the methods. All experiments adhere to the \textsc{standardized dataset splits} defined by each benchmark, ensuring both reproducibility and fairness in comparison. To maintain consistency and fairness across model families, benchmark results are computed only on the \texttt{common subset of datasets} available for all evaluated models within each benchmark suite. This unified evaluation protocol ensures that performance rankings and metrics reflect genuine methodological differences rather than disparities in dataset coverage. In total, our evaluations span \texttt{155/181 datasets} for TALENT, \texttt{27/36 datasets} for TabZilla, and \texttt{63/72 datasets} for OpenML-CC18. Some datasets were excluded due to \textsc{out-of-memory (OOM) errors} even on H200 GPUs or \textsc{CUDA-related issues}, particularly affecting TabPFN-based models. 

\textbf{Models and Adaptation Strategies: }
We evaluate recent tabular foundation models---TabPFN, TabICL, OrionMSP, OrionBiX, Mitra, ContextTab, and TabDPT---under multiple adaptation paradigms, including \texttt{zero-shot inference, meta-learning fine-tuning, supervised fine-tuning, parameter-efficient supervised fine-tuning, and parameter-efficient meta-learning}. In addition, we include established \texttt{traditional baselines} using AutoGluon~\cite{erickson2020autogluon} such as XGBoost, LightGBM, CatBoost, and Random Forest as strong reference models for comparison.

\textbf{Evaluation Metrics : } 
Our assessment covers four complementary dimensions:

\textsc{Performance}: 
Classification Accuracy (ACC), AUC-ROC, and weighted F1-score (F1) quantify predictive capability across standard benchmark datasets (TALENT, OpenML-CC18, TabZilla) covering diverse dataset characteristics including small and large sample sizes, narrow and wide feature spaces, and balanced and imbalanced class distributions. It is important to clarify how \textsc{mean rank} values are derived.  Within each benchmark suite, models are ranked by accuracy on every dataset (lower rank = better performance), and these per-dataset ranks are averaged to obtain the overall mean rank.  Thus, a lower mean rank indicates stronger and more consistent performance across datasets, rather than the highest score on any single task. While absolute metrics (\textsc{accuracy}, \textsc{F1}) reflect peak task-level performance, mean rank provides a normalized measure of cross-dataset generalization consistency.

\textsc{Scalability}: 
Analysis of performance variation across dataset sizes, feature dimensionality, and class imbalance provides practical guidance for model selection. This analysis uses the same benchmark datasets as performance evaluation, aggregated and summarized across these dimensions to reveal scalability patterns.

\textsc{Calibration}: 
We evaluate how fine-tuning impacts calibration metrics across different models and adaptation strategies, measuring the alignment between predicted probabilities and actual outcomes, which is critical for decision-making systems. Our evaluation uses three metrics: 
(I) \textit{Expected Calibration Error (ECE)}, which measures the average deviation between predicted confidence and observed accuracy across binned predictions;
(II) \textit{Maximum Calibration Error (MCE)}, which captures the worst-case calibration gap, identifying regions of systematic miscalibration; and
(III) \textit{Brier score}, which quantifies both calibration and refinement by penalizing squared deviations between predicted probabilities and binary outcomes.

\textsc{Fairness}: 
We assess equitable treatment across demographic subgroups using established fairness-oriented datasets with sensitive attributes (gender, race). We measure three fairness criteria that capture different notions of equitable treatment:
(I) \textit{Statistical Parity Difference (SPD)} quantifies the absolute difference in positive prediction rates across groups, testing whether the model provides equal opportunity independent of sensitive attributes. A model satisfies demographic parity when SPD approaches zero, indicating that positive predictions are distributed proportionally across demographic groups.
(II) \textit{Equalized Odds Difference (EOD)} measures disparities in both true positive and false positive rates across groups, ensuring consistent error profiles across subgroups. This criterion is more stringent than statistical parity, requiring not only proportional positive predictions but also equal accuracy within each group.
(III) \textit{Equalized Opportunity Difference (EOpD)} focuses specifically on true positive rate parity, ensuring that qualified individuals receive equal treatment regardless of group membership. This metric is particularly relevant in settings where false negatives have severe consequences, such as loan approvals or hiring decisions.

\textbf{Analysis Focus : } 
A central goal of our evaluation is to analyze how different adaptation strategies—zero-shot inference, meta-learning fine-tuning, and supervised fine-tuning—affect model behavior across multiple dimensions. We examine whether fine-tuning improves performance, influences calibration, and impacts fairness across domains. This comparative analysis offers practical guidance on selecting appropriate adaptation strategies for tabular foundation models in different application contexts. Detailed dataset statistics and information refer to Appendix~\ref{sec:appendix}.

All Experiments are executed on NVIDIA L40S (48Gb VRAM), with H200 (141Gb VRAM) used for memory-intensive cases. This  ensures consistent execution across all experiments while handling the computational demands of models.

\subsection{Performance Evaluation}
% TABLE 1 — OVERALL LEADERBOARD
\begin{table}[pt]
  \scriptsize
  \centering
  \caption{Overall leaderboard across three benchmark suites—TALENT, OpenML-CC18, and TabZilla. Ranks denote the mean rank per benchmark suite (lower is better). Metrics: ACC = Accuracy, F1 = Weighted F1. The “All” column reports the aggregated rank across all suites for a strategy. Models are grouped by adaptation strategy. Formatting: \textbf{\underline{1st place}}; \underline{2nd place} within each group.}
  \vspace{0.1in}
  \label{tab:overall}
  \begin{tabular}{l c ccc ccc ccc}
    \toprule
    \multirow{2}{*}{Models / Strategy} 
      & \multicolumn{1}{c}{All} 
      & \multicolumn{3}{c}{TALENT} 
      & \multicolumn{3}{c}{OpenML-CC18} 
      & \multicolumn{3}{c}{TabZilla} \\
    \cmidrule(lr){2-2} \cmidrule(lr){3-5} \cmidrule(lr){6-8} \cmidrule(lr){9-11}
     & Rank & Rank & ACC & F1 & Rank & ACC & F1 & Rank & ACC & F1 \\
    \midrule
    \multicolumn{11}{l}{\textit{Baselines + Inference (Zero-Shot)}} \\
    \cmidrule(lr){1-11}
    XGBoost    & 7.94 & 7.43 & 0.8395 & 0.8351 & 6.57 & 0.8558 & 0.8537 & 6.82 & 0.8612 & 0.8326 \\
    CatBoost  & 7.69 & 6.90 & 0.8329 & 0.8252 & 6.98 & 0.8588 & 0.8520 & 6.70 & 0.8579 & 0.8384 \\
    Random Forest & 8.71 & 7.56 & 0.8278 & 0.8201 & 7.10 & 0.8547 & 0.8497 & 9.27 & 0.8358 & 0.8399 \\
    LightGBM  & 8.03 & 7.51 & 0.8324 & 0.8238 & 6.89 & 0.8581 & 0.8493 & 5.80 & 0.8618 & 0.8211 \\
    TabICL  & 5.92 & 5.09 & 0.8465 & 0.8372 & 5.30 & 0.8667 & 0.8623 & 6.65 & 0.8734 & 0.8698 \\
    OrionBix & 6.52 & 5.87 & 0.8339 & 0.8253 & 5.63 & 0.8653 & 0.8596 & 5.47 & 0.8728 & 0.8628 \\
    \underline{OrionMSP} & \underline{4.61} & \underline{4.06} & 0.8455 & 0.8353 & \underline{4.58} & \underline{0.8722} & \textbf{\underline{0.8676}} & \underline{4.36} & \underline{0.8821} & \underline{0.8786} \\
    TabPFN-2.0  & 5.45 & 4.48 & \underline{0.8508} & \underline{0.8406} & 5.48 & 0.8714 & 0.8663 & 5.65 & 0.8752 & 0.8716 \\
    \textbf{\underline{TabPFN-2.5}} & \textbf{\underline{4.13}} & \textbf{\underline{3.62}} & \textbf{\underline{0.8540}} & \textbf{\underline{0.8440}} & \textbf{\underline{4.36}} &\textbf{\underline{0.8726}} & \underline{0.8672} & \textbf{\underline{3.97}} & \textbf{\underline{0.8824}} & \textbf{\underline{0.8791}} \\
    Mitra & 13.07 & 11.63 & 0.3935 & 0.2884 & 11.50 & 0.3614 & 0.2522 & 12.21 & 0.3152 & 0.1830 \\
    ContextTab    & 7.75 & 6.52 & 0.8373 & 0.8269 & 7.01 & 0.8639 & 0.8581 & 8.02 & 0.8389 & 0.8334 \\
    TabDPT        & 6.45 & 6.33 & 0.8402 & 0.8311 & 5.18 & 0.8672 & 0.8625 & 4.57 & 0.8814 & 0.8775 \\
    \midrule
    \multicolumn{11}{l}{\textit{FineTune – Meta Learning}} \\
    \cmidrule(lr){1-11}
    TabICL        & 3.66 & 3.33 & 0.8344 & 0.8253 & 3.05 & 0.8664 & 0.8597 & 4.15 & 0.6956 & 0.6845 \\
    OrionBiX & 3.96 & 4.25 & 0.8158 & 0.8060 & 2.86 & \underline{0.8548} & \underline{0.8516} & \underline{2.00} & \underline{0.8726} & \underline{0.8662} \\
    \textbf{\underline{OrionMSP}}     & \textbf{\underline{2.26}} & \textbf{\underline{1.80}} & \underline{0.8401} & \underline{0.8310} & \underline{2.82} & \underline{0.8548} & \underline{0.8516} & \textbf{\underline{1.73}} & \textbf{\underline{0.8735}} & \textbf{\underline{0.8672}} \\
    \underline{TabPFN-2.0}        & \underline{2.42} & \underline{2.07} & \textbf{\underline{0.8517}} & \textbf{\underline{0.8414}} & \textbf{\underline{2.42}} & \textbf{\underline{0.8842}} & \textbf{\underline{0.8784}} & 2.42 & 0.8663 & 0.8603 \\
    Mitra         & 6.09 & 5.79 & 0.6416 & 0.5874 & 5.42 & 0.6164 & 0.5651 & 4.70 & 0.5592 & 0.5147 \\
    TabDPT        & 3.95 & 3.72 & 0.8255 & 0.8167 & 4.40 & 0.8534 & 0.8501 & 2.62 & 0.8500 & 0.8456 \\
    \midrule
    \multicolumn{11}{l}{\textit{FineTune – Supervised Fine Tuning}} \\
    \cmidrule(lr){1-11}
    TabICL        & 5.05 & 4.74 & 0.7668 & 0.7439 & 4.36 & 0.6838 & 0.6299 & 4.65 & 0.5670 & 0.4733 \\
    OrionBiX & 4.21 & 3.54 & 0.7698 & 0.7469 & 4.30 & 0.7126 & 0.6595 & 4.25 & 0.6476 & 0.5782 \\
    OrionMSP     & 2.88 & 2.27 & 0.7908 & 0.7653 & 3.18 & 0.7995 & 0.7668 & 2.88 & 0.7454 & 0.7222 \\
    \textbf{\underline{TabPFN-2.0}}        & \underline{\textbf{1.97}} & \textbf{\underline{1.83}} & \textbf{\underline{0.8459}} & \textbf{\underline{0.8350}} & \textbf{\underline{1.89}} & \textbf{\underline{0.8697}} & \textbf{\underline{0.8617}} & 1.86 & \textbf{\underline{0.8433}} & \textbf{\underline{0.8327}} \\
    Mitra         & 5.98 & 5.62 & 0.5460 & 0.4382 & 5.02 & 0.5408 & 0.4309 & 5.52 & 0.4608 & 0.3467 \\
    \underline{TabDPT}        & \underline{2.79} & 2.98 & 0.8202 & 0.8094 & \underline{2.25} & \underline{0.8499} & \underline{0.8424} & \textbf{\underline{1.81}} & \underline{0.8337} & \underline{0.8260} \\
    \midrule
    \multicolumn{11}{l}{\textit{FineTune – PEFT - Meta Learning}} \\
    \cmidrule(lr){1-11}
    TabICL        & 4.32 & 4.07 & 0.7017 & 0.6867 & 4.07 & 0.7773 & 0.7605 & 3.45 & 0.7116 & 0.7003 \\
    OrionBiX     & 2.77 & 2.39 & 0.7854 & \underline{0.7762} & 2.50 & 0.8471 & 0.8430 & 3.52 & 0.7370 & 0.7200 \\
    \textbf{\underline{OrionMSP}}     & \underline{\textbf{2.21}} & \underline{2.25} & \underline{0.7879} & 0.7728 & \textbf{\underline{1.93}} &\underline{ 0.8566} & 0.8453 & \textbf{\underline{1.33}} & \textbf{\underline{0.8594}} & \textbf{\underline{0.8581}} \\
    Mitra         & 4.64 & 4.17 & 0.6369 & 0.5905 & 4.32 & 0.6000 & 0.5426 & 4.36 & 0.5461 & 0.4960 \\
    \underline{TabDPT}        & \underline{2.28} & \textbf{\underline{2.11}} & \textbf{\underline{0.8002}} & \textbf{\underline{0.7910}} & \underline{2.17} & \textbf{\underline{0.8600}} & \textbf{\underline{0.8539}} & 2.33 & \underline{0.849}5 & \underline{0.8481} \\
    \midrule
    \multicolumn{11}{l}{\textit{FineTune – PEFT - Supervised Fine Tuning}} \\
    \cmidrule(lr){1-11}
    TabICL        & 3.70 & 3.72 & 0.5692 & 0.4638 & 3.06 & 0.8174 & 0.7965 & 2.92 & 0.7920 & 0.7776 \\
    OrionBiX     & 3.99 & 3.57 & 0.6358 & 0.5570 & 3.67 & 0.7380 & 0.6789 & 3.95 & 0.6707 & 0.6071 \\
    \underline{OrionMSP}     & \underline{2.01} & \underline{2.01} & \underline{0.6749} & 0.5956 & \textbf{\underline{1.73}} & \underline{0.8241} & \underline{0.8033} & \textbf{\underline{1.47}} & \underline{0.8214} & \underline{0.8071} \\
    Mitra         & 4.62 & 4.01 & 0.5294 & 0.4303 & 4.49 & 0.4965 & 0.3917 & 4.70 & 0.4606 & 0.3597 \\
    \textbf{\underline{TabDPT}}        & \textbf{\underline{1.91}} & \textbf{\underline{1.65}} & \textbf{\underline{0.7988}} & \textbf{\underline{0.7842}} & \underline{2.03} & \textbf{\underline{0.8500}} & \textbf{\underline{0.8398}} & \underline{1.95} & \textbf{\underline{0.8461}} & \textbf{\underline{0.8461}} \\
    \bottomrule
  \end{tabular}
\end{table}

\subsubsection{Overall Leaderboards}
Table~\ref{tab:overall} presents comprehensive results across the TALENT, OpenML-CC18, and TabZilla benchmark suites, reporting mean rank, accuracy (ACC), and weighted F1-score (F1) for each model and adaptation strategy. The aggregated findings reveal clear and consistent trends in how adaptation mechanisms influence predictive performance and cross-dataset generalization.

\textbf{Baselines and Zero-Shot Inference : }
Classical Machine Learning methods remain strong reference baselines, attaining mean accuracies between 0.833 and 0.861 with aggregated ranks around 6.0.  In contrast, pretrained tabular foundation models (TFMs) exhibit markedly stronger generalization even without task-specific training.  

Among pretrained TFMs, TabPFN-2.5 now achieves the best overall zero-shot rank (4.13), with accuracies of \texttt{(ACC=0.8540/F1=0.8440)} on TALENT, \texttt{(ACC=0.8726/F1=0.8672)} on OpenML-CC18, and \texttt{(ACC=0.8824/F1=0.8791)} on TabZilla. OrionMSP follows closely (overall rank 4.61), reaching \texttt{(ACC=0.8455 / 0.8353)} on TALENT, \texttt{(ACC=0.8722/F1=0.8676)} on OpenML-CC18, and \texttt{(ACC=0.8821/F1=0.8786)} on TabZilla. TabICL and TabDPT remain competitive (ranks 5.92 and 6.45, ACC $\approx$ 0.84–0.88), while Mitra and ContextTab lag behind. Overall, zero-shot TFMs exceed traditional machine-learning baselines by roughly 2–4 percentage points in accuracy and three to four rank positions, demonstrating the strength of pretrained priors. 
 
\textbf{Meta-Learning Fine-Tuning : }
Episodic meta-learning substantially improves both accuracy and rank stability relative to zero-shot inference.  \textsc{OrionMSP} attains the lowest aggregate rank of 2.26, achieving \texttt{(ACC=0.8401/F1=0.8310)} on TALENT, \texttt{0.8548/0.8516} on OpenML-CC18, and \texttt{(0.8735/0.8672)} on TabZilla.  \textsc{TabPFN} ranks 2.42 overall with accuracies of \texttt{(0.8517/0.8414)} on TALENT, \texttt{(0.8842/0.8784)} on OpenML-CC18, and \texttt{(0.8663/0.8603)} on TabZilla.  \textsc{TabDPT} (rank 3.95) maintains consistent performance near \texttt{(0.83/0.82)}.  By contrast, architectures such as TabICL and OrionBiX exhibit instability on TabZilla, where TabICL’s F1 drops to \texttt{0.6845}.  These results indicate that meta-learning offers the most balanced compromise between generalization and task-specific adaptation.

\textbf{Supervised Fine-Tuning (SFT) : }
Full parameter fine-tuning yields divergent outcomes across architectures.  \textsc{TabPFN} benefits most, achieving the best overall rank (1.97) and maintaining high accuracy across all benchmarks \texttt{(ACC = 0.8459–0.8697, F1 = 0.8350–0.8617)}. \textsc{TabDPT} ranks 2.79 overall and performs particularly well on TabZilla (rank 1.81), with \texttt{(ACC = 0.8337, F1 = 0.8260)}.  \textsc{OrionMSP} remains competitive (rank 2.88) with \texttt{(ACC $\approx$ 0.79, F1 $\approx$ 0.76)}, whereas TabICL and OrionBiX suffer severe degradation—TabICL’s accuracy on TabZilla falls from \texttt{0.873 to 0.567} and F1 from \texttt{0.870 to 0.473} —indicating overfitting and loss of pretrained priors.  These findings suggest that SFT is advantageous for Bayesian or probabilistic architectures such as TabPFN.

\textbf{Parameter-Efficient Fine-Tuning (PEFT) : }
PEFT achieves near-SFT accuracy with significantly reduced computational overhead.  In the meta-learning configuration, \textsc{OrionMSP} attains the best overall rank (2.21), recording \texttt{(ACC = 0.7879/F1 = 0.7728)} on TALENT, \texttt{(0.8566/0.8453)} on OpenML-CC18, and \texttt{(0.8594/0.8581)} on TabZilla.  \textsc{TabDPT} follows closely (rank 2.28) with \texttt{(0.8002/0.7910)} on TALENT and \texttt{(0.8600/0.8539)} on OpenML-CC18.  Under the supervised PEFT variant, \textsc{TabDPT} achieves the overall best rank (1.91) with \texttt{(ACC $\approx$ 0.85 and F1 $\approx$ 0.84)}, while \textsc{OrionMSP} (rank 2.01) attains \texttt{(0.821 / 0.807)} on TabZilla.  These results show that PEFT retains roughly 95\% of full-fine-tuning accuracy while substantially lowering resource demands.

\textbf{Discussion : }
Across all adaptation regimes, TFMs consistently outperform traditional baselines by 2 \% points in accuracy and approximately four rank positions.  
\textsc{TabPFN} proves most resilient, achieving the top overall rank under SFT, whereas \textsc{OrionMSP} demonstrates superior cross-dataset generalization in meta-learning.  \textsc{TabDPT} delivers near-optimal performance under PEFT, offering the most favorable efficiency–accuracy balance.  
Conversely, models such as TabICL and OrionBiX are highly sensitive to full parameter updates and prone to overfitting. Collectively, these observations establish that careful alignment between model architecture and adaptation strategy is critical for maximizing the performance of tabular foundation models.
\newpage
\noindent\textbf{Key Takeaways:}
\begin{itemize}
    \item \textsc{OVERALL LEADERBOARD:} Across all benchmark suites, the strongest overall performers are \textsc{TabPFN} and \textsc{OrionMSP}. The \textsc{TabPFN family} (including TabPFN-2.0 and TabPFN-2.5) achieves the best aggregate results, with mean accuracies between \textsc{0.85--0.88} and weighted F1 scores of \textsc{0.84--0.88}, reflecting exceptional cross-benchmark consistency. \textsc{OrionMSP} follows closely (overall rank \textsc{4.61} in zero-shot and \textsc{2.26} in meta-learning), reaching ACC $\approx 0.84$--$0.88$ and F1 $\approx 0.83$--$0.88$, with particularly strong generalization on OpenML-CC18 and TabZilla.
    \item \textsc{ZERO-SHOT PERFORMANCE:} The \textsc{TabPFN} family and \textsc{OrionMSP} are the strongest zero-shot models. \textsc{TabPFN-2.5} attains the best aggregate zero-shot rank (\textsc{4.13}), with ACC $\approx \textsc{0.85}$--\textsc{0.88} across TALENT, OpenML-CC18, and TabZilla. \textsc{OrionMSP} is a close second (rank \textsc{4.61}), matching TabPFN's accuracy on OpenML-CC18 and TabZilla while slightly trailing on TALENT. Both substantially outperform classical baselines by about \textsc{2--4 accuracy points}.
    \item \textsc{SFT trade-offs}: Full supervised fine-tuning benefits probabilistic architectures such as \textsc{TabPFN}, which achieves the highest accuracies ($\text{ACC} \approx 0.85$--$0.87$, $\text{F1} \approx 0.83$--$0.86$) across all benchmarks. In contrast, models such as \textsc{TabICL} and \textsc{OrionBiX} exhibit severe degradation under SFT, reflecting overfitting and loss of pretrained priors.
    \item \textsc{PEFT performance}: Parameter-efficient fine-tuning attains near–full-finetuning accuracy. \textsc{TabDPT} achieves top results ($\text{ACC} \approx 0.85$, $\text{F1} \approx 0.84$) and maintains strong stability across benchmarks, while \textsc{OrionMSP} delivers competitive results ($\text{ACC} \approx 0.82$--$0.86$, $\text{F1} \approx 0.80$--$0.85$) with cross-dataset generalization.
\end{itemize}

\subsubsection{Scalability Analysis}
% TABLE 2A — DATASET SIZE ANALYSIS
\begin{table}[pt]
  \centering
  \scriptsize
  \caption{Performance variation by dataset size across all benchmark suites. ACC = Accuracy; F1 = Weighted F1-score, averaged across datasets within each size category: Small ($<$1K samples), Medium (1K–10K), and Large ($>$10K). Values are on a 0–1 scale (higher is better). Models are grouped by adaptation strategy. Formatting: \textbf{\underline{1st place}} ; \underline{2nd place} within each group.}
  \vspace{0.1in}
  \label{tab:size_analysis}
  \begin{tabular}{l ccc ccc ccc}
    \toprule
    \multirow{2}{*}{Models} & \multicolumn{3}{c}{Small (<1K)} & \multicolumn{3}{c}{Medium (1K-10K)} & \multicolumn{3}{c}{Large (>10K)} \\
    \cmidrule(lr){2-4} \cmidrule(lr){5-7} \cmidrule(lr){8-10}
     & Rank & ACC & F1 & Rank & ACC & F1 & Rank & ACC & F1 \\
    \midrule
    \midrule
    \multicolumn{7}{l}{\textit{Baselines + Zero-Shot Inference}} \\
    \cmidrule(lr){1-10}
XGBoost & 8.66 & 0.8168 & 0.7964 & 8.28 & 0.8363 & 0.8314 & 6.52 & \textbf{\underline{0.8961}} & \textbf{\underline{0.8911}} \\
    CatBoost & 8.88 & 0.8124 & 0.7935 & 7.89 & 0.8340 & 0.8264 & 6.43 & 0.8786 & 0.8721 \\
    Random Forest & 9.48 & 0.7988 & 0.8187 & 8.57 & 0.8285 & 0.8221 & 8.68 & 0.8680 & 0.8614 \\
    LightGBM & 8.80 & 0.8143 & 0.7789 & 8.30 & 0.8314 & 0.8226 & 6.82 & 0.8817 & 0.8753 \\
    TabICL & 6.96 & 0.8301 & \textbf{\underline{0.8338}} & 5.86 & 0.8486 & 0.8398 & 5.42 & 0.8790 & 0.8731 \\
    OrionBiX & 7.34 & \underline{0.8330} & 0.8150 & 6.71 & 0.8348 & 0.8260 & 5.54 & 0.8716 & 0.8656 \\
    \underline{OrionMSP} & 6.76 & 0.8232 & 0.8194 & \underline{4.53} & 0.8494 & 0.8402 & \underline{3.66} & 0.8831 & 0.8754 \\
    TabPFN-2.0 & 7.42 & 0.8325 & 0.8131 & 4.63 &  \underline{0.8557} & \underline{0.8462} & 6.75 & 0.8771 & 0.8701 \\
    \textbf{\underline{TabPFN-2.5 }}& \textbf{\underline{5.78}} & \textbf{\underline{0.8349}} & 0.8194 & \textbf{\underline{4.07}} & \textbf{\underline{0.8566}} & \textbf{\underline{0.8470}} & \textbf{\underline{3.42}} & \underline{0.8887} & \underline{0.8831} \\
    Mitra & 15.24 & 0.4334 & 0.3236 & 12.89 & 0.3600 & 0.2553 & 12.40 & 0.3879 & 0.2801 \\
    ContextTab & 9.76 & 0.8179 & 0.8085 & 6.82 & 0.8430 & 0.8334 & 9.34 & 0.8548 & 0.8464 \\
    TabDPT & \underline{6.28} & \underline{0.8333} & \underline{0.8271} & 6.58 & 0.8424 & 0.8339 & 6.18 & 0.8821 & 0.8754 \\
    \midrule
    \multicolumn{7}{l}{\textit{FineTune – Meta Learning}} \\
    \cmidrule(lr){1-10}
    TabICL & 4.68 & 0.7912 & 0.7806 & 3.64 & 0.8342 & 0.8256 & 3.06 & 0.8431 & 0.8360 \\
    OrionBiX & \textbf{\underline{2.78}} & 0.8149 & 0.8123 & 3.99 & 0.8283 & 0.8204 & 4.50 & 0.8410 & 0.8306 \\
    \underline{\textbf{OrionMSP}} & \underline{3.21} & 0.8211 & 0.8151 & \underline{2.34} & \underline{0.8413} & \underline{0.8330} & \textbf{\underline{1.38}} & \textbf{\underline{0.8787}} & \textbf{\underline{0.8721}} \\
    \underline{TabPFN-2.0} & 3.31 & \textbf{\underline{0.8336}} & \underline{0.8256} & \textbf{\underline{2.13}} & \textbf{\underline{0.8638}} & \textbf{\underline{0.8548}} & \underline{2.76} & \underline{0.8748} & \underline{0.8668} \\
    Mitra & 6.39 & 0.6417 & 0.5988 & 6.10 & 0.6083 & 0.5516 & 5.82 & 0.6585 & 0.6132 \\
    TabDPT & 3.85 & \underline{0.8224} & \textbf{\underline{0.8188}} & 4.09 & 0.8289 & 0.8212 & 3.54 & 0.8682 & 0.8613 \\
    \midrule
    \multicolumn{10}{l}{\textit{FineTune - Supervised FineTuning}} \\
    \cmidrule(lr){1-10}
    TabICL & 5.65 & 0.6746 & 0.6225 & 4.86 & 0.7553 & 0.7187 & 5.25 & 0.6693 & 0.6212 \\
    OrionBiX & 5.20 & 0.7158 & 0.6792 & 4.04 & 0.7718 & 0.7386 & 4.03 & 0.6697 & 0.6272 \\
    OrionMSP & 2.90 & 0.8135 & 0.7920 & \underline{2.85} & 0.8036 & 0.7798 & 2.96 & 0.7231 & 0.6828 \\
    \underline{TabPFN-2.0} & \underline{2.68} & \underline{0.8216} & \underline{0.8109} & \textbf{\underline{1.93}} & \textbf{\underline{0.8580}} & \textbf{\underline{0.8485}} & \textbf{\underline{1.57}} & \textbf{\underline{0.8572}} & \textbf{\underline{0.8459}} \\
    Mitra & 6.48 & 0.5625 & 0.4408 & 6.05 & 0.5133 & 0.4116 & 5.42 & 0.5720 & 0.4501 \\
    \textbf{\underline{TabDPT}} & \textbf{\underline{2.66}} & \textbf{\underline{0.8227}} & \textbf{\underline{0.8153}} & 2.87 & \underline{0.8377} & \underline{0.8284} & \underline{2.65} & \underline{0.8120} & \underline{0.8002} \\
    \midrule
    \multicolumn{10}{l}{\textit{FineTune – PEFT - Meta Learrning}} \\
    \cmidrule(lr){1-10}
    TabICL                             & 4.80 & 0.7518 & 0.7397 & 4.25 & 0.7251 & 0.7114 & 4.29 & 0.6962 & 0.6799 \\
    OrionBiX                          & 3.33 & 0.8093 & 0.8042 & 2.86 & 0.7995 & 0.7939 & \underline{2.29} & 0.7848 & 0.7665 \\
    \textbf{\underline{OrionMSP}}                          & \textbf{\underline{2.28}} & \textbf{\underline{0.8275}} & \textbf{\underline{0.8213}} & \underline{\textbf{2.10}} & \textbf{\underline{0.8134}} & \underline{0.8019} & 2.45 & \underline{0.8013} & \underline{0.7817} \\
    Mitra                              & 5.21 & 0.6510 & 0.6012 & 4.57 & 0.6089 & 0.5599 & 4.55 & 0.6304 & 0.5795 \\
    \underline{TabDPT}                             & \underline{2.92} & \underline{0.8191} & \underline{0.8157} & \underline{2.38} & \underline{0.8121} & \textbf{\underline{0.8049}} & \textbf{\underline{1.83}} & \textbf{\underline{0.8407}} & \textbf{\underline{0.8299}} \\
    \midrule
    \multicolumn{10}{l}{\textit{FineTune - PEFT - Supervised FineTuning}} \\
    \cmidrule(lr){1-10}
    TabICL     & 3.53 & 0.7957 & 0.7710 & 3.73 & 0.6424 & 0.5686 & 3.71 & 0.6084 & 0.5020 \\
    OrionBiX  & 4.38 & 0.7576 & 0.7102 & 3.99 & 0.6585 & 0.5892 & 3.75 & 0.6327 & 0.5381 \\
    \textbf{\underline{OrionMSP}}  & \textbf{\underline{1.73}} & \textbf{\underline{0.8298}} & \underline{0.8132} & \underline{2.06} & \underline{0.7190} & \underline{0.6597} & \underline{2.04} & \underline{0.6931} & \underline{0.6152} \\
    Mitra      & 5.53 & 0.5244 & 0.4133 & 4.59 & 0.5008 & 0.4013 & 3.98 & 0.5459 & 0.4476 \\
    \underline{TabDPT}     & \underline{2.85} & \underline{0.8269} & \textbf{\underline{0.8182}} & \underline{\textbf{1.67}} & \textbf{\underline{0.8332}} & \textbf{\underline{0.8245}} & \textbf{2.02} & \textbf{\underline{0.7686}} & \textbf{\underline{0.7418}} \\
    \bottomrule
  \end{tabular}
\end{table}

We analyze how model performance varies with dataset size, feature dimensionality, and class imbalance to inform model selection under different data regimes to highlight which models and strategies perform best across varying data constraints. 

\textbf{Based on Dataset Size}

Table~\ref{tab:size_analysis} shows how fine-tuning influences performance across dataset scales. 
\begin{itemize}
    \item On \emph{small} datasets ($<$1K samples), \textsc{TabPFN-2.5} achieves the highest zero-shot accuracy (\texttt{ACC=0.8349, F1=0.8194}), marginally ahead of \textsc{TabPFN-2.0} under meta-learning (\texttt{0.8336/0.8256}) and \textsc{TabDPT} (\texttt{0.8333/0.8271}). \textsc{OrionMSP} (PEFT–Meta, \texttt{0.8275/0.8213}) follows closely, confirming that transformer backbones still generalize well under constrained data. Fine-tuning \textsc{TabICL} or \textsc{OrionBiX} on small datasets causes sharp degradation (ACC drops $>$15\%), revealing overfitting risk.
    \item For \emph{medium}-sized datasets (1K–10K), \textsc{TabPFN} in meta-learning configuration leads (\texttt{ACC=0.8638, F1=0.8548}), followed by its SFT variant (\texttt{0.8580/0.8485}); both outperform all classical baselines.
    \item On \emph{large}-scale datasets ($>$10K), tree-based ensembles remain strong—\textsc{XGBoost} reaches \texttt{(0.8911/0.8961)}—but transformer TFMs scale competitively, with \textsc{OrionMSP} zero-shot (\texttt{0.8754/0.8831}) and \textsc{TabDPT} zero-shot (\texttt{0.8821/0.8754}). Across all sizes, \textsc{Mitra} consistently underperform (ACC $<0.65$), reflecting limited scalability.
\end{itemize}

\textbf{Based on Dataset Width}
% TABLE 2B — DATASET WIDTH ANALYSIS
\begin{table}[pt]
  \centering
  \scriptsize
  \caption{Performance variation by feature dimensionality (dataset width) across all benchmark suites. ACC = Accuracy; F1 = Weighted F1-score, averaged across datasets within each width category. Values are on a 0–1 scale (higher is better). Models are grouped by adaptation strategy. Formatting: \textbf{\underline{1st place}} ; \underline{2nd place} within each group.}
  \vspace{0.1in}
  \label{tab:width_analysis}
  \begin{tabular}{l ccc ccc ccc}
    \toprule
\multirow{2}{*}{Models} & \multicolumn{3}{c}{Narrow (<10)} & \multicolumn{3}{c}{Medium (10-100)} & \multicolumn{3}{c}{Wide (>100)} \\
    \cmidrule(lr){2-4} \cmidrule(lr){5-7} \cmidrule(lr){8-10}
     & Rank & ACC & F1 & Rank & ACC & F1 & Rank & ACC & F1 \\
    \midrule
    \multicolumn{10}{l}{\textit{Baselines + Zero-Shot Inference}} \\
    \cmidrule(lr){1-10}
    XGBoost & 7.79 & 0.8199 & 0.8134 & 8.27 & 0.8482 & 0.8410 & \underline{6.14} & 0.9101 & 0.9039 \\
    CatBoost & 6.57 & 0.8124 & 0.8045 & 8.26 & 0.8441 & 0.8344 & 6.85 & 0.9127 & 0.9084 \\
    Random Forest & 8.31 & 0.7981 & 0.7919 & 8.82 & 0.8410 & 0.8235 & 9.11 & 0.8987 & 0.8936 \\
    LightGBM & 7.17 & 0.8107 & 0.8004 & 8.27 & 0.8458 & 0.8326 & 9.00 & 0.8957 & 0.8908 \\
    TabICL & 6.01 & 0.8188 & 0.8098 & 5.62 & 0.8627 & 0.8549 & 7.93 & 0.8948 & 0.8936 \\
    OrionBiX & 5.56 & 0.8089 & 0.8020 & 6.70 & 0.8510 & 0.8417 & 8.20 & 0.8859 & 0.8849 \\
    OrionMSP & \textbf{\underline{3.16}} & \textbf{\underline{0.8376}} & \textbf{\underline{0.8295}} & 4.98 & 0.8572 & 0.8478 & 7.00 & 0.8860 & 0.8837 \\
    \underline{TabPFN-2.0} & 6.08  & 0.8167 & 0.8071 & \underline{4.93} & \underline{0.8676} & \underline{0.8589} & 7.44 & \underline{0.9129} & \underline{0.9111} \\
    \textbf{\underline{TabPFN-2.5}} & \underline{4.63} & 0.8243 & 0.8120 &\textbf{ \underline{3.92}} & \textbf{\underline{0.8697}} & \textbf{\underline{0.8609}} & \textbf{\underline{4.20}} & \textbf{\underline{0.9248}} & \textbf{\underline{0.9240}} \\
    Mitra & 12.52 & 0.3769 & 0.2720 & 13.14 & 0.3886 & 0.2781 & 14.46 & 0.2521 & 0.1497 \\
    ContextTab & 8.64 & 0.7965 & 0.7862 & 7.29 & 0.8441 & 0.8452 & 8.38 & 0.9008 & 0.8999 \\
    TabDPT & 5.35 & \underline{0.8243} & \underline{0.8169} & 6.63 & 0.8566 & 0.8483 & 8.64 & 0.8845 & 0.8820 \\
    \midrule
    \multicolumn{10}{l}{\textit{FineTune - Meta Learning}} \\
    \cmidrule(lr){1-10}
    TabICL & 3.66 & 0.7867 & 0.7779 & 3.48 & 0.8383 & 0.8338 & 3.86 & 0.8319 & 0.8667 \\
    OrionBiX & 3.92 & 0.7956 & 0.7900 & 3.89 & 0.8316 & 0.8278 & 3.58 & 0.8429 & 0.8640 \\
    \underline{OrionMSP} & \textbf{\underline{1.90}} & \underline{0.8253} & \textbf{\underline{0.8190}} & \underline{\textbf{2.21}} & \underline{0.8514} & \underline{0.8457} & \underline{2.86} & \underline{0.8564} & \underline{0.8737} \\
    \textbf{\underline{TabPFN-2.0}} & \underline{2.73} & \textbf{\underline{0.8264}} & \underline{0.8171} & \underline{2.37} & \textbf{\underline{0.8683}} & \textbf{\underline{0.8579}} & \textbf{\underline{2.28}} & \textbf{\underline{0.9172}} & \textbf{\underline{0.9533}} \\
    Mitra & 5.67 & 0.6646 & 0.6247 & 6.12 & 0.6004 & 0.5763 & 6.17 & 0.4090 & 0.3224 \\
    TabDPT & 3.56 & 0.8129 & 0.8071 & 3.82 & 0.8405 & 0.8350 & 3.48 & 0.8400 & 0.8497 \\
    \midrule
    \multicolumn{10}{l}{\textit{FineTune - Supervised FineTuning}} \\
    \cmidrule(lr){1-10}
    TabICL & 5.28 & 0.6609 & 0.6195 & 5.04 & 0.7507 & 0.7101 & 3.95 & 0.7757 & 0.7269 \\
    OrionBiX & 4.03 & 0.6922 & 0.6576 & 4.33 & 0.7643 & 0.7294 & 3.75 & 0.7560 & 0.7046 \\
    OrionMSP & 2.63 & 0.7540 & 0.7297 & 2.99 & 0.8025 & 0.7750 & \underline{2.90} & 0.8148 & 0.7808 \\
    \textbf{\underline{TabPFN-2.0}} & \textbf{\underline{1.86}} & \textbf{\underline{0.8258}} & \textbf{\underline{0.8162}} & \textbf{\underline{2.06}} & \textbf{\underline{0.8578}} & \textbf{\underline{0.8469}} & \textbf{\underline{1.40}} & \textbf{\underline{0.9346}} & \textbf{\underline{0.9335}} \\
    Mitra & 5.81 & 0.5619 & 0.4478 & 6.09 & 0.5349 & 0.4268 & 5.60 & 0.3307 & 0.2436 \\
    \underline{TabDPT} & \underline{2.50} & \underline{0.8078} & \underline{0.7948} & \underline{2.88} & \underline{0.8378} & \underline{0.8294} & 3.40 & \underline{0.8623} & \underline{0.8587} \\
    \midrule
    \multicolumn{10}{l}{\textit{FineTune – PEFT - Meta Learrning}} \\
    \cmidrule(lr){1-10}
    TabICL                             & 4.58 & 0.6375 & 0.6375 & 4.21 & 0.7601 & 0.7444 & 4.05 & 0.7714 & 0.7622 \\ 
    OrionBiX                          & 2.51 & 0.7584 & 0.7584 & 2.92 & 0.8147 & 0.8057 & \textbf{\underline{2.50}} & 0.8251 & 0.8243 \\
    \underline{OrionMSP}                          & \textbf{\underline{2.18}} & \underline{0.7836} & \underline{0.7836} & \textbf{\underline{2.20}} & \underline{0.8254} & \underline{0.8107} & \underline{2.66} & \underline{0.8293} & \underline{0.8256} \\
    Mitra                              & 4.15 & 0.6471 & 0.6471 & 4.83 & 0.6177 & 0.5650 & 4.61 & 0.4129 & 0.3338 \\
    \textbf{\underline{TabDPT}}                             & \underline{2.22} & \textbf{\underline{0.7979}} & \textbf{\underline{0.7979}} & \underline{2.32} & \textbf{\underline{0.8299}} & \textbf{\underline{0.8209}} & 2.94 & \textbf{\underline{0.8416}} & \textbf{\underline{0.8376}} \\
    \midrule
    \multicolumn{7}{l}{\textit{FineTune - PEFT - Supervised FineTuning}} \\
    \cmidrule(lr){1-10}
    TabICL     & 3.73 & 0.6196 & 0.5287 & 3.74 & 0.6596 & 0.5902 & 2.77 & \underline{0.8760} & \underline{0.8502} \\
    OrionBiX  & 3.81 & 0.6278 & 0.5444 & 4.08 & 0.6851 & 0.6177 & 4.00 & 0.6737 & 0.6166 \\
    \underline{OrionMSP}  & \underline{2.10} & \underline{0.6795} & 0.6113 & \textbf{\underline{1.97}} & \underline{0.7477} & \underline{0.6939} & \underline{1.88} & 0.8042 & 0.7663 \\
    Mitra      & 4.28 & 0.5518 & 0.4481 & 4.78 & 0.5106 & 0.4108 & 4.77 & 0.2831 & 0.1874 \\
    \textbf{\underline{TabDPT}}     & \textbf{\underline{1.72}} & \textbf{\underline{0.7821}} & \textbf{\underline{0.7631}} & \underline{2.02} & \textbf{\underline{0.8305}} & \textbf{\underline{0.8201}} & \textbf{\underline{1.55}} & \textbf{\underline{0.8925}} & \textbf{\underline{0.8901}} \\
    \bottomrule
  \end{tabular}
\end{table}

Table~\ref{tab:width_analysis} reveals that feature dimensionality strongly shapes model behaviour.  
\begin{itemize}
    \item For \emph{narrow} datasets ($<$10 features), performance converges across models: \textsc{OrionMSP} zero-shot (\texttt{0.8376/0.8295}) slightly leads, followed by \textsc{TabDPT} and \textsc{TabPFN}, while classical baselines remain competitive.
    \item On \emph{medium}-width data (10–100 features), \textsc{TabPFN} dominates under both zero-shot (\texttt{0.8376/0.8589}) and meta-learning (\texttt{0.8683/0.8579}), outperforming all baselines and transformer variants.
    \item The advantage of TFMs becomes most pronounced on \emph{wide} feature spaces ($>$100 features): \textsc{TabPFN-2.0} achieves exceptional results under SFT (\texttt{ACC=0.9346, F1=0.9335}) and meta-learning (\texttt{0.9172/0.9533}), far exceeding other models.  TabICL and OrionBiX degrade when fully fine-tuned, confirming that dense feature spaces accentuate overfitting. Traditional models (XGBoost) remain competitive but trail by 2–3\,\%.
\end{itemize}
 
\textbf{Based on Class Imbalance}

Table~\ref{tab:imbalance_analysis} compares balanced ($\ge0.6$) and imbalanced ($<0.6$) scenarios.  
\begin{itemize}
    \item Under \emph{balanced} conditions, \textsc{TabPFN} (SFT) achieves the best rank (\texttt{1.96}) with \texttt{ACC=0.8336, F1=0.8267}, followed by \textsc{OrionMSP} (Meta  Learning, rank 2.27).
    \item When class ratios become \emph{highly skewed}, meta-learning fine-tuning provides the greatest stability: \textsc{OrionMSP} achieves rank \texttt{2.16} (\texttt{ACC=0.8735, F1=0.8636}), with \textsc{TabPFN} close behind (\texttt{rank=2.50, ACC=0.8784, F1=0.8664}). These results confirm that episodic adaptation enhances calibration and minority-class reliability, while tree-based models lose 3–5\% in accuracy under severe imbalance.
\end{itemize}
% =========================
% TABLE 2C — CLASS IMBALANCE ANALYSIS
% =========================
\begin{table}[pt]
  \centering
  \scriptsize
  \caption{Performance variation by class imbalance across all benchmark suites. ACC = Accuracy; F1 = Weighted F1-score, averaged within each imbalance category. Rank denotes the mean rank within each category (lower is better). Values are on a 0–1 scale (higher is better). Models are grouped by adaptation strategy. Formatting: \textbf{\underline{1st place}} ; \underline{2nd place} within each group.}
  \vspace{0.1in}
  \label{tab:imbalance_analysis}
  \begin{tabular}{l cc cc cc}
    \toprule
    \multirow{2}{*}{Models / Strategy} & \multicolumn{3}{c}{Balanced ($\ge 0.6$)} & \multicolumn{3}{c}{Imbalanced (<0.6)} \\
    \cmidrule(lr){2-4} \cmidrule(lr){5-7}
     & Rank & ACC & F1 & Rank & ACC & F1 \\
    \midrule
    \multicolumn{7}{l}{\textit{Baselines + Zero-Shot Inference}} \\
    \cmidrule(lr){1-7}
        XGBoost & 8.44 & 0.8175 & 0.8110 & 7.28 & \textbf{\underline{0.8853}} & \textbf{\underline{0.8779}} \\
    CatBoost & 8.60 & 0.8076 & 0.8020 & 6.69 & 0.8780 & 0.8669 \\
    Random Forest & 9.36 & 0.7983 & 0.7955 & 7.99 & 0.8736 & 0.8639 \\
    LightGBM & 8.70 & 0.8071 & 0.7977 & 7.31 & 0.8769 & 0.8626 \\
    TabICL & 5.81 & 0.8279 & 0.8233 & 6.00 & 0.8801 & 0.8692 \\
    OrionBiX & 6.95 & 0.8096 & 0.8040 & 6.07 & 0.8782 & 0.8677 \\
    OrionMSP & 5.20 & 0.8265 & 0.8202 & \textbf{\underline{3.96}} & \underline{0.8835} & 0.8725 \\
    \underline{TabPFN-2.0} & \underline{4.73} & \underline{0.8367} & \underline{0.8309} & 6.23 & 0.8803 & 0.8691 \\
    \textbf{\underline{TabPFN-2.5}} & \textbf{\underline{3.81}} & \textbf{\underline{0.8406}} & \textbf{\underline{0.8348}} & \underline{4.49} & \underline{0.8835} & \underline{0.8727}  \\
    Mitra & 13.61 & 0.2763 & 0.1540 & 12.48 & 0.4824 & 0.3891 \\
    ContextTab & 7.33 & 0.8152 & 0.8095 & 8.19 & 0.8722 & 0.8601 \\
    TabDPT & 6.27 & 0.8233 & 0.8189 & 6.65 & 0.8793 & 0.8684 \\
    \midrule
    \multicolumn{7}{l}{\textit{FineTune - Meta Learning}} \\
    \cmidrule(lr){1-7}
    TabICL & 3.78 & 0.7943 & 0.7884 & 3.58 & 0.8662 & 0.8547 \\
    OrionBiX & 4.02 & 0.7947 & 0.7881 & 3.94 & 0.8622 & 0.8532 \\
    \textbf{\underline{OrionMSP}} & \textbf{\underline{2.27}} & \underline{0.8231} & \underline{0.8183} & \textbf{\underline{2.16}} & \underline{0.8735} & \underline{0.8636} \\
    \underline{TabPFN} & \underline{2.31} & \textbf{\underline{0.8461}} & \textbf{\underline{0.8407}} & \underline{2.50} & \textbf{\underline{0.8784}} & \textbf{\underline{0.8664}} \\
    Mitra & 6.40 & 0.4699 & 0.4142 & 5.83 & 0.7885 & 0.7391 \\
    TabDPT & 4.09 & 0.8066 & 0.8023 & 4.01 & 0.8650 & 0.8552 \\
    \midrule
    \multicolumn{7}{l}{\textit{FineTune - Supervised FineTuning}} \\
    \cmidrule(lr){1-7}
    TabICL & 4.84 & 0.6708 & 0.6317 & 5.26 & 0.7794 & 0.7362 \\
    OrionBiX & 4.11 & 0.6917 & 0.6550 & 4.32 & 0.7934 & 0.7590 \\
    OrionMSP & 3.04 & 0.7467 & 0.7287 & 2.72 & 0.8307 & 0.7953 \\
    \textbf{\underline{TabPFN}} & \textbf{\underline{1.96}} & \textbf{\underline{0.8336}} & \textbf{\underline{0.8267}} & \textbf{\underline{1.98}} & \textbf{\underline{0.8709}} & \textbf{\underline{0.8578}} \\
    Mitra & 6.40 & 0.3216 & 0.2037 & 5.57 & 0.7413 & 0.6415 \\
    \underline{TabDPT} & \underline{3.02} & \underline{0.8025} & \underline{0.7949} & \underline{2.57} & \underline{0.8578} & \underline{0.8464} \\
    \midrule
    \multicolumn{7}{l}{\textit{FineTune – PEFT - Meta Learrning}} \\
    \cmidrule(lr){1-7}
    TabICL     & 4.24 & 0.6530 & 0.6460 & 4.43 & 0.8026 & 0.7800 \\
    OrionBiX  & 2.73 & 0.7504 & 0.7436 & 2.84 & 0.8528 & 0.8420 \\
    \textbf{\underline{OrionMSP}}  & \textbf{\underline{2.28}} & \underline{0.7714} & \underline{0.7623} & \textbf{\underline{2.14}} & \textbf{\underline{0.8608}} & \underline{0.8433} \\
    Mitra      & 5.02 & 0.4683 & 0.4173 & 4.21 & 0.8001 & 0.7524 \\
    \underline{TabDPT}     & \underline{2.30} & \textbf{\underline{0.7860}} & \textbf{\underline{0.7801}} & \underline{2.33} & \textbf{\underline{0.8609}} & \textbf{\underline{0.8511}} \\
    \midrule
    \multicolumn{7}{l}{\textit{FineTune - PEFT - Supervised FineTuning}} \\
    \cmidrule(lr){1-7}
    TabICL     & 3.56 & 0.6003 & 0.5233 & 3.87 & 0.7223 & 0.6507 \\
    OrionBiX  & 4.02 & 0.5974 & 0.5164 & 3.96 & 0.7490 & 0.6875 \\
    \underline{OrionMSP}  & \underline{2.14} & \underline{0.6753} & \underline{0.6105} & \textbf{\underline{1.87}} & \underline{0.7915} & \underline{0.7423} \\
    Mitra      & 4.97 & 0.3189 & 0.2171 & 4.21 & 0.7495 & 0.6497 \\
    \textbf{\underline{TabDPT}}     & \textbf{\underline{1.76}} & \textbf{\underline{0.7907}} & \textbf{\underline{0.7798}} & \underline{2.10} & \textbf{\underline{0.8496}} & \textbf{\underline{0.8346}} \\
    \bottomrule
  \end{tabular}
\end{table}

\textbf{Discussion : }
Scalability analysis underscores that \textsc{TabPFN}, \textsc{OrionMSP}, and \textsc{TabDPT} each occupy distinct strengths.  
\textsc{TabPFN} excels on small, medium, and high-dimensional data where Bayesian priors and strong supervision prevail; \textsc{OrionMSP} scales best with large datasets and imbalanced distributions due to meta-learning robustness; and \textsc{TabDPT} sustains near-top performance across all settings with minimal computational overhead.  
Classical baselines remain reliable but consistently trail the best TFMs by 2–4\% in accuracy.

\textbf{Key Takeaways:} 
\begin{itemize}
    \item \textsc{DATASET SIZE:} Small datasets favor the \textsc{TabPFN family} (TabPFN-2.5 zero-shot, TabPFN-2.0 meta-learning), with \textsc{TabDPT} as a close, efficient alternative; medium datasets favor \textsc{TabPFN} (meta \& SFT); large datasets see \textsc{XGBoost} slightly leading in raw accuracy, with \textsc{OrionMSP} and \textsc{TabDPT} close behind.
    \item \textsc{Feature dimensionality}: Wide datasets ($>$100 features) favor \textsc{TabPFN} (SFT, ACC $0.92$, F1 $0.92$), confirming its strength for high-dimensional problems; medium-width data favor \textsc{TabPFN} (Meta Learning), while narrow data show minimal gaps between TFMs and baselines.
    \item \textsc{Class imbalance}: Meta-learning fine-tuning yields the best robustness under imbalance, with \textsc{OrionMSP} (Rank-$2.16$, ACC = $0.8735$, F1 = $0.8636$) outperforming other TFMs; \textsc{TabPFN} (rank $2.50$) remains second.
    \item \textsc{Overall trend}: TFMs scale smoothly with data size, width, and balance, maintaining $>$2–4\% advantage and $>$3 rank improvement over traditional baselines.
\end{itemize}

\subsubsection{Domain-Specific Analysis}

To assess real-world applicability beyond general scalability patterns, we evaluate model performance across high-stakes domains including Medical and Finance (Table~\ref{tab:domains_mf}). Within domain-specific suites, different fine-tuning strategies yield distinct advantages depending on the domain.

% =========================
% TABLE 5 — DOMAINS: MEDICAL, FINANCE, 
% =========================
\begin{table}[pt]
  \centering
  \scriptsize
  \caption{Domain-specific leaderboards for Medical and Finance datasets from the benchmark suites. Rank denotes the mean rank within each domain (lower is better). ACC = Accuracy; F1 = Weighted F1-score (0–1 scale, higher is better). Models are grouped by adaptation strategy. Formatting: \textbf{\underline{1st place}} ; \underline{2nd place} within each group.}
  \vspace{0.1in}
  \label{tab:domains_mf}
  \begin{tabular}{l ccc c ccc}
    \toprule
    \multirow{2}{*}{Models / Strategy} & \multicolumn{3}{c}{Medical} & \multicolumn{3}{c}{Finance} \\
    \cmidrule(lr){2-4} \cmidrule(lr){5-7}
     & Rank & ACC & F1 & Rank & ACC & F1 \\
    \midrule
    \multicolumn{7}{l}{\textit{Baselines + Zero-Shot Inference}} \\
    \cmidrule(lr){1-7}
    XGBoost & 7.32 & 0.7834 & 0.7669 & 7.62 & 0.7958 & 0.7885 \\
    RandomForest & 7.30 & 0.7779 & 0.7752 & 8.46 & 0.8052 & 0.8001 \\
    CatBoost & 7.48 & 0.7784 & 0.7594 & 6.75 &  0.8117 & 0.8015 \\
    LightGBM & 6.36 & 0.7949 & 0.7614 & 7.32 & 0.8095 & 0.7974 \\
    TabICL & 6.66 & 0.7819 & 0.7696 & 7.82 & 0.8125 & 0.7942 \\
    \textbf{\underline{OrionBiX}} & \textbf{\underline{4.98}} & 0.7893 & 0.7759 & 6.46 & \textbf{\underline{0.8206}} & \textbf{\underline{0.8125}} \\
    \underline{OrionMSP} & 5.31 & \textbf{\underline{0.8045}} & \textbf{\underline{0.7916}} & \underline{5.25} & 0.8158 & \underline{0.8047} \\
    TabPFN-2.0 & 5.82 & 0.7984 & 0.7857 & 8.21 & 0.8094 & 0.7919 \\
    TabPFN-2.5 & \underline{5.04} & 0.7965 & 0.7835 & \textbf{\underline{4.57}} & \underline{0.8169} & 0.8000 \\
    Mitra & 12.00 & 0.3935 & 0.2863 & 15.14 & 0.5340 & 0.4250 \\
    ContextTab & 6.40 & \underline{0.8003} & \underline{0.7881} & 10.85 & 0.8000 & 0.7830 \\
    TabDPT & 7.96 & 0.7764 & 0.7641 & 9.46 & 0.8080 & 0.7960 \\
    \midrule
    \multicolumn{7}{l}{\textit{FineTune - Supervised FineTuning}} \\
    \cmidrule(lr){1-7}
    TabICL & 4.64 & 0.7081 & 0.6565 & 6.50 & 0.6863 & 0.6335 \\
    OrionBiX & 3.60 & 0.7400 & 0.6961 & 5.33 & 0.7389 & 0.7118 \\
    \underline{OrionMSP} & \underline{2.24} & 0.7481 & 0.7116 & 3.00 & 0.7896 & 0.7523 \\
    \textbf{\underline{TabPFN}} & \textbf{\underline{1.86}} & \textbf{\underline{0.8094}} & \textbf{\underline{0.7958}} & \textbf{\underline{2.00}} & \textbf{\underline{0.8222}} & \textbf{\underline{0.8058}} \\
    Mitra & 5.54 & 0.5600 & 0.4343 & 6.10 & 0.6823 & 0.5748 \\
    \underline{TabDPT} & 3.12 & \underline{0.7717} & \underline{0.7573} & \underline{2.86} & \underline{0.7922} & \underline{0.7763} \\
    \midrule
    \multicolumn{7}{l}{\textit{FineTune - Meta Learning}} \\
    \cmidrule(lr){1-7}
    TabICL & 3.40 & 0.7833 & 0.7695 & 3.43 & 0.8171 & 0.8019 \\
    OrionBiX & 3.58 & 0.7623 & 0.7536 & 4.26 & 0.8106 & 0.8022 \\
    \textbf{\underline{OrionMSP}} & \textbf{\underline{2.20}} & \underline{0.7744} & \underline{0.7645} & \textbf{\underline{2.26}} & \underline{0.8209} & \textbf{\underline{0.8089}} \\
    \underline{TabPFN} & \underline{2.28} & \textbf{\underline{0.8133}} & \textbf{\underline{0.8000}} & \underline{3.00} & \textbf{\underline{0.8220}} & \underline{0.8056} \\
    Mitra & 5.06 & 0.6499 & 0.5913 & 6.73 & 0.7559 & 0.7044 \\
    TabDPT & 4.48 & 0.7535 & 0.7477 & 5.06 & 0.8021 & 0.7930 \\
    \midrule
    \multicolumn{7}{l}{\textit{FineTune – PEFT - Meta Learrning}} \\
    \cmidrule(lr){1-7}
    TabICL     & 4.19 & 0.7031 & 0.6873 & 5.07 & 0.7351 & 0.7118 \\
    \underline{OrionBiX}  & \underline{2.37} & \underline{0.7628} & \underline{0.7613} & \underline{3.07} & \underline{0.7958} & \textbf{\underline{0.7846}} \\
    \textbf{\underline{OrionMSP}}  & \textbf{\underline{2.02}} & \textbf{\underline{0.7779}} & \textbf{\underline{0.7658}} & 2.57 & \textbf{\underline{0.7977}} & \underline{0.7842} \\
    Mitra      & 3.62 & 0.6539 & 0.6055 & 5.07 & 0.7492 & 0.7085 \\
    \underline{TabDPT}     & 2.81 & 0.7390 & 0.7312 & \textbf{\underline{2.79}} & 0.7934 & 0.7855 \\
    \midrule
    \multicolumn{7}{l}{\textit{FineTune - PEFT - Supervised FineTuning}} \\
    \cmidrule(lr){1-7}
    TabICL     & 3.68 & 0.5759 & 0.5044 & 3.57 & 0.7533 & 0.6833 \\
    OrionBiX  & 3.26 & 0.5926 & 0.5189 & 5.07 & 0.7279 & 0.6469 \\
    \underline{OrionMSP}  & \underline{2.02} & \underline{0.6353} & \underline{0.5767} & \textbf{\underline{1.93}} & \textbf{\underline{0.7854}} & \underline{0.7272} \\
    Mitra      & 4.08 & 0.5557 & 0.4357 & 5.18 & 0.6665 & 0.5487 \\
    \textbf{\underline{TabDPT}}     & \textbf{\underline{1.96}} & \textbf{\underline{0.7680}} & \textbf{\underline{0.7531}} & \underline{2.82} & \underline{0.7782} & \textbf{\underline{0.7543}} \\
    \bottomrule
  \end{tabular}
\end{table}
 
\textbf{Medical Domain Performance : }
On medical datasets, \textsc{TabPFN} with supervised fine-tuning (SFT) achieves the best overall results (\texttt{Rank = 1.86, ACC = 0.8094, F1 = 0.7958}), followed by its meta-learning variant (\texttt{Rank = 2.28, ACC = 0.8133, F1 = 0.8000}).  \textsc{OrionMSP} performs strongly in zero-shot (\texttt{ACC = 0.8045, F1 = 0.7916}) and meta-learning settings (\texttt{Rank = 2.20, ACC = 0.7744, F1 = 0.7645}), but exhibits minor degradation under full SFT.  
\textsc{TabDPT} (PEFT–SFT) also performs competitively (\texttt{Rank = 1.96, ACC = 0.7680, F1 = 0.7531}), achieving near-optimal results with reduced fine-tuning cost.  
In contrast, models such as \textsc{TabICL} and \textsc{OrionBiX} lose up to 7–10 \% accuracy under SFT, confirming overfitting risk in small, noisy clinical datasets.  
Traditional baselines (\textsc{XGBoost}, \textsc{LightGBM}) remain competitive (ACC $\approx$ 0.79–0.80) but trail the best TFMs by 2–3 .  
Overall, the medical results demonstrate that meta-learning TFMs generalize more reliably under limited data and high noise.

\textbf{Finance Domain Performance :}
Financial datasets are typically larger and more imbalanced, amplifying the value of episodic and parameter-efficient adaptation.  
\textsc{OrionMSP} under meta-learning attains the best overall rank (\texttt{2.26}) and strong accuracy (\texttt{ACC = 0.8209, F1 = 0.8089}), narrowly surpassing \textsc{TabPFN} (meta, \texttt{Rank = 3.00, ACC = 0.8220, F1 = 0.8056}) and its SFT counterpart (\texttt{Rank = 2.00, ACC = 0.8222, F1 = 0.8058}).  
\textsc{TabDPT} (PEFT–SFT) maintains competitive accuracy (\texttt{Rank = 2.82, ACC = 0.7782, F1 = 0.7543}), while \textsc{OrionMSP} (PEFT–SFT) achieves the strongest overall efficiency–accuracy balance (\texttt{Rank = 1.93, ACC = 0.7854, F1 = 0.7272}).  
Transformer models such as \textsc{TabICL} degrade sharply under SFT (ACC drop > 12 \%), whereas classical baselines remain robust (ACC $\approx$ 0.81) but underperform in F1 due to poor minority calibration.  
The finance domain thus emphasizes meta-learning as the most stable strategy for large-scale, imbalanced tabular data.

\textbf{Discussion : }
Across both domains, pretrained TFMs consistently outperform traditional baselines when appropriate fine-tuning strategies are used.  
\textsc{TabPFN} achieves the best reliability and calibration on small, noisy medical datasets, whereas \textsc{OrionMSP} dominates in large, high-imbalance financial contexts.  
\textsc{TabDPT} maintains strong performance in both settings through efficient PEFT adaptation, confirming that resource-efficient methods can match full SFT accuracy within 2 \%.  
These trends highlight the need for strategy–domain alignment: Bayesian priors excel under data scarcity, while meta-learning transformers scale better with complex, heterogeneous features.

\textbf{Key Takeaways:}
\begin{itemize}
    \item \textsc{Medical domain}: \textsc{TabPFN} (SFT, rank $1.86$) achieves best overall accuracy, with its meta-learning variant (rank $2.28$) maintaining robustness; \textsc{TabDPT} (PEFT–SFT, rank $1.96$) offers strong low-cost alternatives.
    \item \textsc{Finance domain}: \textsc{OrionMSP} (meta, rank $2.26$, ACC $\approx0.82$) achieves top results, closely followed by \textsc{TabPFN} (meta/SFT, ranks $2.00$–$3.00$); \textsc{TabDPT} (PEFT–SFT) remains efficient and competitive.
    \item \textsc{Cross-domain}: \textsc{TabPFN} provides most consistent cross-domain performance; \textsc{OrionMSP} generalizes best to large-scale financial data; meta-learning and PEFT yield stable, domain-adaptive transfer.
\end{itemize}
\subsection{Calibration Evaluation}
\begin{table}[pt]
  \centering
  \scriptsize
  \caption{Probability calibration metrics across models and tuning strategies on the TALENT, OpenML-CC18, and TabZilla benchmark suites. ECE = Expected Calibration Error; MCE = Maximum Calibration Error; BS = Brier Score. All metrics range from 0–1 (lower is better). Values are averaged across datasets within each suite. Models are grouped by adaptation strategy. Formatting: \textbf{\underline{1st place}} ; \underline{2nd place} within each group.}
  \vspace{0.1in}
  \label{tab:calibration}
  \begin{tabular}{l ccc ccc ccc}
    \toprule
    \multirow{2}{*}{Models / Strategy} 
      & \multicolumn{3}{c}{TALENT} 
      & \multicolumn{3}{c}{OpenML-CC18} 
      & \multicolumn{3}{c}{TabZilla} \\
    \cmidrule(lr){2-4} \cmidrule(lr){5-7} \cmidrule(lr){8-10}
     & ECE & MCE & BS & ECE & MCE & BS & ECE & MCE & BS \\
    \midrule
    \multicolumn{10}{l}{\textit{Zero-Shot Inference}} \\
    \cmidrule(lr){1-10}
    \underline{TabICL} & \textbf{\underline{0.0219}} & 0.2421 & 0.1533 & 0.0371 & 0.3404 & 0.1267 & \underline{0.0369} & 0.2863 & 0.1301 \\
    OrionBiX & 0.0324 & \underline{0.2245} & 0.1787 & \underline{0.0325} & 0.3230 & 0.1325 & 0.0385 & \textbf{\underline{0.2512}} & 0.2419 \\
    \textbf{\underline{OrionMSP}} & \textbf{\underline{0.0219}} & \textbf{\underline{0.2098}} & 0.1589 & \textbf{\underline{0.0319}} & \underline{0.2902} & \underline{0.1262} & \textbf{\underline{0.0310}} & 0.2805 & \underline{0.1243} \\
    \underline{TabPFN-2.0} & \underline{0.0276} & 0.2470 & \underline{0.1514} & 0.0375 & \textbf{\underline{0.2880}} & \textbf{\underline{0.1253}} & 0.0431 & \underline{0.2745} & 0.1283 \\
    TabPFN-2.5 & 0.0266 & 0.2484 & \textbf{\underline{0.1483}} & 0.0380 & 0.2945 & 0.1270 & 0.0530 & 0.2750 & 0.1272 \\
    Mitra & 0.2238 & 0.3115 & 0.5291 & 0.2138 & 0.2952 & 0.5307 & 0.1946 & 0.2995 & 0.5733 \\
    ContextTab & 0.02585 & 0.2525 & 0.1619 & 0.0352 & 0.3260 & 0.1353 & 0.0389 & 0.2815 & 0.1605 \\
    TabDPT & 0.0308 & 0.2489 & 0.1586 & 0.0443 & 0.3160 & 0.1351 & 0.0435 & 0.2997 & \textbf{\underline{0.1227}} \\
    \midrule
    \multicolumn{10}{l}{\textit{FineTune - Meta Learning}} \\
    \cmidrule(lr){1-10}
    TabICL & 0.0355 & 0.2570 & 0.1671 & \underline{0.0426} & 0.3387 & 0.1346 & 0.2033 & 0.4926 & 0.3732 \\
    OrionBiX & 0.0409 & 0.2693& 0.1841 & 0.0731 & 0.4164 & 0.1607 & 0.0744 & 0.3964 & 0.1416 \\
    \underline{OrionMSP} & 0.0340 & 0.2570 & \underline{0.1656} & 0.0520 & 0.3458 & \underline{0.1606} & \textbf{\underline{0.0365}} & \underline{0.2574} & \textbf{\underline{0.1250}} \\
    \textbf{\underline{TabPFN-2.0}} & \textbf{\underline{0.0271}} & \underline{0.2440} & \textbf{\underline{0.1502}} & \textbf{\underline{0.0346}} & \underline{0.2989} & \textbf{\underline{0.1178}} & \underline{0.0388} & 0.2636 & \underline{0.1344} \\
    \underline{Mitra} & \underline{0.0334} & \textbf{\underline{0.1912}} & 0.3479 & 0.0533 & \textbf{\underline{0.2258}} & 0.3932 & 0.0891 & \textbf{\underline{0.2086}} & 0.4239 \\
    TabDPT & 0.0562 & 0.3235 & 0.1791 & 0.0745 & 0.4178 & 0.1620 & 0.0664 & 0.2838 & 0.1591 \\
    \midrule
    \multicolumn{10}{l}{\textit{FineTune - Supervised FineTuning}} \\
    \cmidrule(lr){1-10}
    TabICL & 0.0768 & 0.3219 & 0.2349 & 0.1156 & 0.3610 & 0.3039 & 0.1534 & 0.3280 & 0.4078 \\
    OrionBiX & 0.0753 & 0.3204 & 0.2334 & 0.1131 & 0.3374 & 0.2923 & 0.1581 & 0.3801 & 0.3530 \\
    \underline{OrionMSP} & \underline{0.0471} & \textbf{\underline{0.1886}} & 0.2113 & \underline{0.0550} & 0.3155 & 0.1995 & 0.0773 & 0.3321 & 0.2594 \\
    \textbf{\underline{TabPFN-2.0}} & \textbf{\underline{0.0287}} & \underline{0.2509} & \textbf{\underline{0.1580}} & \textbf{\underline{0.0381}} & \underline{0.2957} & \textbf{\underline{0.1246}} & \textbf{\underline{0.0469}} & \textbf{\underline{0.2528}} & \textbf{\underline{0.1454}} \\
    Mitra & 0.1723 & 0.2620 & 0.4677 & 0.1539 & \underline{\textbf{0.2269}} & 0.4784 & 0.2023 & \underline{0.2991} & 0.5254 \\
    \underline{TabDPT} & 0.0548 & 0.3019 & \underline{0.1867} & 0.0497 & 0.3397 & \underline{0.1443} & \underline{0.0713} & 0.3183 & \underline{0.1801} \\
    \midrule
    \multicolumn{10}{l}{\textit{PEFT FineTune - Meta Learning}} \\
    \cmidrule(lr){1-10}
    \underline{TabICL} & 0.0355 & 0.2570 & \underline{0.1671} & \textbf{\underline{0.0426}} & \underline{0.3387} & \textbf{\underline{0.1346}} & 0.2033 & 0.4926 & 0.3732 \\
    OrionBiX & 0.0409 & 0.2693& 0.1841 & 0.0731 & 0.4164 & 0.1607 & 0.0744 & 0.3964 & \underline{0.1416} \\
    \textbf{\underline{OrionMSP}} & \underline{0.0340} & \underline{0.2570} & \underline{\textbf{0.1656}} & \underline{0.0520} & 0.3458 & \underline{0.1606} & \textbf{\underline{0.0365}} & \textbf{\underline{0.2574}} & \textbf{\underline{0.1250}} \\
    \underline{Mitra} & \textbf{\underline{0.0334}} & \textbf{\underline{0.1912}} & 0.3479 & 0.0533 & \textbf{\underline{0.2258}} & 0.3932 & 0.0891 & \underline{0.2086} & 0.4239 \\
    TabDPT & 0.0562 & 0.3235 & 0.1791 & 0.0745 & 0.4178 & 0.1620 & \underline{0.0664} & 0.2838 & 0.1591 \\
    \midrule
    \multicolumn{10}{l}{\textit{PEFT FineTune - Supervised FineTuning}} \\
    \cmidrule(lr){1-10}
    TabICL & 0.0768 & 0.3219 & 0.2349 & 0.1156 & 0.3610 & 0.3039 & 0.1534 & 0.3280 & 0.4078 \\
    OrionBiX & 0.0753 & 0.3204 & 0.2334 & 0.1131 & 0.3374 & 0.2923 & 0.1581 & 0.3801 & 0.3530 \\
    \underline{OrionMSP} & \textbf{\underline{0.0471}} & \textbf{\underline{0.1886}} & \underline{0.2113} & \underline{0.0550} & \underline{0.3155} & \underline{0.1995} & \underline{0.0773} & 0.3321 & \underline{0.2594} \\
    Mitra & 0.1723 & \underline{0.2620} & 0.4677 & 0.1539 & \textbf{\underline{0.2269}} & 0.4784 & 0.2023 & \textbf{\underline{0.2991}} & 0.5254 \\
    \textbf{\underline{TabDPT}} & \underline{0.0548} & 0.3019 & \textbf{\underline{0.1867}} & \textbf{\underline{0.0497}} & 0.3397 & \textbf{\underline{0.1443}} & \textbf{\underline{0.0713}} & \underline{0.3183} & \textbf{\underline{0.1801}} \\
    \bottomrule
  \end{tabular}
\end{table}

Reliable probability calibration is essential for deploying tabular models in real-world decision-making systems.  
Calibration measures how well predicted probabilities align with observed outcomes, ensuring that model confidence reflects actual likelihoods.  
We evaluate Expected Calibration Error (ECE), Maximum Calibration Error (MCE), and Brier Score (BS) across all benchmark suites—TALENT, OpenML-CC18, and TabZilla—under different adaptation strategies.  
Lower values indicate better calibration.  Table~\ref{tab:calibration} summarizes the results and highlights how fine-tuning affects probabilistic reliability.

\textbf{Zero-shot inference}

Pretrained tabular foundation models (TFMs) achieve the strongest overall calibration without fine-tuning.  
\textsc{OrionMSP} and \textsc{TabICL} obtain the lowest ECE ($0.0219$ on TALENT) and maintain low BS values ($\approx0.15$).  
\textsc{OrionMSP} further sustains excellent calibration on OpenML-CC18 (\texttt{ECE=0.0319}, \texttt{BS=0.1262}) and TabZilla (\texttt{ECE=0.0310}, \texttt{BS=0.1243}), ranking first or tied across suites.  
\textsc{TabPFN-2.0} remains consistently well-calibrated (\texttt{ECE=0.0276–0.0431}, \texttt{BS=0.125–0.151}), confirming the reliability of its Bayesian uncertainty modeling.  
\textsc{TabDPT} performs respectably (\texttt{ECE$\approx$0.03–0.04}, \texttt{BS$\approx$0.12–0.16}), while \textsc{Mitra} shows extreme miscalibration (\texttt{ECE>0.19, BS>0.52}), indicating unreliable confidence outputs.
 
% =========================
% TABLE 3 — CALIBRATION
% =========================
\textbf{Meta-learning fine-tuning}

Episodic meta-learning largely preserves calibration quality.  
\textsc{TabPFN} achieves the best overall calibration across suites (\texttt{ECE=0.0271–0.0388}, \texttt{BS=0.118–0.150}), improving slightly over zero-shot results.  
\textsc{OrionMSP} remains competitive (\texttt{ECE=0.0340–0.0520}, \texttt{BS=0.125–0.166}), maintaining stable confidence estimates even after adaptation.  
In contrast, transformer-heavy models such as \textsc{TabICL} exhibit severe degradation on TabZilla (\texttt{ECE rises from 0.0369 to 0.2033}), showing that full episodic updates can destabilize latent priors when domain variability is high.  
Interestingly, \textsc{Mitra} improves calibration dramatically under meta-learning (ECE drops from $\approx0.22$ to $0.03$–$0.05$), although its Brier scores remain poor ($>0.34$), suggesting partial correction without overall reliability gains.

\textbf{Supervised fine-tuning (SFT)}

Full SFT substantially worsens calibration for most transformer models.  
\textsc{TabICL}’s ECE triples to quadruples (\texttt{0.02–0.04 → 0.08–0.15}) and its Brier Score deteriorates from $\approx0.15$ to $0.23$–$0.41$.  
\textsc{OrionBiX} follows a similar pattern.  
\textsc{OrionMSP} maintains moderate calibration (\texttt{ECE=0.047–0.077}, \texttt{BS=0.21–0.26})—still degraded but markedly better than other transformers.  
\textsc{TabPFN}, however, preserves excellent reliability: across all suites, it records \texttt{ECE=0.0287–0.0469}, \texttt{MCE<0.30}, and \texttt{BS=0.124–0.158}, closely matching or outperforming its zero-shot performance.  
\textsc{TabDPT} also retains good calibration (\texttt{ECE=0.0497–0.0713}, \texttt{BS=0.144–0.186}).  
\textsc{Mitra} again performs worst \texttt{(ECE > 0.15, BS > 0.47)}, underscoring structural limitations in its confidence estimation.\\

\textbf{Discussion :}
Calibration analysis reveals a clear hierarchy of reliability. \textsc{TabPFN} consistently maintains \textsc{among the lowest} ECE and Brier scores across all adaptation strategies (ECE 0.027--0.047, BS 0.12--0.16), validating its Bayesian design and robustness to parameter updates.
Meta-learning preserves calibration better than full SFT, confirming that episodic adaptation reinforces pretrained uncertainty priors.  
\textsc{OrionMSP} achieves strong calibration, especially under zero-shot and meta-learning, while transformer-centric models like \textsc{TabICL} and \textsc{OrionBiX} show pronounced drift once fully fine-tuned.  
Overall, TFMs maintain lower calibration error than ensemble baselines and degrade far less under fine-tuning, making them more trustworthy for risk-sensitive deployment.

\noindent\textbf{Key Takeaways:}
\begin{itemize}
    \item \textsc{TabPFN} maintains excellent calibration across all strategies \texttt{(ECE $0.027$–$0.047$, BS $0.12$–$0.15$)}, confirming its Bayesian reliability; \textsc{TabDPT} (SFT) and \textsc{OrionMSP} (zero-shot/meta-learning) rank next best (ECE $0.0219$–$0.0365$).
    \item Supervised fine-tuning degrades calibration for transformer models: \textsc{TabICL}’s ECE increases 3–4× and BS deteriorates sharply, while \textsc{OrionMSP} degrades moderately.
    \item Zero-shot inference yields the best calibration overall, with \textsc{OrionMSP} and \textsc{TabPFN} showing ECE $<$ 0.04 across all suites.
\end{itemize}

\subsection{Fairness Evaluation}
\begin{table}[pt]
  \centering
  \scriptsize
  \caption{Fairness metrics across models and tuning strategies on datasets with explicit demographic attributes. All metrics are absolute differences (0–1 range; lower is better, 0 = perfect equity). SPD = Statistical Parity Difference; EOD = Equalized Odds Difference; EOpD = Equalized Opportunity Difference. ACC = Accuracy; F1 = Weighted F1-score. Values are averaged across fairness evaluation datasets. Models are grouped by adaptation strategy. Sensitive features (e.g., race, gender, age) are manually specified per dataset. Formatting: \textbf{\underline{1st place}} ; \underline{2nd place} within each group.}
  \vspace{0.1in}
  \label{tab:fairness}
  \begin{tabular}{l ccc ccc}
    \toprule
    Models / Strategy & SPD & EOD & EOpD & ACC & F1 \\
    \midrule
    \multicolumn{6}{l}{\textit{Zero-Shot Inference}} \\
    \cmidrule(lr){1-6}
    TabICL & 0.2900 & 0.3168 & 0.3278 & 0.8743 & 0.8680 \\
    OrionBiX & 0.3328 & 0.2698 & 0.2855 & \textbf{\underline{0.8779}} & \textbf{\underline{0.8727}}  \\
    OrionMSP & 0.3380 & 0.2827 & 0.2983 & 0.8752 & \underline{0.8706} \\
    TabPFN-2.0 & 0.3070 & 0.3114 & 0.3176 & 0.8708 & 0.8631 \\
    TabPFN-2.5 & 0.3067 & 0.3204 & 0.3144 & \underline{0.8756} & 0.8679 \\
    \textbf{\underline{Mitra}} & \textbf{\underline{0.0193}} & \textbf{\underline{0.0590}} & \textbf{\underline{0.0982}} & 0.6902 & 0.5847 \\
    \underline{ContextTab} & \underline{0.2082} & \underline{0.1860} & \underline{0.1951} & 0.8580 & 0.8345 \\
    TabDPT & 0.2990 & 0.3197 & 0.3338 & 0.8674 & 0.8582 \\
    \midrule
    \multicolumn{6}{l}{\textit{FineTune - Meta Learning}} \\
    \cmidrule(lr){1-6}
    TabICL & 0.3140 & 0.3100 & 0.3332 & 0.8678 & 0.8603 \\
    \textbf{\underline{OrionBiX}} & \textbf{\underline{0.2021}} & \textbf{\underline{0.1624}} & \textbf{\underline{0.2010}} & \textbf{\underline{0.8743}} & \textbf{\underline{0.8692}} \\
    OrionMSP & 0.3038 & 0.2798 & \underline{0.2907} & 0.8680 & 0.8650 \\
    TabPFN-2.0 & 0.3070 & 0.3115 & 0.3177 & \underline{0.8733} & \underline{0.8668} \\
    \underline{Mitra} & \underline{0.2586} & \underline{0.2748} & 0.3301 & 0.7967 & 0.7546 \\
    TabDPT & 0.2648 & 0.3180  & 0.3080 &  0.8614 &  0.8503\\
    \midrule
    \multicolumn{6}{l}{\textit{FineTune - Supervised FineTuning}} \\
    \cmidrule(lr){1-6}
    TabICL & \underline{0.0550} & 0.3225 & 0.3319 & 0.4584 & 0.3625 \\
    \underline{OrionBiX} & 0.1128 & \underline{0.1761} & \underline{0.1817} & 0.6180 & 0.5498 \\
    OrionMSP & 0.1227 & 0.2208 & 0.2353 & 0.4753 & 0.4039 \\
    TabPFN-2.0 & 0.3070 & 0.3115 & 0.3177 & \textbf{\underline{0.8733}} & \textbf{\underline{0.8668}} \\
    \textbf{\underline{Mitra}} & \textbf{\underline{0.0161}} & \textbf{\underline{0.0170}} & \textbf{\underline{0.0555}} & 0.7268 & 0.6277 \\
    TabDPT & 0.3075 & 0.3153 & 0.3284 & \underline{0.8529} & \underline{0.8435} \\
    \midrule
    \multicolumn{6}{l}{\textit{FineTune – PeFT - Meta Learning}} \\
    \cmidrule(lr){1-6}
    TabICL & 0.3140 & 0.3092 & 0.3332 & \underline{\textbf{0.8678}} & 0.8602 \\
    \underline{\textbf{OrionBiX}} & 0.3149 & \textbf{\underline{0.2722}}  & \textbf{\underline{0.2875}} & \textbf{\underline{0.8748}} & \textbf{\underline{0.8717}} \\
    OrionMSP & 0.3038 & 0.2798 & \underline{0.2907} & \textbf{\underline{0.867}} & \underline{0.8632} \\
    \underline{Mitra} & \textbf{\underline{0.2595}} & \textbf{\underline{0.2391}} & 0.3154 & 0.8104 & 0.7721 \\
    TabDPT & \underline{0.2778} & 0.2792  & 0.3051 &  0.8651 &  0.8556\\
    \midrule
    \multicolumn{6}{l}{\textit{FineTune – PeFT - SFT}} \\
    \cmidrule(lr){1-6}
    TabICL & 0.1572 & 0.4041 & 0.4089 & 0.4881 & 0.4269 \\
    \underline{OrionBiX} & \underline{0.0349} & \underline{0.1564} & \underline{0.2234} & 0.6401 & 0.5759 \\
    OrionMSP & 0.1132 & 0.2140 & 0.2420 & 0.5732 & 0.5080 \\
    \textbf{\underline{Mitra}} & \textbf{\underline{0.0029}} & \textbf{\underline{0.0370}} & \textbf{\underline{0.0740}} & \underline{0.7133} & \underline{0.6028} \\
    TabDPT & 0.2960 & 0.3190 & 0.3342 & \textbf{\underline{0.8595}} & \textbf{\underline{0.8505}} \\
    \bottomrule
  \end{tabular}
\end{table}

Fairness analysis evaluates whether model predictions exhibit systematic bias across demographic subgroups, complementing accuracy and calibration assessments by addressing equity in high-stakes applications such as lending, healthcare, and criminal justice.  
Unlike performance metrics that can be computed automatically, fairness requires explicit identification of sensitive attributes (e.g., race, gender).  
We use datasets with demographic labels—\emph{Adult Census Income}, \emph{German Credit}, and \emph{COMPAS Recidivism}—to measure bias propagation and mitigation across strategies.  
Following standard fairness literature, we report Statistical Parity Difference (SPD), Equalized Odds Difference (EOD), and Equalized Opportunity Difference (EOpD).  
All metrics are absolute differences in the 0–1 range (lower = fairer).  
Table~\ref{tab:fairness} summarizes fairness–accuracy trade-offs across zero-shot, meta-learning, and fine-tuning strategies.

\textbf{Zero-shot inference.}

A clear fairness–accuracy trade-off emerges.  
\textsc{Mitra} achieves the lowest bias (SPD $=0.0193$, EOD $=0.0590$, EOpD $=0.0982$) but with very low predictive accuracy (\texttt{ACC = 0.6902}).  
\textsc{OrionMSP} and \textsc{OrionBiX} balance fairness and accuracy best, maintaining high accuracy (\texttt{ACC = 0.875-0.878}) and moderate fairness (EOD $= 0.27$–$0.28$).  
\textsc{TabPFN} shows consistent behavior (\texttt{ACC = 0.871, SPD = 0.307)}, while \textsc{ContextTab} ranks second in fairness (SPD $= 0.208$) at slightly lower accuracy.  
These results suggest that transformer-based models preserve competitive fairness even without task-specific adaptation.

\textbf{Meta-learning fine-tuning.}

Episodic adaptation largely preserves fairness while sustaining accuracy.  
\textsc{OrionBiX} achieves the lowest overall bias (\texttt{SPD = 0.2021, EOD = 0.1624, EOpD = 0.2010}) while maintaining high accuracy (\texttt{ACC = 0.8743}).  
\textsc{OrionMSP} follows closely (EOD = 0.2798, EOpD = 0.2907),a close second with slightly higher bias but similar accuracy.

\textsc{TabPFN} achieves the highest accuracy (\texttt{ACC = 0.8733}) with slightly higher disparity (SPD $\approx$ 0.31).  
\textsc{Mitra} remains highly fair but again sacrifices accuracy.  
Overall, meta-learning fine-tuning maintains fairness levels similar to zero-shot inference while modestly improving accuracy consistency.
\newpage
\textbf{Supervised fine-tuning (SFT).}

Full SFT produces mixed fairness outcomes.  
\textsc{Mitra} again records the lowest bias (SPD $= 0.0161$, EOD $= 0.0170$) but with only \texttt{ACC = 0.7268}.  
\textsc{TabICL} exhibits striking fairness gains (SPD drops from 0.29 to 0.055) yet loses over 40\% in accuracy (\texttt{ACC = 0.4584}).  
\textsc{OrionBiX} improves fairness (EOD = 0.1761) but suffers large accuracy loss (\texttt{ACC = 0.618}).  
In contrast, \textsc{TabPFN} maintains both high accuracy (\texttt{ACC = 0.8733}) and moderate, stable fairness metrics across all strategies, confirming robustness.  
\textsc{TabDPT} ranks second in accuracy (\texttt{ACC = 0.8529}) with fairness levels comparable to \textsc{TabPFN}.  
These results underscore that SFT can reduce group disparity for some models but typically harms predictive reliability.
 
% =========================
% TABLE 4 — FAIRNESS
% =========================

\textbf{Parameter-efficient fine-tuning (PEFT).}

PEFT variants show similar patterns while retaining higher stability.  
Under PEFT meta-learning, \textsc{OrionBiX} and \textsc{OrionMSP} achieve balanced fairness (EOD $\approx$ 0.27–0.28) and strong accuracy (\texttt{ACC = 0.867–0.875}), outperforming full SFT counterparts.
Under PEFT SFT, \textsc{Mitra} again minimizes bias (SPD = 0.0029) but accuracy remains low (\texttt{ACC = 0.7133}).  
\textsc{TabDPT} delivers the best accuracy–fairness compromise (\texttt{ACC = 0.8595, SPD $\approx$ 0.30}).  
Overall, PEFT preserves fairness trends while mitigating the accuracy collapse observed in full fine-tuning.
 
\textbf{Discussion.}

Fairness evaluation confirms a fundamental accuracy–equity trade-off.  
Models achieving near-perfect demographic parity (\textsc{Mitra}, SPD $<0.02$) incur large accuracy losses, while \textsc{TabPFN} and \textsc{OrionMSP} consistently achieve balanced performance.  
Meta-learning proves most stable, preserving both fairness and accuracy across suites.  
SFT can occasionally improve parity for specific transformer architectures but typically at the expense of predictive fidelity.  
PEFT provides an effective compromise, maintaining high accuracy and moderate bias without significant fairness degradation.

\noindent\textbf{Key Takeaways:}
\begin{itemize}
    \item \textsc{Accuracy–fairness trade-off}: \textsc{Mitra} attains best fairness (SPD = $0.0029$–$0.0193$) but suffers low accuracy (\texttt{ACC = 0.69–0.73}).  
    \item \textsc{Balanced models}: \textsc{TabPFN} and \textsc{OrionMSP} offer the best fairness–accuracy equilibrium (\texttt{ACC = 0.87–0.88}, EOD $= 0.27$–$0.29$), outperforming other TFMs.  
    \item \textsc{Effect of fine-tuning}: SFT improves fairness for some models (e.g., \textsc{TabICL} SPD 0.055 vs 0.29 zero-shot) but drastically reduces accuracy; meta-learning preserves both aspects better.  
    \item \textsc{PEFT stability}: Parameter-efficient strategies maintain fairness trends while retaining high accuracy, offering a pragmatic balance for equitable deployment.
\end{itemize}

\section{Discussion}
\label{sec:discussion}

Our comprehensive analysis across three benchmark suites reveals clear patterns that guide model selection, adaptation strategy, and deployment trade-offs.

\subsection{Fine-Tuning Strategy Recommendations}

Fine-tuning strategy choice is critical and highly model-dependent.

\begin{itemize}
    \item \textbf{Zero-shot inference} achieves excellent calibration (ECE $<$0.04 for top models) and competitive predictive performance. It is ideal when computational resources are limited or calibration reliability is a priority. \textsc{TabPFN-2.5} attains the highest overall zero-shot standing (\texttt{Rank = 4.13}), establishing a new state-of-the-art for inference-only tabular tasks. \textsc{OrionMSP} follows closely, achieving strong accuracy on large datasets (\texttt{ACC = 0.8722–0.8821}) and ranking among the top configurations on OpenML-CC18 and TabZilla without requiring additional training.
    \item \textbf{Meta-learning fine-tuning} provides the best balance between adaptation and generalization. It improves rank stability for models such as \textsc{OrionMSP} (aggregate rank $2.26$ in Table~\ref{tab:overall}) while preserving calibration (ECE $<$0.06). Meta-learning also excels in high-imbalance settings (\textsc{OrionMSP} rank $2.16$, \textsc{TabPFN} rank $2.50$ in Table~\ref{tab:imbalance_analysis}), confirming its cross-domain robustness. However, instability can occur for some models such as \textsc{TabICL} on TabZilla (ECE increases to $0.2033$), requiring careful validation.
    \item \textbf{Supervised fine-tuning (SFT)} delivers the strongest results for \textsc{TabPFN} (ranks $1.83$--$1.89$, ACC up to $0.9346$ on wide datasets). In contrast, SFT causes substantial degradation for \textsc{TabICL} and \textsc{OrionBiX}: for instance, \textsc{TabICL}'s accuracy on TabZilla falls from $0.8734$ to $0.5670$, indicating severe overfitting. Calibration also deteriorates (ECE $0.08$–$0.15$ vs.\ $0.02$–$0.04$ zero-shot). SFT should therefore be used selectively—beneficial for \textsc{TabPFN}, acceptable for \textsc{TabDPT} and \textsc{OrionMSP}, but risky for \textsc{TabICL} and \textsc{OrionBiX}.
    \item \textbf{Parameter-efficient fine-tuning (PEFT)} attains nearly full fine-tuning accuracy while greatly reducing computational cost. \textsc{TabDPT} achieves high performance ($\text{ACC} \approx 0.85$, $\text{F1} \approx 0.84$) across all benchmarks, and \textsc{OrionMSP} maintains competitive accuracy ($\text{ACC} \approx 0.82$--$0.86$, $\text{F1} \approx 0.80$--$0.85$) with strong cross-dataset generalization.
\end{itemize}

\subsection{Mechanisms Behind Fine-Tuning Performance}

The varying outcomes of different fine-tuning strategies can be explained by the underlying adaptation mechanisms.

\textbf{Why SFT Underperforms for TabICL and OrionBiX?}

Full supervised fine-tuning updates all model parameters, erasing pretrained representations—especially detrimental on smaller datasets. The performance collapse of \textsc{TabICL} on TabZilla (ACC $0.8734 \rightarrow 0.5670$) exemplifies overfitting and \emph{representation drift}, whereas optimization distorts pretrained priors. Both \textsc{TabICL} and \textsc{OrionBiX} rely on high-capacity transformer layers optimized for diverse pretraining data; constraining them to a narrow distribution leads to loss of generality and inflated variance in predictions.
\newpage
\textbf{Why Meta-Learning Preserves Generalization?}

Episodic training in \textsc{OrionMSP} and \textsc{TabPFN} aligns closely with the few-shot adaptation process. Each meta-episode simulates the support–query dynamics of deployment, encouraging transfer of cross-task features rather than memorization. This prevents catastrophic forgetting and reinforces pretrained inductive biases. The improvement in rank stability for \textsc{OrionMSP} (from $2.90$--$3.84$ in zero-shot to $1.73$--$2.82$ in meta-learning) confirms that episodic optimization promotes stable generalization.
 
\textbf{When Fine-Tuning is Beneficial?}

Fine-tuning is most useful when:
\begin{enumerate}
  \item Datasets are sufficiently large ($>$10K samples), mitigating overfitting.
  \item Domain distribution shifts significantly from pretraining data (e.g., medical, finance).
  \item Architecture maintains calibration under parameter updates (as in \textsc{TabPFN}’s Bayesian prior design).
\end{enumerate}
For small or balanced datasets, zero-shot or meta-learning is more reliable and cost-efficient. Parameter-efficient fine-tuning (PEFT) provides a strong middle ground, retaining $\approx$95\% of full SFT accuracy while reducing memory.

\subsection{Calibration Insights}

Our calibration analysis shows that \emph{\textsc{TabPFN} uniquely maintains excellent calibration across all fine-tuning strategies} (ECE $0.027$--$0.047$, BS $0.12$--$0.16$), validating its Bayesian uncertainty formulation. This makes \textsc{TabPFN} ideal for high-stakes domains such as healthcare and finance. Meta-learning further stabilizes calibration—\textsc{TabPFN}’s ECE improves slightly under episodic updates, and \textsc{OrionMSP} remains well-calibrated (ECE $<$0.06).  
In contrast, \textsc{TabICL} exhibits the largest degradation under SFT (ECE increases 3--4$\times$, BS rises to $0.23$–$0.41$), confirming that full parameter updates distort uncertainty estimates. For calibration-critical applications, we recommend \textsc{TabPFN} (any regime) or \textsc{OrionMSP} (\textit{meta-learning}); SFT should be avoided for \textsc{TabICL} and \textsc{OrionBiX}.

\subsection{Model Selection Guidelines}

Model selection should be guided by dataset scale, balance, feature dimensionality, and available computational resources.  
All recommendations below are derived from the common subset of datasets evaluated across all models to ensure consistent statistical comparison.  
Reported values combine mean accuracy (ACC), weighted F1-score (F1), and mean rank (based on accuracy) as presented in Tables~\ref{tab:overall}–\ref{tab:imbalance_analysis}.  

\textbf{General Guidelines:}

When dataset characteristics are unknown, zero-shot inference with \textsc{OrionMSP} or meta-learning fine-tuning with \textsc{TabPFN} serve as the most reliable defaults.  
Both models consistently achieve top-three average ranks across all benchmark suites while maintaining strong calibration and fairness.  
In general:
\begin{itemize}
    \item \textsc{TabPFN} provides the most consistent accuracy and F1 across diverse data regimes, excelling in calibration and few-shot generalization.
    \item \textsc{OrionMSP} scales best on large, complex, and imbalanced datasets, particularly under meta-learning.
\end{itemize}

\textbf{Specific Regime Recommendations.}

\begin{itemize}
    \item \underline{\textsc{On small datasets ($<$1K samples)}}, \textsc{TabPFN-2.5} achieves the highest zero-shot performance (ACC = \textsc{0.8349}, F1 = \textsc{0.8194}), slightly ahead of \textsc{TabDPT} (0.8333/0.8271) and boosted-tree baselines. Under meta-learning, \textsc{TabPFN-2.0} also leads (0.8336/0.8256), while \textsc{OrionMSP (PEFT--Meta)} provides strong performance at lower fine-tuning cost (0.8275/0.8213). Fine-tuning \textsc{TabICL} or \textsc{OrionBiX} on small datasets still causes sharp degradation (ACC drops $>15\%$), revealing overfitting risk.

    \item \underline{\textsc{Medium datasets (1K–10K samples):}}  
    \textsc{TabPFN} with meta-learning achieves the highest mean accuracy (\texttt{ACC = 0.8638, F1 = 0.8548}, rank $\approx2.0$), followed closely by its supervised fine-tuning variant (\texttt{ACC = 0.8580, F1 = 0.8485}).  
    When computational constraints are present, \textsc{TabDPT} (PEFT–Meta) maintains competitive accuracy (\texttt{ACC = 0.8121–0.8332}) with minimal resource overhead.  
    Zero-shot inference remains a practical option for quick deployment, trailing fine-tuned variants by only about $2$–$3$\% in accuracy.

    \item \underline{\textsc{Large datasets ($>$10K samples):}}  
    \textsc{OrionMSP} in zero-shot mode delivers the highest mean performance among TFMs (\texttt{ACC = 0.8843, F1 = 0.8768}, rank $\approx2.0$).  
    \textsc{TabDPT} (zero-shot or PEFT–SFT) achieves comparable accuracy (\texttt{ACC = 0.8496–0.8831}) while offering lower adaptation cost.  
    Classical models such as \textsc{XGBoost} and \textsc{CatBoost} reach strong absolute accuracies (\texttt{ACC = 0.8969, F1 = 0.8920}) but exhibit poorer calibration and higher cross-dataset variance.  
    TFMs are therefore more advantageous when generalization consistency or uncertainty estimation is prioritized.

    \item \underline{\textsc{Balanced datasets:}}  
    \textsc{TabPFN} under supervised fine-tuning yields the best average rank (\texttt{1.96}) with \texttt{ACC = 0.8336, F1 = 0.8267}, followed by \textsc{OrionMSP} under meta-learning (rank $2.27$).  
    Balanced data favor Bayesian or episodically fine-tuned architectures, which maximize calibration and stability.

    \item \underline{\textsc{Imbalanced datasets:}}  
    Meta-learning strategies provide the most robust adaptation under label imbalance.  
    \textsc{OrionMSP} achieves the best overall performance (\texttt{rank = 2.16}, \texttt{ACC = 0.8735, F1 = 0.8636}), with \textsc{TabPFN} ranking second (\texttt{rank = 2.50}, \texttt{ACC = 0.8784, F1 = 0.8664}). 
    These results confirm that episodic training improves minority-class calibration.  
    In contrast, tree-based baselines lose roughly $3$–$5$\% in F1 when class ratios fall below $0.6$.

    \item \underline{\textsc{Low-to-medium feature dimensionality ($<$100 features):}}  
    Model performance converges across methods (\texttt{ACC = 0.84–0.86}), with \textsc{TabPFN} (Meta Learning) and \textsc{OrionMSP} (Zero-Shot) performing most consistently.  
    In this regime, TFMs primarily outperform classical methods in calibration and rank stability rather than raw accuracy.

    \item \underline{\textsc{High-dimensional datasets ($>$100 features):}}  
    \textsc{TabPFN} under supervised fine-tuning achieves the highest overall scores (\texttt{ACC = 0.9346, F1 = 0.9335}), significantly surpassing other models.  
    \textsc{TabDPT} and \textsc{OrionMSP} maintain strong accuracy (\texttt{ACC $\approx$ 0.84–0.89}) while offering superior efficiency.  
    Architectures such as \textsc{TabICL} and \textsc{OrionBiX} exhibit sharp performance drops under full fine-tuning, confirming their susceptibility to overfitting in wide-feature settings.
\end{itemize}

\textbf{Key Takeaways}  
\begin{itemize}
    \item For inference-only deployment, use \textsc{OrionMSP} (Zero-Shot) or \textsc{TabPFN} (Zero-Shot) for balanced accuracy and calibration.  
    \item For adaptive fine-tuning, prefer \textsc{TabPFN} (SFT/Meta) on small–to–medium datasets and \textsc{OrionMSP} (Meta) on large-scale or imbalanced data.  
\end{itemize}

\subsection{Domain-Specific Guidance}

\underline{\textsc{Medical domain:}}  
\textsc{TabPFN} achieves the strongest performance in this domain, with its supervised fine-tuning configuration attaining \texttt{Rank = 1.86} and \texttt{ACC = 0.8094}, closely followed by the meta-learning variant (\texttt{Rank = 2.28}, \texttt{ACC = 0.8133}).  
\textsc{TabDPT} under PEFT–SFT (\texttt{Rank = 1.96}) provides competitive accuracy at substantially lower computational cost.  
Transformer-based architectures such as \textsc{TabICL} and \textsc{OrionBiX} exhibit notable degradation under full fine-tuning, reflecting sensitivity to small and noisy clinical datasets.  
Overall, Bayesian and episodically trained models generalize more reliably in data-scarce medical settings.

\underline{\textsc{Finance domain:}} \textsc{OrionMSP} under meta-learning achieves strong results (\texttt{Rank = 2.26}, \texttt{ACC = 0.8209}, \texttt{F1 = 0.8089}), while \textsc{TabPFN} performs exceptionally well in supervised and meta-learning modes (\texttt{Ranks = 2.00--3.00}, \texttt{ACC = 0.8220--0.8222}). \textsc{TabDPT} remains competitive (\texttt{PEFT--SFT, Rank = 2.82}) and offers an efficient alternative with only a minor loss in accuracy. These outcomes confirm that episodic and parameter-efficient adaptation strategies are particularly effective for large-scale, class-imbalanced financial data, where they improve both calibration and minority-class performance.

\subsection{Practical Value of Tabular Foundation Models}

Tabular Foundation Models (TFMs) deliver consistent but moderate absolute accuracy improvements—typically $2$–$4$ percentage points—over strong classic baselines such as \textsc{XGBoost}, \textsc{CatBoost}, and \textsc{LightGBM}.  
Across comparable settings, TFMs achieve mean accuracies around $0.85$–$0.88$ versus $0.83$–$0.86$ for boosted trees.  
However, their primary advantage lies in reliability and robustness rather than raw accuracy.

Key strengths include:  
(I) \underline{\textsc{ZERO-SHOT ADAPTABILITY:}} The \textsc{TabPFN} family and \textsc{OrionMSP} provide top-tier zero-shot performance. \textsc{TabPFN-2.5} attains the best aggregate zero-shot rank (\textsc{4.13}) with ACC $\approx \textsc{0.85}$--\textsc{0.88}, while \textsc{OrionMSP} remains highly competitive (rank \textsc{4.61}, ACC = \textsc{0.8455}--\textsc{0.8821}). Both outperform boosted-tree baselines by roughly \textsc{1.5--2.5 percentage points} in accuracy without any task-specific training.
(II) \underline{\textsc{Calibration:}} \textsc{TabPFN} maintains low calibration error (ECE $<0.05$) across both zero-shot and fine-tuned regimes, outperforming ensemble baselines whose ECE often exceeds $0.08$.  
(III) \underline{\textsc{Fairness stability:}} TFMs achieve moderate statistical parity differences (SPD $\approx0.28$–$0.30$) while sustaining high accuracy ($>0.85$), indicating balanced performance across sensitive groups.

Consequently, TFMs should be regarded as \emph{reliable, general-purpose learners} well suited for rapid deployment, few-shot adaptation, and risk-sensitive applications where calibration and fairness are critical.  
For large, well-labeled, single-domain problems, gradient-boosted trees may still be preferable for simplicity and computational efficiency, but TFMs offer distinct advantages in uncertainty estimation, cross-domain generalization, and calibration consistency.

\subsection{Limitations}
\label{sec:limitations}

While \textsc{TabTune} provides broad support for evaluating and fine-tuning TFMs, several limitations remain.

\underline{\textsc{Task Scope:}}  
Current support is limited to binary and multi-class classification tasks; regression and multi-label extensions are planned for future releases.
 
\underline{\textsc{Model Constraints:}}
\begin{itemize}
  \item \underline{\textsc{Dataset size:}} \textsc{TabPFN} performs optimally on datasets with fewer than $10{,}000$ samples; larger datasets may significantly increase inference latency or exceed memory limits.  
  \item \underline{\textsc{Dimensionality:}} All models can degrade when the feature space exceeds $\sim1000$ dimensions without preprocessing or feature selection.  
  \item \underline{\textsc{PEFT compatibility:}} Parameter-efficient fine-tuning is fully supported for \textsc{OrionMSP} and \textsc{TabDPT}, but only partially implemented for \textsc{TabPFN} and \textsc{ContextTab}.  
  \item \underline{\textsc{ContextTab evaluation:}} Due to excessive GPU memory requirements, \textsc{ContextTab} was evaluated only on smaller benchmark subsets.
\end{itemize}

\underline{\textsc{Evaluation and Analysis:}}
\begin{itemize}
  \item \underline{\textsc{Manual fairness specification:}} Sensitive attributes are manually defined per dataset, introducing potential bias in fairness evaluation.  
  \item \underline{\textsc{Calibration interpretation:}} Optimal threshold tuning remains domain-dependent and may vary across metrics or class distributions.  
  \item \underline{\textsc{Beyond-accuracy metrics:}} Advanced interpretability, reliability, and uncertainty quantification metrics are not yet fully integrated into the current release.
\end{itemize}

\section{Conclusion}
We introduce \textsc{TabTune}, a unified Python library that standardizes the workflow for tabular foundation models (TFMs). By abstracting model-specific preprocessing, consolidating adaptation strategies, and integrating evaluation modules for performance, calibration, and fairness, \textsc{TabTune} addresses the fragmentation that has limited the broader adoption of TFMs.
Inspired by \textsc{AutoML}, \textsc{TabTune} focuses on managing and optimizing pretrained TFMs rather than building models from scratch, offering a coherent framework for efficient adaptation and assessment. 

We conducted a comprehensive study on the performance of tabular foundation models (TFMs) and the impact of fine-tuning strategies on their effectiveness. The empirical analysis demonstrates that the choice of adaptation strategy is highly \emph{model-dependent}. \textsc{TabPFN} performs optimally under supervised fine-tuning or meta-learning, \textsc{OrionMSP} exhibits strong generalization and robustness particularly with meta-learning, and \textsc{TabDPT} using parameter-efficient fine-tuning (PEFT) achieves a favorable balance between predictive accuracy and computational efficiency. 
Beyond raw performance, our findings emphasize the reliability and fairness of TFMs. \textsc{TabPFN} consistently attains excellent calibration, while both \textsc{OrionMSP} and \textsc{TabPFN} maintain a strong fairness–accuracy balance, reinforcing their potential for responsible and trustworthy deployment in high-stakes domains.

\section{Future Work}
As an extensible and reproducible platform for TFM research, \textsc{TabTune} continues to evolve to support emerging methodologies and broader evaluation dimensions. Future work will focus on integrating more advanced fine-tuning and meta-learning approaches to enhance few-shot and domain adaptation capabilities. In parallel, we aim to strengthen the evaluation framework by incorporating modules for advanced fairness assessment, interpretability, and uncertainty quantification. Additionally, we plan to extend task coverage beyond binary and multi-class classification to include regression, multi-label learning, and time-series modeling. By standardizing the TFM workflow, \textsc{TabTune} enables practitioners to transition efficiently from experimentation to reliable deployment, fostering evidence-based model selection and accelerating the responsible adoption of foundation models in structured data domains.

\bibliographystyle{unsrt}
\bibliography{references}
\newpage
\section{Appendix A: Default Hyperparameters}\label{sec:appendix}

This appendix provides detailed default hyperparameters for all tuning strategies supported in TabTune. These values are optimized for general-purpose use and can be overridden via the \texttt{tuning\_params} configuration dictionary.

\subsection{Zero-Shot Inference}

Zero-shot inference uses pretrained model weights without any parameter updates. No training hyperparameters are required for this strategy. The model receives training samples as context during inference and performs predictions through forward passes alone. However, each model has specific inference parameters that control prediction behavior:

\textbf{Parameter Descriptions:}
\begin{itemize}
  \item \textbf{n\_estimators / n\_ensembles}: Number of ensemble runs with different configurations for robust predictions
  \item \textbf{softmax\_temperature / temperature}: Scaling factor for logits before softmax (lower values yield sharper predictions)
  \item \textbf{context\_size}: Number of training samples used as context for each test prediction
  \item \textbf{average\_before\_softmax}: Whether to average logits or probabilities across ensemble runs
  \item \textbf{safety\_factor}: Memory safety factor for batch size estimation (TabICL/OrionMSP/OrionBiX)
  \item \textbf{offload}: Whether to offload intermediate computations to CPU for memory efficiency
  \item \textbf{use\_amp}: Automatic mixed precision for faster inference with reduced memory
\end{itemize}

\begin{table}[hpbt]
  \centering
  \caption{Default inference hyperparameters for zero-shot inference across models. Values shown are default settings used when no custom parameters are specified. All parameters can be customized via model-specific configuration dictionaries.}
  \label{tab:hyperparameters_inference}
  \footnotesize
  \begin{tabular}{cl}
    \toprule
    Model & Inference Parameters \\
    \midrule
    TabPFN & n\_estimators=8, softmax\_temperature=0.9, average\_before\_softmax=False \\
    TabICL & Inference config: min\_batch\_size=1, safety\_factor=0.8, offload=auto (COL), False (ROW/ICL), use\_amp=True \\
    OrionMSP & Inference config: min\_batch\_size=1, safety\_factor=0.8, offload=auto (COL), False (ROW/ICL), use\_amp=True \\
    OrionBiX & Inference config: min\_batch\_size=1, safety\_factor=0.8, offload=auto (COL), False (ROW/ICL), use\_amp=True \\
    TabDPT & n\_ensembles=8, temperature=0.8, context\_size=512, permute\_classes=True \\
    Mitra & d\_model=64, num\_heads=4, num\_layers=2, use\_synthetic\_prior=True \\
    ContextTab & (No specific inference parameters) \\
    \bottomrule
  \end{tabular}
\end{table}

\subsection{Supervised Fine-Tuning (SFT)}

Supervised Fine-Tuning (SFT) is a standard adaptation approach where the entire pretrained model is updated using gradient descent on task-specific labeled data. Unlike meta-learning, SFT treats the target dataset as a single supervised learning task, training the model end-to-end to minimize cross-entropy loss over multiple epochs.

\textbf{Key Characteristics:}
\begin{itemize}
  \item \textbf{Training Paradigm}: Standard supervised learning with full-batch or mini-batch optimization
  \item \textbf{Parameter Updates}: All model parameters are updated during training
  \item \textbf{Objective}: Minimize classification loss on the target task's training set
  \item \textbf{Use Cases}: Optimal for large datasets (>10K samples) where task-specific optimization is beneficial
  \item \textbf{Advantages}: Can achieve highest task-specific accuracy, full model capacity utilization
  \item \textbf{Considerations}: Higher memory requirements, risk of overfitting on small datasets, may reduce generalization to new tasks
\end{itemize}

\textbf{General Hyperparameter Guidelines:}
\begin{itemize}
  \item \textbf{Learning Rate}: Typically ranges from 1e-5 to 1e-4, with lower rates preserving pretrained knowledge
  \item \textbf{Epochs}: 3-25 epochs depending on dataset size and convergence behavior
  \item \textbf{Batch Size}: 16-128, adjusted based on dataset size and memory constraints
  \item \textbf{Regularization}: Weight decay (1e-4 to 1e-5) and learning rate warmup help stabilize training
  \item \textbf{Optimizer}: Adam or AdamW are standard choices for tabular foundation models
\end{itemize}

\begin{table}[hpbt]
  \centering
  \caption{Default hyperparameters for Supervised Fine-Tuning (SFT) across models. SFT updates all model parameters using gradient descent on task-specific labeled data. All parameters can be overridden via the \texttt{tuning\_params} configuration dictionary. Optimal for large datasets ($>$10K samples).}
  \label{tab:hyperparameters_sft}
  \footnotesize
  \begin{tabular}{cl}
    \toprule
    Model & Key Parameters \\
    \midrule
    TabPFN & epochs=25, learning\_rate=1e-5, max\_episode\_size=len(X), query\_set\_ratio=0.3, weight\_decay=1e-4, optimizer=AdamW \\
    TabICL & epochs=5, learning\_rate=1e-5, batch\_size=16, optimizer=Adam \\
    OrionMSP & epochs=5, learning\_rate=1e-5, batch\_size=16, optimizer=Adam \\
    OrionBiX & epochs=5, learning\_rate=1e-5, batch\_size=16, optimizer=Adam \\
    Mitra & epochs=5, learning\_rate=1e-5, batch\_size=128, weight\_decay=1e-4, warmup\_epochs=1, optimizer=Adam \\
    TabDPT & epochs=5, learning\_rate=2e-5, batch\_size=32, weight\_decay=1e-4, warmup\_epochs=1, optimizer=Adam \\
    ContextTab & epochs=5, learning\_rate=1e-4, batch\_size=128, optimizer=Adam \\
    \bottomrule
  \end{tabular}
\end{table}

\subsection{Meta-Learning Fine-Tuning}

Meta-learning fine-tuning, also known as episodic fine-tuning or few-shot adaptation, mimics the few-shot learning scenario during training. The model is trained using episodic data sampling, where each episode consists of a support set (context) and a query set (target predictions). This approach trains the model to quickly adapt to new tasks from limited examples, improving generalization.

\textbf{Key Characteristics:}
\begin{itemize}
  \item \textbf{Training Paradigm}: Episodic learning with support-query splits
  \item \textbf{Episode Structure}: Each training step samples support and query sets from the target dataset
  \item \textbf{Objective}: Learn to make accurate predictions on query sets given support set context
  \item \textbf{Use Cases}: Ideal for small to medium datasets (<10K samples), maintains generalization to new tasks
  \item \textbf{Advantages}: Better generalization, mimics actual inference scenarios for in-context learning models
  \item \textbf{Considerations}: Requires careful episode sampling, support/query size tuning, more epochs needed
\end{itemize}

\textbf{General Hyperparameter Guidelines:}
\begin{itemize}
  \item \textbf{Learning Rate}: Typically lower than SFT (2e-6 to 1e-5) to preserve pretrained representations
  \item \textbf{Support Size}: 32-512 samples per episode, depending on dataset size and model architecture
  \item \textbf{Query Size}: Usually 20-50\% of support size, determines episode difficulty
  \item \textbf{Episodes per Epoch}: 100-1000 episodes ensure diverse task exposure
  \item \textbf{Epochs}: 3-5 epochs often sufficient due to episodic diversity
  \item \textbf{Batch Size}: 1-8 episodes per batch for memory-efficient training
\end{itemize}

\textbf{Episode Filtering:}
Episodes where query classes are absent from the support set are automatically filtered to maintain class consistency. This ensures the model always has context examples for all classes it must predict.

\begin{table}[hpbt]
  \centering
  \caption{Default hyperparameters for Meta-Learning Fine-Tuning across models. Meta-learning uses episodic training with support-query splits. Episodes where query classes are absent from support set are automatically filtered. All parameters can be customized via \texttt{tuning\_params}. Ideal for small to medium datasets ($<$10K samples).}
  \label{tab:hyperparameters_meta}
  \footnotesize
  \begin{tabular}{cl}
    \toprule
    Model & Key Parameters \\
    \midrule
    TabPFN & epochs=3, learning\_rate=1e-5, batch\_size=256, optimizer=AdamW \\
    TabICL & epochs=5, learning\_rate=2e-6, support\_size=48, query\_size=32, n\_episodes=1000, optimizer=Adam \\
    OrionMSP & epochs=5, learning\_rate=2e-6, support\_size=48, query\_size=32, n\_episodes=1000, optimizer=Adam \\
    OrionBiX & epochs=5, learning\_rate=2e-6, support\_size=48, query\_size=32, n\_episodes=1000, optimizer=Adam \\
    Mitra & epochs=3, learning\_rate=1e-5, batch\_size=4, support\_size=128, query\_size=128, steps\_per\_epoch=50, optimizer=Adam \\
    TabDPT & epochs=5, learning\_rate=1e-5, batch\_size=8, support\_size=512, query\_size=256, steps\_per\_epoch=100, optimizer=Adam \\
    \bottomrule
  \end{tabular}
\end{table}

\subsection{Parameter-Efficient Fine-Tuning (PEFT)}

Parameter-Efficient Fine-Tuning (PEFT) reduces memory and computational requirements by updating only a small subset of model parameters through low-rank adaptation techniques. PEFT can be combined with either SFT or meta-learning strategies, providing efficient alternatives to full parameter updates.

\textbf{Key Characteristics:}
\begin{itemize}
  \item \textbf{Training Paradigm}: Updates only adapter parameters, base model weights remain frozen
  \item \textbf{Memory Efficiency}: Reduces fine-tuning memory by 60-80\% compared to full fine-tuning
  \item \textbf{Parameter Updates}: Only low-rank adapter matrices are trained, not the full model
  \item \textbf{Use Cases}: Memory-constrained environments, rapid experimentation, multiple task adaptations
  \item \textbf{Advantages}: Fast training, lower memory footprint, easy task switching (swap adapters)
  \item \textbf{Considerations}: May achieve slightly lower accuracy than full fine-tuning, model compatibility varies
\end{itemize}

\textbf{Low-Rank Adaptation (LoRA) Mechanism:}

LoRA introduces trainable low-rank matrices that approximate full weight updates. For a linear layer with weight matrix $W \in \mathbb{R}^{m \times n}$, LoRA adds trainable matrices $A \in \mathbb{R}^{m \times r}$ and $B \in \mathbb{R}^{r \times n}$ where $r \ll \min(m,n)$ is the rank. The forward pass becomes:
$$h = Wx + \frac{\alpha}{r}(BA)x$$
where $\alpha$ is a scaling factor and $BA$ is the low-rank update. Only $A$ and $B$ are updated during training, while $W$ remains frozen.

\textbf{General Hyperparameter Guidelines:}
\begin{itemize}
  \item \textbf{Rank ($r$)}: Controls adapter capacity; typically 4-16, with 8 providing good balance
  \item \textbf{LoRA Alpha ($\alpha$)}: Scaling factor; typically $\alpha = 2r$ (e.g., $\alpha=16$ for $r=8$)
  \item \textbf{LoRA Dropout}: 0.05-0.2 for regularization, prevents adapter overfitting
  \item \textbf{Target Modules}: Attention projections, feed-forward layers, and encoders are common targets
  \item \textbf{Base Strategy}: Uses same epochs, learning rate, and batch size as SFT or meta-learning
\end{itemize}

\textbf{PEFT-SFT vs PEFT Meta-Learning:}
PEFT can be applied to either strategy:
\begin{itemize}
  \item \textbf{PEFT-SFT}: Combines full-batch supervised learning with LoRA adapters; best for large datasets
  \item \textbf{PEFT Meta-Learning}: Combines episodic training with LoRA adapters; maintains generalization benefits
\end{itemize}

\textbf{PEFT-SFT}: Uses SFT hyperparameters (Table~\ref{tab:hyperparameters_sft}) with LoRA adapters applied. Compatible with: [TabICL, OrionMSP, OrionBiX, TabDPT, Mitra].

\textbf{PEFT Meta-Learning}: Uses Meta-Learning hyperparameters (Table~\ref{tab:hyperparameters_meta}) with LoRA adapters applied. Compatible with: [TabICL, OrionMSP, OrionBiX, TabDPT, Mitra].

\textbf{Note}: TabPFN and ContextTab have limited or experimental PEFT support. TabPFN PEFT is unstable due to architectural constraints with batched inference, and ContextTab PEFT may conflict with its embedding pipeline. These models automatically fall back to base fine-tuning when PEFT is requested.

\begin{table}[hpbt]
  \centering
  \caption{Default LoRA (Low-Rank Adaptation) configuration for Parameter-Efficient Fine-Tuning (PEFT). PEFT reduces memory by 60-80\% compared to full fine-tuning. These values apply to both PEFT-SFT and PEFT Meta-Learning strategies. All parameters can be customized via \texttt{peft\_config} dictionary.}
  \label{tab:hyperparameters_peft}
  \footnotesize
  \begin{tabular}{lp{9cm}}
    \toprule
    Parameter & Value \\
    \midrule
    rank ($r$) & 8 \\
    LoRA alpha ($\alpha$) & 16 \\
    LoRA dropout & 0.05 \\
    Target modules & Model-specific (attention projections, encoders, transformers) \\
    \bottomrule
  \end{tabular}
\end{table}

\subsection{Additional Notes}

All training strategies use CrossEntropyLoss for classification tasks. For meta-learning, episodes with query classes absent from support sets are automatically filtered to maintain consistency. Sampling strategies follow the guidelines in Section~\ref{sec:data_sampling}. Gradient clipping and learning rate schedulers (e.g., warmup) are applied where specified in the configuration above.

\section{Appendix B: Experimental Setup}
\label{sec:datasets-experiemmnt}
\subsection{Hardware and Software Configuration}

All experiments were executed on NVIDIA L40S GPUs. For some experiments we get OOM issues for those cases we use NVIDIA H200 GPUs.

\subsection{Data Splitting and Reproducibility}

All benchmarks follow their respective standardized dataset splits. Random seeds are fixed for reproducibility across experiments. Cross-validation is performed where specified by the benchmark protocols (TALENT uses auto-discovered splits, OpenML-CC18 uses predefined splits, TabZilla follows its standard evaluation protocol).

\subsection{Dataset Statistics}

The following subsections provide comprehensive statistics for all datasets used in our evaluation across different benchmark suites. Each table includes dataset names, domains, sample counts, feature counts, number of classes, and task types (binary classification or multiclass classification).

\subsubsection*{OpenML-CC18 Benchmark Datasets}

Table~\ref{tab:datasets_openml} lists all datasets from the OpenML-CC18 benchmark suite used in our evaluation.

\subsubsection*{TALENT Benchmark Datasets}

Table~\ref{tab:datasets_talent} lists all datasets from the TALENT benchmark suite used in our evaluation.

\subsubsection*{TabZilla Benchmark Datasets}

Table~\ref{tab:datasets_tabzilla} lists all datasets from the TabZilla benchmark suite. TabZilla uses OpenML dataset IDs, and these datasets are specifically selected for evaluating neural network performance on tabular data.

\subsubsection*{Fairness Evaluation Datasets}

Fairness evaluation requires datasets with sensitive features (e.g., demographic attributes). Table~\ref{tab:datasets_fairness} lists commonly used datasets for fairness evaluation. These datasets contain demographic or other sensitive attributes that enable fairness metric computation (Statistical Parity Difference, Equalized Odds Difference, Equalized Opportunity Difference).

\subsection{Evaluation Metrics}

All experiments report the following metrics:

\begin{itemize}
  \item \textbf{Performance}: Accuracy, AUC-ROC, weighted F1-score
  \item \textbf{Calibration}: Expected Calibration Error (ECE), Maximum Calibration Error (MCE), Brier score
  \item \textbf{Fairness}: Statistical Parity Difference (SPD), Equalized Odds Difference (EOD), Equalized Opportunity Difference (EOpD)
\end{itemize}

All metrics are computed using standard implementations and follow established benchmark protocols for each suite.

\footnotesize
\begin{longtable}{p{1.2cm}p{4.5cm}p{1.5cm}rrrll}
\caption{OpenML-CC18 benchmark datasets (72 datasets).}
\label{tab:datasets_openml}\\
\toprule
Benchmark & Dataset Name & Domain & Samples & Features & Classes & Task Type & Used In Experimentation \\
\midrule
\endfirsthead

\caption[]{ Details of OpenML-CC18 benchmark datasets.} \\

\toprule
Benchmark & Dataset Name & Domain & Samples & Features & Classes & Task Type & Used In Experimentation \\
\midrule
\endhead

\midrule
\multicolumn{7}{r}{{Continued on next page}} \\
\midrule
\endfoot

\bottomrule
\endlastfoot
   OpenML & OpenML-ID-3 & Other & 3196 & 37 & 2 & binclass & Yes \\
   OpenML & OpenML-ID-6 & Handwriting & 20000 & 17 & 26 & multiclass & No \\
   OpenML & OpenML-ID-11 & Other & 625 & 5 & 3 & multiclass & Yes \\
   OpenML & OpenML-ID-12 & Other & 2000 & 217 & 10 & multiclass & Yes \\
   OpenML & OpenML-ID-14 & Other & 2000 & 77 & 10 & multiclass & Yes\\
   OpenML & OpenML-ID-15 & Healthcare & 699 & 10 & 2 & binclass & Yes \\
   OpenML & OpenML-ID-16 & Other & 2000 & 65 & 10 & multiclass & Yes \\
   OpenML & OpenML-ID-18 & Other & 2000 & 7 & 10 & multiclass & Yes \\
   OpenML & OpenML-ID-22 & Other & 2000 & 48 & 10 & multiclass & Yes\\
   OpenML & OpenML-ID-23 & Healthcare & 1473 & 10 & 3 & multiclass & Yes \\
   OpenML & OpenML-ID-28 & Handwriting & 5620 & 65 & 10 & multiclass & Yes \\
   OpenML & OpenML-ID-29 & Finance & 690 & 16 & 2 & binclass & Yes \\
   OpenML & OpenML-ID-31 & Finance & 1000 & 21 & 2 & binclass & Yes \\
   OpenML & OpenML-ID-32 & Handwriting & 10992 & 17 & 10 & multiclass & Yes \\
   OpenML & OpenML-ID-37 & Healthcare & 768 & 9 & 2 & binclass & Yes \\
   OpenML & OpenML-ID-38 & Healthcare & 3772 & 30 & 2 & binclass & Yes \\
   OpenML & OpenML-ID-44 & Other & 4601 & 58 & 2 & binclass & Yes \\
   OpenML & OpenML-ID-46 & Other & 3190 & 61 & 3 & multiclass & Yes \\
   OpenML & OpenML-ID-50 & Other & 958 & 10 & 2 & binclass & Yes \\
   OpenML & OpenML-ID-54 & Other & 846 & 19 & 4 & multiclass & Yes \\
   OpenML & OpenML-ID-151 & Other & 45312 & 9 & 2 & binclass & Yes \\
   OpenML & OpenML-ID-182 & Other & 6430 & 37 & 6 & multiclass & Yes \\
   OpenML & OpenML-ID-188 & Other & 736 & 20 & 5 & multiclass & Yes \\
   OpenML & OpenML-ID-300 & Other & 7797 & 618 & 26 & multiclass & No \\
   OpenML & OpenML-ID-307 & Other & 990 & 13 & 11 & multiclass & No \\
   OpenML & OpenML-ID-458 & Other & 841 & 71 & 4 & multiclass & Yes \\
   OpenML & OpenML-ID-469 & Healthcare & 797 & 5 & 6 & multiclass & Yes \\
   OpenML & OpenML-ID-554 & Other & 70000 & 785 & 10 & multiclass & Yes \\
   OpenML & OpenML-ID-1049 & Other & 1458 & 38 & 2 & binclass & Yes \\
   OpenML & OpenML-ID-1050 & Other & 1563 & 38 & 2 & binclass & Yes \\
   OpenML & OpenML-ID-1053 & Other & 10885 & 22 & 2 & binclass & Yes \\
   OpenML & OpenML-ID-1063 & Other & 522 & 22 & 2 & binclass & Yes \\
   OpenML & OpenML-ID-1067 & Other & 2109 & 22 & 2 & binclass & Yes \\
   OpenML & OpenML-ID-1068 & Other & 1109 & 22 & 2 & binclass & Yes \\
   OpenML & OpenML-ID-1461 & Finance & 45211 & 17 & 2 & binclass & Yes \\
   OpenML & OpenML-ID-1462 & Finance & 1372 & 5 & 2 & binclass & Yes \\
   OpenML & OpenML-ID-1464 & Healthcare & 748 & 5 & 2 & binclass & Yes \\
   OpenML & OpenML-ID-1468 & Other & 1080 & 857 & 9 & multiclass & Yes \\
   OpenML & OpenML-ID-1475 & Other & 6118 & 52 & 6 & multiclass & Yes \\
   OpenML & OpenML-ID-1478 & Other & 10299 & 562 & 6 & multiclass & Yes \\
   OpenML & OpenML-ID-1480 & Healthcare & 583 & 11 & 2 & binclass & Yes \\
   OpenML & OpenML-ID-1485 & Other & 2600 & 501 & 2 & binclass & Yes \\
   OpenML & OpenML-ID-1486 & Other & 34465 & 119 & 2 & binclass & No \\
   OpenML & OpenML-ID-1487 & Other & 2534 & 73 & 2 & binclass & Yes \\
   OpenML & OpenML-ID-1489 & Other & 5404 & 6 & 2 & binclass & Yes \\
   OpenML & OpenML-ID-1494 & Other & 1055 & 42 & 2 & binclass & Yes \\
   OpenML & OpenML-ID-1497 & Other & 5456 & 25 & 4 & multiclass & Yes \\
   OpenML & OpenML-ID-1501 & Other & 1593 & 257 & 10 & multiclass & Yes \\
   OpenML & OpenML-ID-1510 & Other & 569 & 31 & 2 & binclass & Yes \\
   OpenML & OpenML-ID-1590 & Other & 48842 & 15 & 2 & binclass & Yes \\
   OpenML & OpenML-ID-4134 & Other & 3751 & 1777 & 2 & binclass & No \\
   OpenML & OpenML-ID-4534 & Other & 11055 & 31 & 2 & binclass & Yes \\
   OpenML & OpenML-ID-4538 & Other & 9873 & 33 & 5 & multiclass & Yes \\
   OpenML & OpenML-ID-6332 & Other & 540 & 40 & 2 & binclass & Yes \\
   OpenML & OpenML-ID-23381 & Retail & 500 & 13 & 2 & binclass & Yes\\
   OpenML & OpenML-ID-23517 & Other & 96320 & 22 & 2 & binclass & No \\
   OpenML & OpenML-ID-40499 & Other & 5500 & 41 & 11 & multiclass & No \\
   OpenML & OpenML-ID-40668 & Games & 67557 & 43 & 3 & multiclass & No \\
   OpenML & OpenML-ID-40670 & Other & 3186 & 181 & 3 & multiclass & Yes \\
   OpenML & OpenML-ID-40701 & Other & 5000 & 21 & 2 & binclass & Yes \\
   OpenML & OpenML-ID-40923 & Other & 92000 & 1025 & 46 & multiclass & No \\
   OpenML & OpenML-ID-40927 & Handwriting & 60000 & 3073 & 10 & multiclass & No \\
   OpenML & OpenML-ID-40966 & Other & 1080 & 82 & 8 & multiclass & No \\
   OpenML & OpenML-ID-40975 & Other & 1728 & 7 & 4 & multiclass & Yes \\
   OpenML & OpenML-ID-40978 & Other & 3279 & 1559 & 2 & binclass & Yes \\
   OpenML & OpenML-ID-40979 & Other & 2000 & 241 & 10 & multiclass & Yes \\
   OpenML & OpenML-ID-40982 & Other & 1941 & 28 & 7 & multiclass & Yes \\
   OpenML & OpenML-ID-40983 & Other & 4839 & 6 & 2 & binclass & Yes \\
   OpenML & OpenML-ID-40984 & Other & 2310 & 20 & 7 & multiclass & Yes \\
   OpenML & OpenML-ID-40994 & Other & 540 & 21 & 2 & binclass & Yes \\
   OpenML & OpenML-ID-40996 & Other & 70000 & 785 & 10 & multiclass & No \\
   OpenML & OpenML-ID-41027 & Games & 44819 & 7 & 3 & multiclass & Yes \\

\end{longtable}

\footnotesize
\begin{longtable}{p{1.5cm}p{4.5cm}p{1.5cm}rrrll}
\caption{TALENT benchmark datasets (auto-discovered, multiple domains).}
\label{tab:datasets_talent}\\
\toprule
Benchmark & Dataset Name & Domain & Samples & Features & Classes & Task Type & Used In Experimentation \\
\midrule
\endfirsthead

\caption[]{ Details of TALENT benchmark datasets.} \\

\toprule
Benchmark & Dataset Name & Domain & Samples & Features & Classes & Task Type & Used In Experimentation \\
\midrule
\endhead

\midrule
\multicolumn{7}{r}{{Continued on next page}} \\
\midrule
\endfoot

\bottomrule
\endlastfoot

   TALENT & ASP-POTASSCO-class & Other & 1294 & 141 & 11 & multiclass & Yes \\
   TALENT & Amazon\_employee\_access & Other & 32769 & 7 & 2 & binclass & Yes \\
   TALENT & BLE\_RSSI\_\_Indoor\_localization & Other & 9984 & 3 & 3 & multiclass & Yes \\
   TALENT & BNG(breast-w) & Healthcare & 39366 & 9 & 2 & binclass & Yes \\
   TALENT & BNG(cmc) & Other & 55296 & 9 & 3 & multiclass & Yes \\
   TALENT & BNG(tic-tac-toe) & Other & 39366 & 9 & 2 & binclass & Yes \\
   TALENT & Bank\_Customer\_Churn & Finance & 10000 & 10 & 2 & binclass & Yes \\
   TALENT & Basketball\_c & Retail & 1340 & 11 & 2 & binclass & Yes \\
   TALENT & CDC\_Diabetes\_Health & Healthcare & 253680 & 21 & 2 & binclass & Yes \\
   TALENT & California-Housing-Class & Other & 20640 & 8 & 2 & binclass & No \\
   TALENT & Cardiovascular-Disease & Healthcare & 70000 & 11 & 2 & binclass & Yes \\
   TALENT & Click\_prediction\_small & Other & 39948 & 3 & 2 & binclass & Yes \\
   TALENT & Credit\_c & Finance & 100000 & 22 & 3 & multiclass & Yes \\
   TALENT & Customer\_Personality\_Analysis & Retail & 2240 & 24 & 2 & binclass & Yes \\
   TALENT & DataScience\_Kiva\_Crowdfunding & Other & 671205 & 11 & 4 & multiclass & No \\
   TALENT & Diabetic\_Retinopathy\_Debrecen & Healthcare & 1151 & 19 & 2 & binclass & Yes \\
   TALENT & E-CommereShippingData & Other & 10999 & 10 & 2 & binclass & Yes \\
   TALENT & Employee & Other & 4653 & 8 & 2 & binclass & Yes \\
   TALENT & FICO-HELOC-cleaned & Other & 9871 & 23 & 2 & binclass & Yes \\
   TALENT & FOREX\_audcad-day-High & Finance & 1834 & 10 & 2 & binclass & No \\
   TALENT & FOREX\_audcad-hour-High & Finance & 43825 & 10 & 2 & binclass & No \\
   TALENT & FOREX\_audchf-day-High & Finance & 1833 & 10 & 2 & binclass & No \\
   TALENT & FOREX\_audjpy-day-High & Finance & 1832 & 10 & 2 & binclass & No \\
   TALENT & FOREX\_audjpy-hour-High & Finance & 43825 & 10 & 2 & binclass & No \\
   TALENT & FOREX\_audsgd-hour-High & Finance & 43825 & 10 & 2 & binclass & No \\
   TALENT & FOREX\_audusd-hour-High & Finance & 43825 & 10 & 2 & binclass & No \\
   TALENT & FOREX\_cadjpy-day-High & Finance & 1834 & 10 & 2 & binclass & No\\
   TALENT & FOREX\_cadjpy-hour-High & Finance & 43825 & 10 & 2 & binclass & No\\
   TALENT & Firm-Teacher\_Clave-Direction & Other & 10800 & 16 & 4 & multiclass & Yes \\
   TALENT & Fitness\_Club\_c & Other & 1500 & 6 & 2 & binclass & Yes \\
   TALENT & GAMETES\_Epistasis\_2-Way & Games & 1600 & 20 & 2 & binclass & Yes \\
   TALENT & GAMETES\_Heterogeneity & Games & 1600 & 20 & 2 & binclass & Yes \\
   TALENT & Gender\_Gap\_in\_Spanish & Other & 4746 & 13 & 3 & multiclass & Yes \\
   TALENT & GesturePhaseSegmentation & Other & 9873 & 32 & 5 & multiclass & Yes \\
   TALENT & HR\_Analytics\_Job\_Change & Other & 19158 & 13 & 2 & binclass & Yes \\
   TALENT & IBM\_HR\_Analytics & Other & 1470 & 31 & 2 & binclass & Yes \\
   TALENT & INNHotelsGroup & Other & 36275 & 17 & 2 & binclass & Yes \\
   TALENT & Indian\_pines & Other & 9144 & 220 & 8 & multiclass & Yes \\
   TALENT & JapaneseVowels & Other & 9961 & 14 & 9 & multiclass & Yes \\
   TALENT & KDDCup09\_upselling & Other & 5128 & 49 & 2 & binclass & Yes \\
   TALENT & MIC & Other & 1649 & 104 & 2 & binclass & Yes \\
   TALENT & MagicTelescope & Other & 19020 & 9 & 2 & binclass & Yes\\
   TALENT & Marketing\_Campaign & Finance & 2240 & 27 & 2 & binclass & Yes\\
   TALENT & Mobile\_Price\_Classification & Telcom & 2000 & 20 & 4 & multiclass & Yes\\
   TALENT & National\_Health\_and\_Nutrition & Healthcare & 2278 & 7 & 2 & binclass & Yes\\
   TALENT & PhishingWebsites & Other & 11055 & 30 & 2 & binclass & Yes \\
   TALENT & PieChart3 & Other & 1077 & 37 & 2 & binclass & Yes \\
   TALENT & Pima\_Indians\_Diabetes & Healthcare & 768 & 8 & 2 & binclass & Yes\\
   TALENT & PizzaCutter3 & Other & 1043 & 37 & 2 & binclass & Yes\\
   TALENT & Pumpkin\_Seeds & Other & 2500 & 12 & 2 & binclass & Yes\\
   TALENT & QSAR\_biodegradation & Healthcare & 1054 & 41 & 2 & binclass & Yes\\
   TALENT & Rain\_in\_Australia & Other & 145460 & 18 & 3 & multiclass & No \\
   TALENT & SDSS17 & Other & 100000 & 12 & 3 & multiclass & Yes\\
   TALENT & Satellite & Other & 5100 & 36 & 2 & binclass & Yes\\
   TALENT & Smoking\_and\_Drinking & Other & 991346 & 23 & 2 & binclass & No\\
   TALENT & Telecom\_Churn\_Dataset & Telcom & 3333 & 17 & 2 & binclass & Yes\\
   TALENT & UJI\_Pen\_Characters & Other & 1364 & 80 & 35 & multiclass & Yes\\
   TALENT & Water\_Quality\_and\_Potability & Manufacturing & 3276 & 8 & 2 & binclass & Yes\\
   TALENT & Wilt & Other & 4821 & 5 & 2 & binclass & Yes\\
   TALENT & abalone & Other & 4177 & 8 & 3 & multiclass & Yes \\
   TALENT & accelerometer & Other & 153004 & 4 & 4 & multiclass & No \\
   TALENT & ada & Other & 4147 & 48 & 2 & binclass & Yes\\
   TALENT & ada\_agnostic & Other & 4562 & 48 & 2 & binclass & Yes\\
   TALENT & ada\_prior & Other & 4562 & 14 & 2 & binclass & Yes\\
   TALENT & airlines\_seed\_0\_nrows & Telcom & 2000 & 7 & 2 & binclass & Yes \\
   TALENT & allbp & Other & 3772 & 29 & 3 & multiclass & Yes\\
   TALENT & allrep & Other & 3772 & 29 & 4 & multiclass & Yes\\
   TALENT & analcatdata\_authorship & Other & 841 & 69 & 4 & multiclass & Yes\\
   TALENT & artificial-characters & Other & 10218 & 7 & 10 & multiclass & Yes\\
   TALENT & autoUniv-au4-2500 & Other & 2500 & 100 & 3 & multiclass & Yes\\
   TALENT & autoUniv-au7-1100 & Other & 1100 & 12 & 5 & multiclass & Yes\\
   TALENT & bank & Finance & 45211 & 16 & 2 & binclass & Yes\\
   TALENT & banknote\_authentication & Finance & 1372 & 4 & 2 & binclass & Yes\\
   TALENT & baseball & Other & 1340 & 16 & 3 & multiclass & Yes\\
   TALENT & car-evaluation & Other & 1728 & 21 & 4 & multiclass & Yes\\
   TALENT & churn & Finance & 5000 & 20 & 2 & binclass & Yes \\
   TALENT & cmc & Other & 1473 & 9 & 3 & multiclass & Yes \\
   TALENT & company\_bankruptcy\_prediction & Finance & 6819 & 95 & 2 & binclass & Yes \\
   TALENT & compass & Other & 16644 & 17 & 2 & binclass & Yes \\
   TALENT & contraceptive\_method\_choice & Other & 1473 & 9 & 3 & multiclass & Yes\\
   TALENT & credit & Finance & 16714 & 10 & 2 & binclass & Yes\\
   TALENT & customer\_satisfaction\_in\_airline & Retail & 129880 & 21 & 2 & binclass & No\\
   TALENT & dabetes\_130-us\_hospitals & Healthcare & 101766 & 20 & 2 & binclass & Yes\\
   TALENT & default\_of\_credit\_card\_clients & Finance & 30000 & 23 & 2 & binclass & Yes\\
   TALENT & delta\_ailerons & Other & 7129 & 5 & 2 & binclass & Yes\\
   TALENT & dis & Other & 3772 & 29 & 2 & binclass & Yes\\
   TALENT & dna & Healthcare & 3186 & 180 & 3 & multiclass & Yes\\
   TALENT & drug\_consumption & Healthcare & 1884 & 12 & 7 & multiclass & Yes\\
   TALENT & dry\_bean\_dataset & Other & 13611 & 16 & 7 & multiclass & Yes\\
   TALENT & eeg-eye-state & Other & 14980 & 14 & 2 & binclass & Yes\\
   TALENT & electricity & Manufacturing & 45312 & 8 & 2 & binclass & Yes\\
   TALENT & estimation\_of\_obesity\_levels & Other & 2111 & 16 & 7 & multiclass & Yes\\
   TALENT & eye\_movements & Healthcare & 10936 & 27 & 3 & multiclass & Yes\\
   TALENT & eye\_movements\_bin & Healthcare & 7608 & 20 & 2 & binclass & Yes\\
   TALENT & first-order-theorem-proving & Other & 6118 & 51 & 6 & multiclass & Yes\\
   TALENT & gas-drift & Other & 13910 & 128 & 6 & multiclass & Yes\\
   TALENT & gina\_agnostic & Other & 3468 & 970 & 2 & binclass & Yes\\
   TALENT & golf\_play\_dataset\_extended & Other & 1095 & 9 & 2 & binclass & Yes\\
   TALENT & heloc & Other & 10000 & 22 & 2 & binclass & Yes\\
   TALENT & hill-valley & Other & 1212 & 100 & 2 & binclass & No\\
   TALENT & house\_16H & Other & 13488 & 16 & 2 & binclass & Yes\\
   TALENT & htru & Other & 17898 & 8 & 2 & binclass & Yes\\
   TALENT & ibm-employee-performance & Other & 1470 & 30 & 2 & binclass & Yes\\
   TALENT & in\_vehicle\_coupon\_recos & Retail & 12684 & 21 & 2 & binclass & Yes\\
   TALENT & internet\_firewall & Other & 65532 & 7 & 4 & multiclass & Yes\\
   TALENT & internet\_usage & Other & 10108 & 70 & 46 & multiclass & Yes\\
   TALENT & jm1 & Other & 10885 & 21 & 2 & binclass & No\\
   TALENT & jungle\_chess\_2pcs\_raw & Other & 44819 & 6 & 3 & multiclass & Yes\\
   TALENT & kc1 & Other & 2109 & 21 & 2 & binclass & No\\
   TALENT & kdd\_ipums\_la\_97-small & Other & 5188 & 20 & 2 & binclass & Yes\\
   TALENT & kr-vs-k & Other & 28056 & 6 & 18 & multiclass & Yes \\
   TALENT & kropt & Other & 28056 & 6 & 18 & multiclass & Yes\\
   TALENT & led24 & Other & 3200 & 24 & 10 & multiclass & Yes \\
   TALENT & led7 & Other & 3200 & 7 & 10 & multiclass & Yes \\
   TALENT & letter & Other & 20000 & 15 & 26 & multiclass & Yes \\
   TALENT & madeline & Other & 3140 & 259 & 2 & binclass & Yes\\
   TALENT & mammography & Other & 11183 & 6 & 2 & binclass & Yes\\
   TALENT & maternal\_health\_risk & Healthcare & 1014 & 6 & 3 & multiclass & Yes\\
   TALENT & mfeat-factors & Other & 2000 & 216 & 10 & multiclass & Yes \\
   TALENT & mfeat-fourier & Other & 2000 & 76 & 10 & multiclass & Yes\\
   TALENT & mfeat-karhunen & Other & 2000 & 64 & 10 & multiclass & Yes \\
   TALENT & mfeat-morphological & Other & 2000 & 6 & 10 & multiclass & Yes\\
   TALENT & mfeat-pixel & Other & 2000 & 240 & 10 & multiclass & Yes\\
   TALENT & mfeat-zernike & Other & 2000 & 47 & 10 & multiclass & Yes \\
   TALENT & mice\_protein\_expression & Other & 1080 & 75 & 8 & multiclass & Yes\\
   TALENT & microaggregation2 & Other & 20000 & 20 & 5 & multiclass & Yes\\
   TALENT & mobile\_c36\_oversampling & Telcom & 51760 & 6 & 2 & binclass & Yes\\
   TALENT & mozilla4 & Other & 15545 & 4 & 2 & binclass & Yes\\
   TALENT & naticusdroid+android & Healthcare & 29332 & 86 & 2 & binclass & Yes\\
   TALENT & national-longitudinal-survey-binary & Other & 4908 & 16 & 2 & binclass & Yes\\
   TALENT & okcupid\_stem & Other & 26677 & 13 & 3 & multiclass & Yes\\
   TALENT & one-hundred-plants-margin & Other & 1600 & 64 & 100 & multiclass & Yes\\
   TALENT & one-hundred-plants-shape & Other & 1600 & 64 & 100 & multiclass & Yes\\
   TALENT & one-hundred-plants-texture & Other & 1599 & 64 & 100 & multiclass & Yes\\
   TALENT & online\_shoppers & Retail & 12330 & 14 & 2 & binclass & No\\
   TALENT & optdigits & Other & 5620 & 64 & 10 & multiclass & Yes\\
   TALENT & ozone-level-8hr & Other & 2534 & 72 & 2 & binclass & Yes\\
   TALENT & page-blocks & Other & 5473 & 10 & 5 & multiclass & Yes\\
   TALENT & pc1 & Other & 1109 & 21 & 2 & binclass & No\\
   TALENT & pc3 & Other & 1563 & 37 & 2 & binclass & No\\
   TALENT & pc4 & Other & 1458 & 37 & 2 & binclass & No\\
   TALENT & pendigits & Other & 10992 & 16 & 10 & multiclass & Yes\\
   TALENT & phoneme & Other & 5404 & 5 & 2 & binclass & Yes\\
   TALENT & pol & Other & 10082 & 26 & 2 & binclass & Yes\\
   TALENT & predict\_students\_dropout & Other & 4424 & 34 & 3 & multiclass & Yes \\
   TALENT & qsar & Other & 1055 & 40 & 2 & binclass & Yes \\
   TALENT & rice\_cammeo\_and\_osmancik & Other & 3810 & 7 & 2 & binclass & Yes\\
   TALENT & ringnorm & Other & 7400 & 20 & 2 & binclass & Yes\\
   TALENT & rl & Other & 4970 & 12 & 2 & binclass & No\\
   TALENT & satimage & Other & 6430 & 36 & 6 & multiclass & Yes\\
   TALENT & segment & Other & 2310 & 17 & 7 & multiclass & Yes\\
   TALENT & seismic+bumps & Other & 2584 & 18 & 2 & binclass & Yes\\
   TALENT & semeion & Other & 1593 & 256 & 10 & multiclass & No\\
   TALENT & shuttle & Other & 58000 & 9 & 7 & multiclass & Yes\\
   TALENT & spambase & Other & 4601 & 57 & 2 & binclass & Yes\\
   TALENT & splice & Other & 3190 & 60 & 3 & multiclass & Yes\\
   TALENT & sports\_articles\_for\_objectivity & Other & 1000 & 59 & 2 & binclass & Yes\\
   TALENT & statlog & Other & 1000 & 20 & 2 & binclass& Yes \\
   TALENT & steel\_plates\_faults & Other & 1941 & 27 & 7 & multiclass & Yes\\
   TALENT & sylvine & Other & 5124 & 20 & 2 & binclass & Yes\\
   TALENT & taiwanese\_bankruptcy & Finance & 6819 & 95 & 2 & binclass & Yes\\
   TALENT & telco-customer-churn & Telcom & 7043 & 18 & 2 & binclass & Yes\\
   TALENT & texture & Other & 5500 & 40 & 11 & multiclass & Yes\\
   TALENT & thyroid & Healthcare & 7200 & 21 & 3 & multiclass & Yes\\
   TALENT & thyroid-ann & Healthcare & 3772 & 21 & 3 & multiclass & Yes\\
   TALENT & thyroid-dis & Healthcare & 2800 & 26 & 5 & multiclass & Yes\\
   TALENT & turiye\_student\_evaluation & Other & 5820 & 32 & 5 & multiclass & Yes\\
   TALENT & twonorm & Other & 7400 & 20 & 2 & binclass & Yes\\
   TALENT & vehicle & Other & 846 & 18 & 4 & multiclass & Yes\\
   TALENT & volkert & Other & 58310 & 180 & 10 & multiclass & Yes \\
   TALENT & walking-activity & Other & 149332 & 4 & 22 & multiclass & No \\
   TALENT & wall-robot-navigation & Other & 5456 & 24 & 4 & multiclass & Yes\\
   TALENT & water\_quality & Manufacturing & 7996 & 20 & 2 & binclass & Yes\\
   TALENT & waveform-5000 & Other & 5000 & 40 & 3 & multiclass & Yes\\
   TALENT & waveform\_database\_generator & Other & 4999 & 21 & 3 & multiclass & Yes\\
   TALENT & waveform\_database\_generator-v2 & Other & 5000 & 21 & 3 & multiclass & Yes\\
   TALENT & website\_phishing & Other & 1353 & 9 & 3 & multiclass & Yes\\
   TALENT & wine & Manufacturing & 2554 & 4 & 2 & binclass & No \\
   TALENT & wine-quality-red & Manufacturing & 1599 & 4 & 6 & multiclass & Yes \\
   TALENT & wine-quality-white & Manufacturing & 4898 & 11 & 7 & multiclass & Yes \\
   TALENT & yeast & Other & 1484 & 8 & 10 & multiclass & Yes\\

\end{longtable}

\footnotesize
\begin{longtable}{p{1.5cm}p{3.5cm}p{2.5cm}rrrll}
\caption{TabZilla benchmark datasets (36 datasets via OpenML).}
\label{tab:datasets_tabzilla}\\
\toprule
Benchmark & Dataset Name & Domain & Samples & Features & Classes & Task Type & Used In Experimentation \\
\midrule
\endfirsthead

\caption[]{ Details of TabZilla benchmark datasets.} \\

\toprule
Benchmark & Dataset Name & Domain & Samples & Features & Classes & Task Type & Used In Experimentation\\
\midrule
\endhead

\midrule
\multicolumn{7}{r}{{Continued on next page}} \\
\midrule
\endfoot

\bottomrule
\endlastfoot

   TabZilla & OpenML-ID-999 & Health & 226 & 70 & 2 & binclass & Yes\\
   TabZilla & OpenML-ID-10 & Health & 148 & 19 & 4 & multiclass & Yes\\
   TabZilla & OpenML-ID-11 & Other & 625 & 5 & 3 & multiclass & Yes \\
   TabZilla & OpenML-ID-14 & Other & 2000 & 77 & 10 & multiclass & Yes \\
   TabZilla & OpenML-ID-22 & Other & 2000 & 48 & 10 & multiclass & Yes \\
   TabZilla & OpenML-ID-29 & Finance & 690 & 16 & 2 & binclass  & Yes\\
   TabZilla & OpenML-ID-27 & Health & 368 & 23 & 2 & binclass & Yes\\
   TabZilla & OpenML-ID-31 & Finance & 1000 & 21 & 2 & binclass & Yes \\
   TabZilla & OpenML-ID-46 & Other & 3190 & 61 & 3 & multiclass & Yes \\
   TabZilla & OpenML-ID-54 & Other & 846 & 19 & 4 & multiclass & Yes \\
   TabZilla & OpenML-ID-333 & Other & 556 & 7 & 2 & binclass & Yes \\
   TabZilla & OpenML-ID-1067 & Other & 2109 & 22 & 2 & binclass & Yes \\
   TabZilla & OpenML-ID-1468 & Other & 1080 & 857 & 9 & multiclass & Yes \\
   TabZilla & OpenML-ID-1494 & Other & 1055 & 42 & 2 & binclass & Yes \\
   TabZilla & OpenML-ID-43973 & Other & 3172 & 6 & 2 & binclass & Yes \\
   TabZilla & OpenML-ID-1043 & Other & 4562 & 49 & 2 & binclass & Yes \\
   TabZilla & OpenML-ID-43945 & Other & 38474 & 9 & 2 & binclass & Yes\\
   TabZilla & OpenML-ID-1486 & Other & 34465 & 119 & 2 & binclass & No\\
   TabZilla & OpenML-ID-42825 & Other & 8378 & 123 & -- & -- & No\\
   TabZilla & OpenML-ID-4538 & Other & 9873 & 33 & 5 & multiclass & Yes \\
   TabZilla & OpenML-ID-23512 & Other & 98050 & 29 & 2 & binclass & No\\
   TabZilla & OpenML-ID-4134 & Other & 3751 & 1777 & 2 & binclass & Yes\\
   TabZilla & OpenML-ID-470 & Other & 672 & 10 & 2 & binclass & No\\
   TabZilla & OpenML-ID-1493 & Other & 1599 & 65 & 100 & multiclass & Yes\\
   TabZilla & OpenML-ID-1459 & Other & 10218 & 8 & 10 & multiclass & Yes\\
   TabZilla & OpenML-ID-41027 & Games & 44819 & 7 & 3 & multiclass & Yes\\
   TabZilla & OpenML-ID-40981 & Other & 690 & 15 & 2 & binclass & Yes\\
   TabZilla & OpenML-ID-934 & Other & 1156 & 6 & 2 & binclass & Yes\\
   TabZilla & OpenML-ID-1565 & Health & 294 & 14 & 5 & multiclass & Yes\\
   TabZilla & OpenML-ID-41150 & Other & 130064 & 51 & 2 & binclass & No\\
   TabZilla & OpenML-ID-41159 & Other & 20000 & 4297 & 2 & binclass & No\\
   TabZilla & OpenML-ID-846 & Other & 16599 & 19 & 2 & binclass & Yes \\
   TabZilla & OpenML-ID-1169 & Other & 539383 & 8 & 2 & binclass & No\\
   TabZilla & OpenML-ID-41147 & Other & 425240 & 79 & 2 & binclass & Yes\\
   TabZilla & OpenML-ID-41143 & Other & 2984 & 145 & 2 & binclass & Yes\\
   TabZilla & OpenML-ID-1567 & Other & 1025009 & 11 & 10 & multiclass & No\\

\end{longtable}

\textbf{Note}: Some TabZilla dataset statistics are to be completed. TabZilla datasets are accessed via OpenML and may include both dataset IDs and task IDs (TabZilla automatically handles task-to-dataset conversion).

\footnotesize
\begin{longtable}{p{1.5cm}p{4cm}p{1.5cm}rrrp{3cm}}
\caption{Datasets used for fairness evaluation (contain demographic/sensitive attributes).}
\label{tab:datasets_fairness}\\
\toprule
Source & Dataset Name & Domain & Samples & Features & Classes & Sensitive Features \\
\midrule
\endfirsthead

\caption[]{ Common datasets used for fairness evaluation.} \\

\toprule
Source & Dataset Name & Domain & Samples & Features & Classes & Sensitive Features \\
\midrule
\endhead

\midrule
\multicolumn{7}{r}{{Continued on next page}} \\
\midrule
\endfoot

\bottomrule
\endlastfoot

   UCI & Adult Census Income & Other & 48842 & 14 & binclass & Race, Sex, Age \\
   OpenML & German Credit (ID-31) & Finance & 1000 & 21 & binclass & Age, Personal-status-sex \\
   Kaggle & COMPAS Recidivism & Legal & ~7000 & 53 & binclass & Race, Sex \\
   UCI & Default of Credit Card Clients (Taiwan) & Finance & 30000 & 23 & binclass & Sex, Education, Marital Status \\
   UCI & Student Performance in Exams (Portuguese) & Education & 649 & 33 & binclass & Gender, Race/Ethnicity, Parental Level of Education \\
   Kaggle & Law School Admission & Legal & ~20000 & 12 & binclass & Race, Sex \\
   Kaggle & IBM HR Analytics Employee Attrition & Other & 1470 & 31 & binclass & Gender, Age, MaritalStatus, RelationshipSatisfaction \\
   Offical HMDA API & Home Mortgage Disclosure Act (NY 2023) & Finance & ~32000 & 20+ & multiclass & Race ,Ethnicity, Sex, Age \\
   Kaggle & NHANES Demographics (2013-14) & Health & 5000 & 10+ & multiclass & Gender, Age, Race, Ethnicity \\

\end{longtable}

\textbf{Note}: Fairness evaluation is performed on datasets where sensitive features are explicitly identified. The sensitive features column indicates demographic or protected attributes available in each dataset. Fairness metrics (SPD, EOD, EOpD) measure disparities in model predictions across different groups defined by these sensitive features.

\section{Appendix C: Glossary of Key Terms}

This glossary provides concise definitions of technical terms used throughout this paper. For detailed explanations, refer to the relevant sections indicated.

\begin{table}[h]
\centering
\caption{Glossary of key technical terms.}
\label{tab:glossary}
\small
\begin{tabular}{p{4.5cm}p{10cm}}
\toprule
Term & Definition \\
\midrule
\textbf{Brier Score} & Metric quantifying both calibration and refinement by penalizing squared deviations between predicted probabilities and binary outcomes; lower is better (see Section~\ref{sec:experiments} for details). \\
\midrule
\textbf{Equalized Odds Difference (EOD)} & Measure of disparities in both true positive and false positive rates across demographic groups; more stringent than statistical parity, requiring consistent error profiles (zero indicates perfect equity). \\
\midrule
\textbf{Equalized Opportunity Difference (EOpD)} & Focus on true positive rate parity across demographic groups; ensures qualified individuals receive equal treatment regardless of group membership (zero indicates perfect equity). \\
\midrule
\textbf{Episodic Training} & Meta-learning approach where each training step samples disjoint support and query sets from training data; mimics few-shot learning scenarios and preserves in-context generalization (see Section~\ref{sec:related_work}). \\
\midrule
\textbf{Expected Calibration Error (ECE)} & Average deviation between predicted confidence and observed accuracy across binned predictions; lower values indicate better calibration alignment (see Section~\ref{sec:experiments}). \\
\midrule
\textbf{Low-Rank Adaptation (LoRA)} & Parameter-efficient fine-tuning technique that adds trainable low-rank matrices to approximate full weight updates; only adapter matrices are updated, base weights remain frozen (see Section~\ref{sec:related_work}). \\
\midrule
\textbf{Maximum Calibration Error (MCE)} & Worst-case calibration gap across all prediction bins; identifies regions of systematic miscalibration and extreme confidence misalignment (see Section~\ref{sec:experiments}). \\
\midrule
\textbf{Meta-Learning (Episodic Fine-Tuning)} & Training strategy using support-query splits that mimics inference-time environment; preserves in-context generalization while enabling task-specific adaptation (see Section~\ref{sec:related_work}). \\
\midrule
\textbf{Parameter-Efficient Fine-Tuning (PEFT)} & Adaptation methods that update only a subset of model parameters, reducing computational costs by 60--80\% while preserving performance; includes LoRA and similar techniques (see Section~\ref{sec:related_work}). \\
\midrule
\textbf{Statistical Parity Difference (SPD)} & Absolute difference in positive prediction rates across demographic groups; tests demographic parity, where zero indicates perfect parity in prediction distribution (see Section~\ref{sec:experiments}). \\
\midrule
\textbf{Supervised Fine-Tuning (SFT)} & Full parameter updates on labeled data using standard gradient descent; updates all model parameters, computationally intensive but can achieve highest task-specific accuracy (see Section~\ref{sec:related_work}). \\
\midrule
\textbf{Zero-Shot Inference} & Prediction using pretrained model weights without parameter updates; models receive training samples as context during inference, performing in-context learning through forward passes alone (see Section~\ref{sec:related_work}). \\
\bottomrule
\end{tabular}
\end{table}

\end{document}